\documentclass{article}

\PassOptionsToPackage{numbers,sort}{natbib}

\usepackage[final]{neurips_2022}

\usepackage[utf8]{inputenc} % allow utf-8 input
\usepackage[T1]{fontenc}    % use 8-bit T1 fonts
\usepackage{url}            % simple URL typesetting
\usepackage{booktabs}       % professional-quality tables
\usepackage{amsfonts}       % blackboard math symbols
\usepackage{nicefrac}       % compact symbols for 1/2, etc.
\usepackage{microtype}      % microtypography
\usepackage{xcolor}         % colors
\usepackage{float}

\usepackage{amsmath}
\usepackage{bm}
\usepackage{url}
\usepackage{supertabular}
\usepackage{algorithm}
\usepackage[noend]{algorithmic}

\usepackage{wrapfig}
\usepackage{booktabs} 
\usepackage{graphicx}
\usepackage{diagbox}
\usepackage{tabularx}
\usepackage{subfigure}
\usepackage{caption}
\newcommand{\tabincell}[2]{\begin{tabular}{@{}#1@{}}#2\end{tabular}}
\usepackage{multirow}
\usepackage{color}
\usepackage{array}

\usepackage{xspace}

\usepackage[backref]{hyperref} 
\hypersetup{
colorlinks=true,
linkcolor=black,
citecolor=brown
}

\makeatletter
\DeclareRobustCommand\onedot{\futurelet\@let@token\@onedot}
\def\@onedot{\ifx\@let@token.\else.\null\fi\xspace}

\def\eg{\emph{e.g}\onedot} 

\def\ie{\emph{i.e}\onedot}

\def\etc{\emph{etc}\onedot} 

\def\wrt{\emph{w.r.t}\onedot}

\title{Boosting the Transferability of Adversarial Attacks with Reverse Adversarial Perturbation}

\author{
Zeyu Qin\textsuperscript{1*}, \ Yanbo Fan\textsuperscript{2*}, \ Yi Liu\textsuperscript{1}, \ Li Shen\textsuperscript{3}, \ Yong Zhang\textsuperscript{2}, \ 
Jue Wang\textsuperscript{2}, \ Baoyuan Wu\textsuperscript{1$\dagger$}\\
\textsuperscript{1}School of Data Science, Shenzhen Research Institute of Big Data, \\
The Chinese University of Hong Kong, Shenzhen\\
\textsuperscript{2}Tencent AI Lab\\
\textsuperscript{3}JD Explore Academy\\
\texttt{\{zeyu6181136, fanyanbo0124, yiliuhk2000\}@gmail.com}\\
\texttt{\{mathshenli, zhangyong201303, arphid\}@gmail.com} \\
\texttt{wubaoyuan@cuhk.edu.cn}
}

\begin{document}

\maketitle

\begin{abstract}

Deep neural networks (DNNs) have been shown to be vulnerable to adversarial examples, which can produce erroneous predictions by injecting imperceptible perturbations.
In this work, we study the transferability of adversarial examples, which is significant due to its threat to real-world applications where model architecture or parameters are usually unknown.
Many existing works reveal that the adversarial examples are likely to overfit the surrogate model that they are generated from, limiting its transfer attack performance against different target models.
To mitigate the overfitting of the surrogate model, we propose a novel attack method, dubbed {\it reverse adversarial perturbation} (RAP).
Specifically, instead of minimizing the loss of a single adversarial point,
we advocate seeking adversarial example located at a region with unified low loss value, by injecting the worst-case perturbation (\ie, the reverse adversarial perturbation) for each step of the optimization procedure.
The adversarial attack with RAP is formulated as a min-max bi-level optimization problem. 
By integrating RAP into the iterative process for
attacks, our method can find more stable adversarial examples which are less sensitive to the changes of decision boundary, mitigating the overfitting of the surrogate model. 
Comprehensive experimental comparisons demonstrate that RAP can significantly boost adversarial transferability.
Furthermore, RAP can be naturally combined with many existing black-box attack techniques, to further boost the transferability.
When attacking a real-world image recognition system, \ie, Google Cloud Vision API, we obtain 22\% performance improvement of targeted attacks over the compared method.  Our codes are available at: \url{
https://github.com/SCLBD/Transfer_attack_RAP}.

% That also means predictions of the attack’ neighborhoods should be consistent. 
% to demonstrate the significant boost of the method on transferability,
% The detailed comparison and evaluation demonstrate that RAP effectively 

\end{abstract}

\renewcommand{\thefootnote}{\fnsymbol{footnote}}
\footnotetext[1]{Equal contribution. $^\dagger$Corresponding author.} 
% \footnotetext[2]{Corresponding author.}
\footnotetext[3]{This work is done when Zeyu Qin is a research intern at Tencent AI Lab.}
\renewcommand{\thefootnote}{\arabic{footnote}}

\vspace{-0.02in}
\section{Introduction}
\label{introduction}
% \vspace{-0.02in}
Deep neural networks (DNNs) have been successfully applied in many safety-critical tasks, such as autonomous driving, face recognition and verification, \etc. However, it has been shown that DNN models are vulnerable to adversarial examples \citep{szegedy2013intriguing,goodfellow2014explaining,madry2018towards,fan2020sparse,xiao2022adversarial,zheng2022robust,xiao2022stability,qin2021random}, which are indistinguishable from natural examples but make a model produce erroneous predictions. For real-world applications, the DNN models are often hidden from users. Therefore, the attackers need to generate the adversarial examples under black-box setting where they do not know any information of the target model \citep{cheng2018queryefficient, Cheng2020Sign-OPT,ilyas2018black,qin2021random}. For black-box setting, the adversarial transferability matters since it can allow the attackers to attack target models by using adversarial examples generated on the surrogate models. Therefore, learning how to generate adversarial examples with high transferability has gained more attentions in the literature \citep{liu2016delving,tramer2018ensemble,dong2018boosting,xie2019improving,GuoLinBP,Feng_2022_CVPR,liang2021parallel}. 

% The several gradient-based attacks used under white-box setting have been also adopted to generate transfer attacks, such as single-step attacks \citep{goodfellow2014explaining} and iterative attacks \citep{Kurakin_2018}. Under white-box setting, with the knowledge of surrogate model, iterative attacks achieve much better white-box attack performance. Hoewever, they often exhibit the poor success rates for the black-box settings. The previous work \citep{dong2018boosting,xie2019improving,dong2019evading,tramer2018ensemble,Lin2020Nesterov,Why,GuoLinBP} attribute that to the overfitting of adversarial examples to the surrogate models and propose methods to improve the transferability. 

{
% The gradient-based attacks used under white-box setting have been  adopted to generate transfer attacks, such as single-step attacks \citep{goodfellow2014explaining} and iterative attacks \citep{Kurakin_2018}.
Under white-box setting where the complete information of the attacked model (\eg, architecture and parameters) is available, 
the gradient-based attacks such as PGD \cite{madry2018towards} have demonstrated good attack performance.
However, they often exhibit the poor transferiability \citep{xie2019improving,dong2018boosting}, \ie, the adversarial example $\bm{x}^{adv}$ generated from the surrogate model $\mathcal{M}^{S}$ performs poorly against different target models $\mathcal{M}^{T}$. 
The previous works attribute that to the overfitting of adversarial examples to the surrogate models \citep{dong2018boosting,xie2019improving,Lin2020Nesterov}. 
% \zeyu{\sout{As shown in Figure \ref{fig_vis} (b), the current attack $\bm{x}^{pgd}$ successfully attacks $\mathcal{M}^{S}$, but it is not stable and sensitive to changes of $\mathcal{M}^{S}$. When having some changes on current model parameters (new $\mathcal{M}^{S'}$), $\bm{x}^{pgd}$ will have the high loss value and even lead to inconsistent prediction for $\mathcal{M}^{S'}$.}}
Figure~\ref{fig_vis} (b) gives an illustration.
The PGD attack aims to find an adversarial point $\bm{x}^{pgd}$ with minimal attack loss, while doesn't consider the attack loss of the neighborhood regions round $\bm{x}^{pgd}$.
Due to the highly non-convex of deep models, when $\bm{x}^{pgd}$ locates at a sharply local minimum, a slight change on model parameters of $\mathcal{M}^{S}$ could cause a large increase of the attack loss, making $\bm{x}^{pgd}$ fail to attack the perturbed model.

Many techniques have been proposed to mitigate the overfitting and improve the transferability, including input transformation \citep{dong2019evading,xie2019improving}, gradient calibration \citep{GuoLinBP}, feature-level attacks \citep{huang2019enhancing}, and generative models \citep{naseer2021generating}, etc. %
%\citep{dong2018boosting,xie2019improving,dong2019evading,tramer2018ensemble,Lin2020Nesterov,Why,GuoLinBP} 
%
However, there still exists a large gap of attack performance between the transfer setting and the ideal white-box setting, especially for targeted attack, requiring more efforts for boosting the transferability. 
% Through these the overfitting can be and the transferability has been improved accordingly, the attack performance are still inferior compared to the ideal white-box settings, especially for targeted attacks.
}

{
}

% \zeyu{\sout{In this work, we propose a novel attack method called {\it reverse adversarial perturbation} (RAP) to further alleviate the overfitting for source model $\mathcal{M}^{S}$ and boost the transferability of adversarial examples $\bm{x}^{adv}$. 
% The key idea is to seek an adversarial example located at the flat local minimum. We encourage that $\bm{x}^{adv}$ is not only of low attack loss but also locates at a local flat region, \ie, the points within the local neighborhood region around $\bm{x}^{adv}$ should also of low loss values. As shown in Figure \ref{fig_vis} (b), we tend to find the yellow point, $\bm{x}^{adv}$ which locates at the local flat region. Even if the model parameter of $\mathcal{M}^S$ has some slight changes, the loss function will keep relatively low value and $\bm{x}^{adv}$ could still have consistent predictions for these two models.}
% }
%

In this work, we propose a novel attack method called {\it reverse adversarial perturbation} (RAP) to alleviate the overfitting of the surrogate model and boost the transferability of adversarial examples.
% The key idea is to seek an adversarial example $\bm{x}^{adv}$ located at a flat local minimum. 
We encourage that $\bm{x}^{adv}$ is not only of low attack loss but also locates at a local flat region, \ie, the points within the local neighborhood region around $\bm{x}^{adv}$ should also be of low loss values. 
Figure~\ref{fig_vis} (b) illustrates the difference between the sharp local minimum and flat local minimum.
When the model parameter of $\mathcal{M}^S$ has some slight changes, the variation of the attack loss \wrt the flat local minimum is less than that of the sharp one.
Therefore, the flat local minimum is less sensitive to the changes of decision boundary.
To achieve this goal, we formulate a min-max bi-level optimization problem. The inner maximization aims to find the worst-case perturbation (\ie, that with the largest attack loss, and this is why we call it reverse adversarial perturbation) within the local region around the current adversarial example, which can be solved by the projected gradient ascent algorithm. Then, the outer minimization will update the adversarial example to find a new point added with the provided reverse perturbation that leads to lower attack loss. 
% \zeyu{\sout{As shown in Figure \ref{fig_vis} (a), RAP first finds the point $\bm{x}^{0}+\bm{n}^{rap}$ with max loss within the neighborhood region of $\bm{x}^{0}$ and updates $\bm{x}^{0}$ by using the gradient of $\bm{x}^{0}+\bm{n}^{rap}$ rather than its own gradient. After updating, the attack has become the yellow point $\bm{x}^{adv}$ located at the local flat region. Therefore, rather than trapping in the sharp minimum like the original attack $\bm{x}^{pgd}$, RAP could help us escape from the sharp local minimum and find attacks at the flat local minimum.}}
Figure~\ref{fig_vis} (a) provides an illustration of the optimization process. 
For $t$-th iteration and $\bm{x}^{t}$, RAP first finds the point $\bm{x}^{t}+\bm{n}^{rap}$ with max attack loss within the neighborhood region of $\bm{x}^{t}$.
Then it updates $\bm{x}^{t}$ with the gradient calculated by minimizing the attack loss \wrt $\bm{x}^{t}+\bm{n}^{rap}$. 
Compared to directly adopting the gradient at $\bm{x}^{t}$, RAP could help escape from the sharp local minimum and pursue a relatively flat local minimum.
Besides, we design a late-start variant of RAP (RAP-LS) to further boost the attack effectiveness and efficiency, which doesn't insert the reverse perturbation into the optimization procedure in the early stage.
% \zeyu{Extensive experiments on attacking different DNN models have empirically verified that the proposed RAP can effectively mitigate attacks' overfitting for $\mathcal{M}^{S}$ and improve their adversarial transferability.}
Moreover, from the technical perspective, since the proposed RAP method only introduces one specially designed perturbation onto adversarial attacks, one notable advantage of RAP is that it can be naturally combined with many existing black-box attack techniques to further boost the transferability. 
For example, when combined with different input transformations (\eg, the random resizing and padding in Diverse Input \citep{xie2019improving}), our RAP method consistently outperforms the counterparts by a clear margin.

Our main contributions are three-fold:
% {1)} we advocate considering the flatness of loss landscape around the adversarial example to boost the adversarial tranferability;
\textbf{1)}
% \zeyu{\sout{To mitigate the overfitting of attack for source model, we propose the {\it reverse adversarial perturbation} (RAP) attack to find the adversarial examples located at the flat local minimum. RAP explicitly encourages the local flatness around the adversarial example by injecting the worst-case perturbation;}}
Based on a novel perspective, the flatness of loss landscape for adversarial examples, we propose a novel adversarial attack method RAP that encourages both the adversarial example and its neighborhood region be of low loss value;
\textbf{2)} we present a vigorous experimental study and show that RAP can significantly boost the adversarial transferability on both untargeted and targeted attacks for various networks also containing defense models;
\textbf{3)} we demonstrate that RAP can be easily combined with existing transfer attack techniques and outperforms the state-of-the-art performance by a large margin.

\section{Related Work}
\label{related work}
% \vspace{-0.02in}

The black-box attacks can be categorized into two categories: 1) {\textit{query-based attacks}} that conduct the attack based on the feedback of iterative queries to target models, and 2) {\textit{transfer attacks}} that use the adversarial examples generated on some surrogate models to attack the target models.
In this work, we focus on the transfer attacks. 
For surrogate models, existing attack algorithms such as FGSM \citep{goodfellow2014explaining} and I-FGSM \citep{Kurakin_2018} could achieve good attack performance.
However, they often overfit the surrogate models and thus exhibit poor transferability.
Recently, many works have been proposed to generate more transferable adversarial examples \cite{dong2018boosting,xie2019improving,Lin2020Nesterov,dong2019evading,wang2021admix,wang2021adversarial,Wang_2021_CVPR,Wu2020Skip,GuoLinBP,huang2019enhancing,Inkawhich2020Transferable,inkawhich2020perturbing,gubri2022lgv}, which we briefly summarize as below.
% Here we categorise them into five classes: momentum-based attack, input transformation attack, model-specific attack, feature-based attack, and generative attack. 

\textbf{Input transformation:} 
Data augmentation, which has been shown to be effective in improving model generalization, has also been studied to boost the adversarial transferability, such as randomly resizing and padding \citep{xie2019improving}, randomly scaling \citep{Lin2020Nesterov}, and adversarial mixup \citep{wang2021admix}.
In addition, the work of \citet{dong2019evading} uses a set of translated images to compute gradient and get the better performance against defense models. 
Expectation of Transformation (EOT) method \citep{athalye2018synthesizing} synthesizes adversarial examples over a chosen distribution of transformations to enhance its adversarial transferability. 
% It could find the stable $\bm{x}_{adv}$ to be robust to noise, distortion, and affine transformation, which means it also finds $\bm{x}_{adv}$ at the flat local minimum.  
%
\textbf{Gradient modification:} Instead of the I-FGSM, the work of \citet{dong2018boosting} integrates momentum into the updating strategy. And \citet{Lin2020Nesterov} uses the Nesterov accelerated gradient to boost the transferability. 
The work of \citet{Wang_2021_CVPR} aims to find a more stable gradient direction by tuning the variance of each gradient step.
%
% \textbf{Network architecture modification:}
There are also some model-specific designs to boost the adversarial transferability.
For example, \citet{Wu2020Skip} found that the gradient of skip connections is more crucial to generate more transferable attacks. 
The work of \citet{GuoLinBP} proposed LinBP to utilize more gradient of skip connections during the back-propagation. 
However, these methods tend to be specific to a particular model architecture, such as skip connection, and it is nontrivial to extend the findings to other architectures or modules.
\textbf{Intermediate feature attack:}
Meanwhile, \citet{huang2019enhancing,Inkawhich2020Transferable,inkawhich2020perturbing} proposed to exploit feature space constraints to generate more transferable attacks. 
Yet they need to identity the best performing intermediate layers or train one-vs-all binary classifies for all attacked classes.
Recently, \citet{zhao2020success} find iterative attacks with much more iterations and logit loss can achieve relatively high targeted transferability and exceed the feature-based attacks. 
\textbf{Generative models:}
In addition, there have been some methods utilizing the generative models to generate the adversarial perturbations \citep{poursaeed2018generative,naseer2019cross,naseer2021generating}.  
For example, the work of \citet{naseer2021generating} proposed to train a generative model to match the distributions of source and target class, so as to increase the targeted transferability.  
However, the learning of the perturbation generator is nontrivial, especially on large-scale datasets.

In summary, the current performance of transfer attacks is still unsatisfactory, especially for targeted attacks.
In this work, we study adversarial transferability from the prespective of the flatness of adversarial examples. 
We find that adversarial examples located at flat local minimum will be more transferable than those at sharp local minimum and  propose an novel algorithm to find adversarial example that locates at flat local minimum. 
\section{Methodology}
\label{methodology}
% \vspace{-0.07in} 
\subsection{Preliminaries of Transfer Adversarial Attack}

Given an benign sample $(\bm{x}, y) \in (\mathcal{X}, \mathcal{Y})$, the procedure of transfer adversarial attack is firstly constructing the adversarial example $\bm{x}^{adv}$ within the neighborhood region $\mathcal{B}_{\epsilon} (\bm{x})  = \{\bm{x}^{\prime}: \|\bm{x}^{\prime}-\bm{x}\|_{p} \leq \epsilon\}$ by attacking the white-box surrogate model $\mathcal{M}^s(\bm{x}; \bm{\theta}): \mathcal{X} \rightarrow \mathcal{Y}$, then transferring $\bm{x}^{adv}$ to directly attack the black-box target model $\mathcal{M}^t(\bm{x}; \bm{\phi}): \mathcal{X} \rightarrow \mathcal{Y}$. 
The attack goal is to mislead the target model, \ie, $\mathcal{M}^{t}(\bm{x}^{adv};\bm{\phi}) \neq y$ (untargeted attack), or $\mathcal{M}^{t}(\bm{x}^{adv};\bm{\phi}) = y_t$ (targeted attack) with $y_t \in \mathcal{Y}$ indicting the target label. 
Taking the target attack as example, the general formulation of many existing transfer attack methods can be written as follows:
\begin{equation}
    \min_{\bm{x}^{adv}\in \mathcal{B}_{\epsilon} (\bm{x})}  \mathcal{L} (\mathcal{M}^{s} (\mathcal{G}(\bm{x}^{adv});\bm{\theta}), y_{t}).
    \label{eq_general_formulation_of_transfer_attack}
\end{equation}
The loss function $\mathcal{L}$ is often set as the cross entropy (CE) loss \citep{xie2019improving} or the logit loss \citep{zhao2020success}, which will be specified in later experiments. 
Besides, the formulation of untargeted attack can be easily obtained by replacing the loss function $\mathcal{L}$ and $y_t$ by $-\mathcal{L}$ and $y$, respectively. 

Since $\mathcal{M}^{s}$ is white-box, if $\mathcal{G}(\cdot)$ is set as the identity function, then any off-the-shelf white-box adversarial attack method can be adopted to solve Problem (\ref{eq_general_formulation_of_transfer_attack}),  such as I-FSGM \citep{Kurakin_2018}, MI-FGSM \citep{dong2018boosting}, \etc. 
Meanwhile, existing works have designed different $\mathcal{G}(\cdot)$ functions and developed the corresponding optimization algorithms, to boost the adversarial transferability between surrogate and target models. For example, $\mathcal{G}(\cdot)$ is specified as random resizing and padding (DI) \citep{xie2019improving}, translation transformation (TI)  \citep{dong2019evading}, scale transformation (SI)  \citep{Lin2020Nesterov}, and adversarial mixup (Admix) \citep{wang2021admix}. 

%denotes a transformation function, which plays an important role of promoting the adversarial transferability, such as random resizing and padding in Diverse Input (DI) method \citep{xie2019improving}, translation transformation in Translation-Invariant (TI) method \citep{dong2019evading}, scale transformation in Scale-Invariant (SI) method \citep{Lin2020Nesterov}, as well as adversarial mixup in Admix method \citep{wang2021admix}. 

% \vspace{-0.2cm}

\begin{figure*}[t]
\vskip -0.07in
\centering
\scalebox{1}{
\subfigure[]{
\begin{minipage}[htp]{0.45\linewidth}
\centering
\includegraphics[width=2.2in]{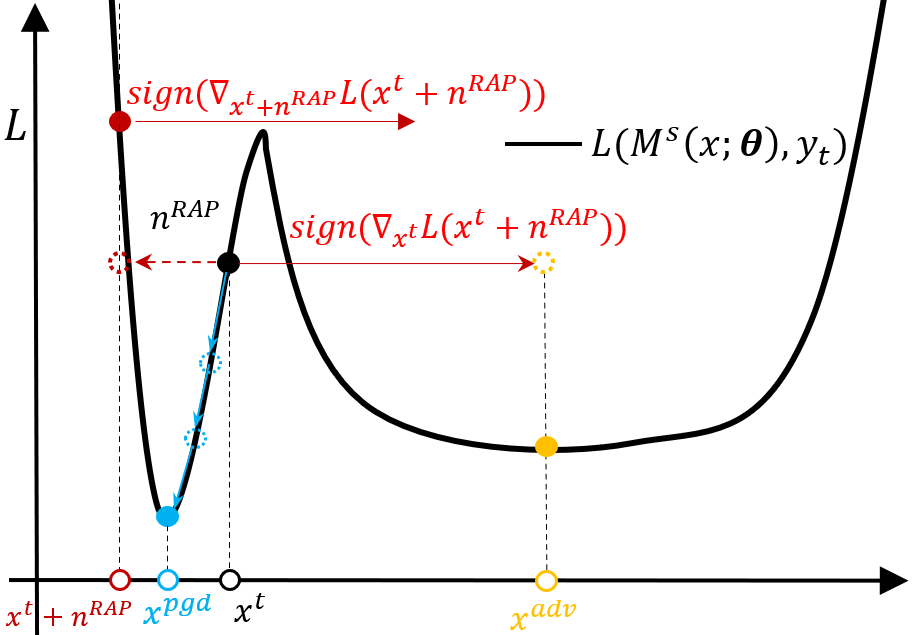}
\end{minipage}%
}%
\subfigure[]{
\begin{minipage}[htp]{0.5\linewidth}
\centering
\includegraphics[width=2.05in]{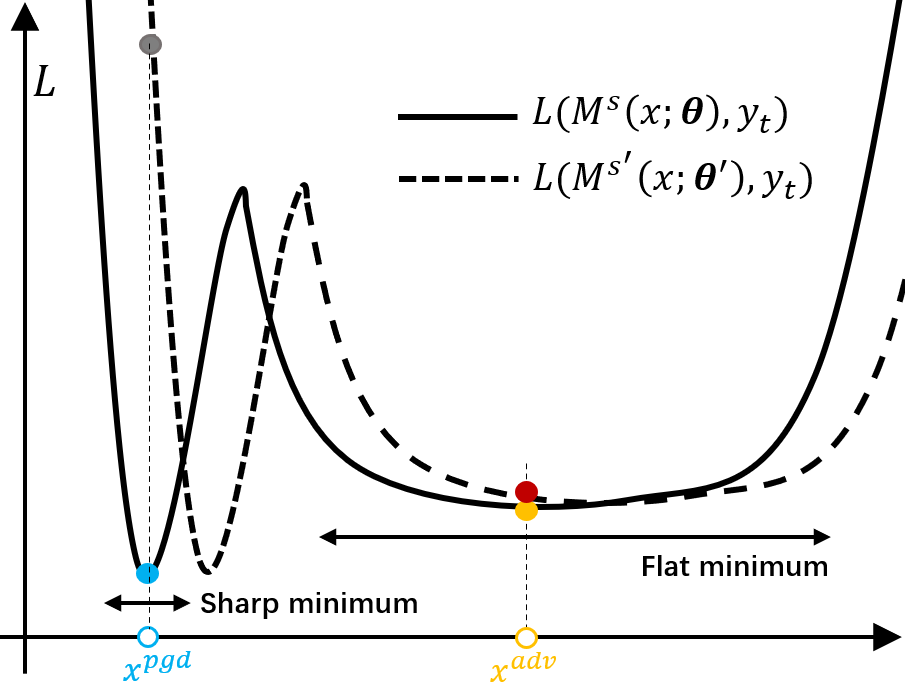}
\end{minipage}%
}}
\centering
\vskip -0.12in
\caption{These two plots are schematic diagrams in 1D space. The x-axis means the value of input $\bm{x}$. The y-axis means the value of attack loss function $\mathcal{L}$. (a) Illustration of our attack method and the original PGD attack. (b) Illustration of attack loss landscape of $\mathcal{M}^{S}$ and $\mathcal{M}^{S'}$. $\mathcal{M}^{S'}$ denotes a slight change on the model parameters of $\mathcal{M}^{S}$. The blue and yellow dots correspond to attacks located at different local minima on $\mathcal{M}^{S}$, respectively. The gray and red points are their counterparts on $\mathcal{M}^{S'}$.
}
\label{fig_vis}
\vskip -0.05in
\end{figure*}

\subsection{Reverse Adversarial Perturbation}
\label{method}

As discussed above, although having good performance for the white-box setting, the adversarial examples generated from $\mathcal{M}^{S}$ exist poor adversarial transferability on $\mathcal{M}^{T}$, especially for targeted attacks. 
The previous works attribute this issue to the overfitting of adversarial attack to $\mathcal{M}^{S}$ \citep{dong2018boosting,xie2018mitigating,tramer2018ensemble,dong2019evading,demontis2019adversarial}. As shown in Figure \ref{fig_vis} (b), when $\bm{x}^{pgd}$ locates at a sharp local minimum, it is not stable and is sensitive to changes of $\mathcal{M}^{S}$. When having some changes on model parameters, $\bm{x}^{pgd}$ could results in a high attack loss against $\mathcal{M}^{S'}$ and lead to a failure attack.
% Here, we take the demonstration in Figure~\ref{fig_vis} (a). The black point is the current attack, $\bm{x}^{PGD}$ which locates at the sharp local minimum. When having some changes on current model parameters (new $\mathcal{M}^{S'}$), $\bm{x}^{PGD}$ will have the high loss value and even lead to inconsistent prediction for $\mathcal{M}^{S'}$. The better adversarial attack is yellow point located at the flat local minimum. As shown in Figure~\ref{fig_vis} (a), it is more stable and not sensitive to the changes of decision boundaries. 

% The reason why $\bm{x}^{adv}$ performs the poor adversarial transferability on $\mathcal{M}^{T}$ is that $\bm{x}^{adv}$ generated on the source model $\mathcal{M}^{S}$ overfitts $\mathcal{M}^{S}$ itself \citep{dong2018boosting,xie2018mitigating,tramer2018ensemble,dong2019evading,demontis2019adversarial}. Although $\bm{x}^{adv}$ attacks $\mathcal{M}^{S}$ successfully, it severely depends decision boundary of $\mathcal{M}^{S}$. Therefore, its attack performance is very sensitive to the changes of decision boundary. Since the architecture of $\mathcal{M}^{S}$ and $\mathcal{M}^{T}$ are different from each other, there will be differences between the decision boundaries \citep{liu2016delving, tramer2017space}. When having difference on decision boundaries between $\mathcal{M}^{S}$ and $\mathcal{M}^{T}$, $\bm{x}^{adv}$ cannot successfully attack $\mathcal{M}^{T}$ {\color{red}seeing the figure}. 

To mitigating the overfitting to $\mathcal{M}^{S}$, 
% we seek to find the stable attack like the yellow point $\bm{x}^{adv}$ in Figure~\ref{fig_vis} (a). As shown in Figure \ref{fig_vis} (b), even if the model parameter of $\mathcal{M}^S$ has some slight changes, the loss function will keep relatively low value and $\bm{x}^{adv}$ could still have consistent predictions for these two models. Therefore, 
we advocate to find $\bm{x}^{adv}$ located at flat local region. That means we encourage that not only $\bm{x}^{adv}$ itself has low loss value, but also the points in the vicinity of $\bm{x}^{adv}$ have similarly low loss values.

To this end, we propose to minimize the maximal loss value within a local neighborhood region around the adversarial example $\bm{x}^{adv}$. 
The maximal loss is implemented by perturbing $\bm{x}^{adv}$ to maximize the attack loss, named \textit{Reverse Adversarial Perturbation} (RAP). 
By inserting the RAP into the formulation (\ref{eq_general_formulation_of_transfer_attack}), we aim to solve the following problem, 
\begin{equation}
\begin{aligned}
 \min_{\bm{x}^{adv}\in \mathcal{B}_{\epsilon}(\bm{x})} \mathcal{L} (\mathcal{M}^s (\mathcal{G}(\bm{x}^{adv}+\bm{n}^{rap});\bm{\theta}), y_t), 
 \label{eq: inner minimization}
\end{aligned}
\end{equation}
where
\begin{flalign}
\bm{n}^{rap} = \underset{\| \bm{n}^{rap}\|_{\infty} \leq \epsilon_n }{\arg\max} \mathcal{L} (\mathcal{M}^s (\bm{x}^{adv}+\bm{n}^{rap};\bm{\theta}), y_t),
\label{eq: outer maximization}
\end{flalign}
with $\bm{n}^{rap}$ indicating the RAP, and $\epsilon_n$ defining its search region. 
The above formulations \eqref{eq: inner minimization} and \eqref{eq: outer maximization} correspond to the targeted attack, and the corresponding untargeted formulations can be easily obtained by replacing the loss function $\mathcal{L}$ and $y_t$ by $-\mathcal{L}$ and $y$, respectively. 

%---------------------------------------------------------------------%
% \ls{Is any intuitive visualization why problem (2) works(about the problem (2) itself rather than transferbility!), such as the influence of decision boundary?}\ls{it can be placed in the introduction section.}
%---------------------------------------------------------------------%

It is a min-max bi-level optimization problem  \citep{liu2021investigating}, and can be solved by iteratively optimizing the inner maximization and the outer minimization problem. 
Specifically, in each iteration, given $\bm{x}^{adv}$, the inner maximization \textit{w.r.t.} $\bm{n}^{rap}$ is solved by the projected gradient ascent algorithm:
\begin{flalign}
\bm{n}^{rap} \leftarrow \bm{n}^{rap} + \alpha_n \cdot \text{sign}( \nabla_{\bm{n}^{rap}} \mathcal{L} (\mathcal{M}^s (\bm{x}^{adv}+\bm{n}^{rap};\bm{\theta}), y_t)). 
\label{eq: update n}
\end{flalign}
The above update is conducted by $T$ steps, and $\alpha_n = \frac{\epsilon_n}{T}$. 
Then, given $\bm{n}^{rap}$, the outer minimization \textit{w.r.t.} $\bm{x}^{adv}$ can be solved by any off-the-shelf algorithm that is developed for solving \eqref{eq_general_formulation_of_transfer_attack}. For example, it can be undated by one step projected gradient descent, as follows:
% \begin{flalign}
% \bm{x}^{adv} \leftarrow \text{Clip}_{\mathcal{B}_{\epsilon}(\bm{x})} \big[ \bm{x}^{adv} - \alpha \cdot \text{sign}( \nabla_{\bm{x}^{adv} + \bm{n}^{rap}} \mathcal{L} (\mathcal{M}^s (\mathcal{G}(\bm{x}^{adv}+\bm{n}^{rap});\bm{\theta}), y_t)) \big], 
% \label{eq: update x}
% \end{flalign}
\begin{flalign}
\bm{x}^{adv} \leftarrow \text{Clip}_{\mathcal{B}_{\epsilon}(\bm{x})} \big[ \bm{x}^{adv} - \alpha \cdot \text{sign}( \nabla_{\bm{x}^{adv}} \mathcal{L} (\mathcal{M}^s (\mathcal{G}(\bm{x}^{adv}+\bm{n}^{rap});\bm{\theta}), y_t)) \big], 
\label{eq: update x}
\end{flalign}
with $\text{Clip}_{\mathcal{B}_{\epsilon}(\bm{x})}(\bm{a})$ clipping $\bm{a}$ into the neighborhood region $\mathcal{B}_{\epsilon}(\bm{x})$. 
The overall optimization procedure is summarized in Algorithm \ref{alg}. 
% \zeyu{We also take the visualization of our method in Figure \ref{fig_vis} (a). RAP first finds the point $\bm{x}^{0}+\bm{n}^{rap}$ with max loss within the neighborhood region of $\bm{x}^{0}$ and updates $\bm{x}^{0}$ by using the gradient of $\bm{x}^{0}+\bm{n}^{rap}$ rather than its own gradient. After updating, the attack has become the yellow point $\bm{x}^{adv}$ located at the local flat region. Therefore, rather than trapping in the sharp minimum like the original attack $\bm{x}^{pgd}$, RAP could help us escape from the sharp local minimum and find attacks at the flat local minimum.}
%
Moreover, since the optimization \textit{w.r.t.} $\bm{x}^{adv}$ can be implemented by any off-the-shelf algorithm for solving Problem \eqref{eq_general_formulation_of_transfer_attack}, one notable advantage of the proposed RAP is that it can be naturally combined with any one of them, such as the input transformation methods \citep{xie2019improving,dong2019evading,Lin2020Nesterov,wang2021admix}.

% \begin{figure*}[t]
% \vskip -0.15in
% \centering
% \scalebox{0.95}{
% \subfigure{
% \begin{minipage}[htp]{0.25\linewidth}
% \centering
% \includegraphics[width=1.4in]{figures/ResNet-50 --> DenseNet-121 (MI).png}
% \end{minipage}%
% }%
% \subfigure{
% \begin{minipage}[htp]{0.25\linewidth}
% \centering
% \includegraphics[width=1.4in]{figures/ResNet-50 --> DenseNet-121 (Admix).png}
% \end{minipage}%
% }%
% \subfigure{
% \begin{minipage}[htp]{0.25\linewidth}
% \centering
% \includegraphics[width=1.4in]{figures/ResNet-50 --> VGG-16 (MI).png}

% \end{minipage}
% }%
% % \rulesep
% \subfigure{
% \begin{minipage}[htp]{0.25\linewidth}
% \centering
% \includegraphics[width=1.4in]{figures/ResNet-50 --> VGG-16 (Admix).png}
% \end{minipage}
% }}
% \centering
% \vskip -0.2in
% \small{\caption{
% Targeted attack success rate ($\%$) on DenseNet-121 and VGG-16. We take the ResNet-50 as the surrogate model and take MI and Admix as baseline methods. 
% % The three different colored lines represent the baseline, the baseline with our method, and baseline with our method and late-start.
% }
% \label{show_late_start}
% }
% \vskip -0.08in
% \end{figure*}

\begin{figure}[t]
    \vspace{-0.02in}
    \centering
    \includegraphics[width=1\textwidth]{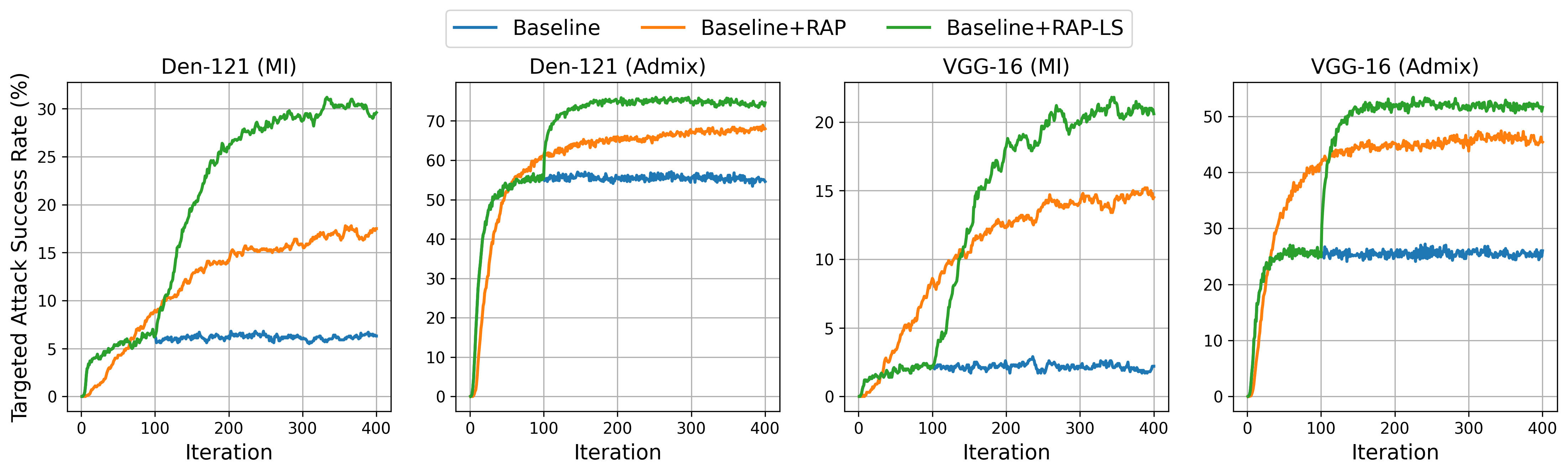}
    \vspace{-0.02in}
    \small{\caption{
    Targeted attack success rate ($\%$) on Dense-121 and VGG-16. We take the Res-50 as the surrogate model and take MI and Admix as baseline methods. 
    }
    \label{show_late_start}
    }
    \vspace{-0.15in}
\end{figure}

\paragraph{A Late-Start (LS) Variant of RAP.} 
In our preliminary experiments, we find that RAP requires more iterations to converge and the performance is slightly lower during the initial iterations, compared to its baseline attack methods. 
As shown in Figure \ref{show_late_start}, we combine MI \citep{dong2018boosting} and Admix \citep{wang2021admix} with RAP, and adopt ResNet-50 as the surrogate model. We take the evaluation on 1000 images from ImageNet (see Sec.\ref{evaluation settings}).
It is observed that the method with RAP (see the orange curves) quickly surpasses its baseline method (see the blue curves) and finally achieves much higher success rate with more iterations, which verify the effect of RAP on enhancing the adversarial transferability. 
However, it is also observed that the performance of RAP is slightly lower than its baseline method in the early stage. 
The possible reason is that the early-stage attack is of very weak attack performance to the surrogate model. In this case, it may be waste to pursue better transferable attacks by solving the min-max problem. 
A better strategy may be only solving the minimization problem \eqref{eq_general_formulation_of_transfer_attack} in the early stage to quickly achieve the region of relatively high adversarial attack performance, then starting RAP to further enhance the attack performance and transferability simultaneously. This strategy is denoted as RAP with late-start (RAP-LS), whose effect is preliminarily supported by the results shown in Figure \ref{show_late_start} (see the green curve) and will be evaluated extensively in later experiments.
\vskip -0.05in

% \vspace{-0.03in}
\subsection{A Closer Look at RAP}
% \vspace{-0.03in}

% Here, we conduct experiments to study the effect of RAP on the flatness of the loss landscape around
% the adversarial examples. 
To verify whether RAP can help us find a $\bm{x}^{adv}$ located at the local flat region or not, we use ResNet-50 as surrogate model and conduct the untargeted attacks. We visualize the loss landscape around $\bm{x}^{adv}$ on $\mathcal{M}^{S}$ by plotting the loss variations when we move $\bm{x}^{adv}$ along a random direction with different magnitudes $a$. The details of the calculation are provided in \textit{Appendix}. Figure~\ref{show_flatness} plots the visualizations. We take I-FGSM \citep{Kurakin_2018} (denoted as I), MI \citep{dong2018boosting}, DI \citep{xie2019improving}, and MI-TI-DI (MTDI) as baselines attacks and combined them with RAP. We can see that comparing to the baselines, RAP could help find $\bm{x}^{adv}$ located at the flat region. 
% \begin{figure*}[t]
% \vskip -0.15in
% \centering
% \scalebox{0.95}{
% \subfigure{
% \begin{minipage}[htp]{0.25\linewidth}
% \centering
% \includegraphics[width=1.4in]{figures/ResNet-50 --> DenseNet-121 (MI).png}
% \end{minipage}%
% }%
% \subfigure{
% \begin{minipage}[htp]{0.25\linewidth}
% \centering
% \includegraphics[width=1.4in]{figures/ResNet-50 --> DenseNet-121 (Admix).png}
% \end{minipage}%
% }%
% \subfigure{
% \begin{minipage}[htp]{0.25\linewidth}
% \centering
% \includegraphics[width=1.4in]{figures/ResNet-50 --> VGG-16 (MI).png}

% \end{minipage}
% }%
% % \rulesep
% \subfigure{
% \begin{minipage}[htp]{0.25\linewidth}
% \centering
% \includegraphics[width=1.4in]{figures/ResNet-50 --> VGG-16 (Admix).png}
% \end{minipage}
% }}
% \centering
% \vskip -0.2in
% \small{\caption{
% Targeted attack success rate ($\%$) on DenseNet-121 and VGG-16. We take the ResNet-50 as the surrogate model and take MI and Admix as baseline methods. 
% % The three different colored lines represent the baseline, the baseline with our method, and baseline with our method and late-start.
% }
% \label{show_late_start}
% }
% \vskip -0.08in
% \end{figure*}

\begin{algorithm}[tp]
% \algsetup{linenosize=\tiny} \scriptsize
  \caption{Reverse Adversarial Perturbation (RAP) Algorithm}
  \label{alg}
      {\bf Input:}  Surrogate model $\mathcal{M}^{s}$, benign data $(\bm{x}, y)$, target label $y_t$, loss function $\mathcal{L}$, transformation $\mathcal{G}$, the global iteration number $K$, the late-start iteration number $K_{LS}$ of RAP, as well as hyper-parameters in optimization (specified in later experiments)
    %   \ls{warp the algorithm to save space}\\
      {\bf Output:} the adversarial example $\bm{x}^{adv}$
    \begin{algorithmic}[1]
        \STATE \textbf{Initialize} $\bm{x}^{adv} \leftarrow \bm{x}$
        \FOR{$k=1, \dots, K$}
            \IF{$k \geq K_{LS}$}
                \STATE \textbf{Initialize} $\bm{n}^{rap} \leftarrow \bm{0}$
                \FOR{$t = 1, \dots, T$}
                    \STATE Update $\bm{n}^{rap}$ using \eqref{eq: update n}
                \ENDFOR
            \vskip 0.02in
            \ENDIF
            \STATE Update $\bm{x}^{adv}$ using \eqref{eq: update x}
        \ENDFOR
\end{algorithmic}
\end{algorithm}
\vspace{-0.05in}

\begin{figure}[thp]
    \vspace{-0.05in}
    \centering
    \includegraphics[width=1.0\textwidth]{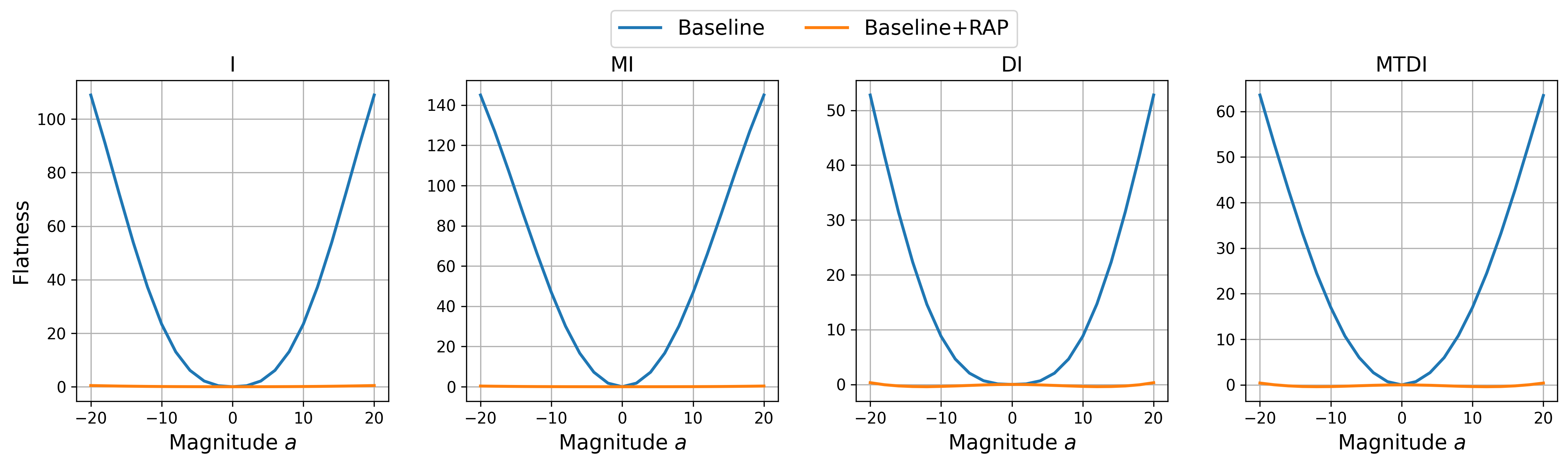}
    \vspace{-0.2in}
    \small{\caption{The flatness visualization of untargeted adversarial examples on $\mathcal{M}^{S}$. The implementation details are shown in Section \ref{implementation details} of \textit{Appendix}.}
    \label{show_flatness}
}
\vspace{-0.1in}
\end{figure}

% \vspace{-0.1in}

\section{Experiments}
\label{experiments}
\vspace{-0.02in}
% \wub{This paragraph should be removed!}
% The evaluation settings are shown in Sec.\ref{evaluation settings}. In Sec.\ref{untarget attacks} and Sec.\ref{targeted attacks}, we conduct the detailed evaluation of our method on targeted and untargeted attacks. In Sec.\ref{comparison}, we take a comparison with other types of iterative attacks and generative attacks. To further show the efficacy of our method, we take the evaluation on more diverse network structures and defense models in Sec.\ref{more models}. In Sec.\ref{ablation study}, we take the ablation study of our method. In Sec.\ref{api}, we evaluate our method against the real world system, Google Cloud Vision API. 

\subsection{Experimental Settings}
\label{evaluation settings}
% \vspace{-0.02in}
\textbf{Dataset and Evaluated Models.}
We conduct the evaluation on the ImageNet-compatible dataset \footnote{Publicly available from \url{https://github.com/cleverhans-lab/cleverhans/tree/master/cleverhans_v3.1.0/examples/nips17_adversarial_competition/dataset}} comprised of
1,000 images. 
% This dataset was used in the NIPS 2017 adversarial competition. 
For the surrogate models, we consider the four widely used network architectures: Inception-v3 (Inc-v3) \citep{szegedy2016rethinking}, ResNet-50 (Res-50) \citep{he2016deep}, DenseNet-121 (Dense-121) \citep{huang2017densely}, and VGG-16bn (VGG-16) \citep{simonyan2014very}. For target models, apart from the above models, we also utilize more diverse architectures: Inception-ResNet-v2 (Inc-Res-v2) \citep{szegedy2017inception}, NASNet-Large (NASNet-L) \citep{zoph2018learning}, and ViT-Base/16 (ViT-B/16) \citep{dosovitskiy2021an}. For defense models, we adopt the two widely used ensemble adversarial training (AT) models: adv-Inc-v3 (Inc-v3$_{adv}$) and ens-adv-Inc-Res-v2 (IncRes-v2$_{ens}$) \citep{tramer2018ensemble}. Besides, we also test multi-step AT model \citep{NEURIPS2020_24357dd0}, imput transformation defense \citep{xie2018mitigating}, feature denoising \citep{xie2019feature}, and purification defense (NRP) \citep{naseer2020self}. 

\textbf{Compared Methods.} 
We adopt I-FGSM \citep{Kurakin_2018} (denoted as I), MI \citep{dong2018boosting}, TI \citep{dong2019evading}, DI \citep{xie2019improving}, SI \citep{Lin2020Nesterov},
Admix \citep{wang2021admix}, VT \citep{Wang_2021_CVPR}, EMI \citep{wang2021boosting}, ILA \citep{huang2019enhancing}, LinBP \citep{GuoLinBP}, Ghost Net \citep{li2020learning}, and the generative targeted attack method TTP \citep{naseer2021generating}. 
We also consider the combination of baseline methods, including MI-TI-DI (MTDI), MI-TI-DI-SI (MTDSI), and MI-TI-DI-Admix (MTDAI). Besides, Expectation of Transformation (EOT) method \citep{athalye2018synthesizing} is also a comparable baseline method. We also conduct the comparison of RAP and EOT.

\textbf{Implementation Details.} For untargeted attack, we adopt the Cross Entropy (CE) loss. For targeted attack, apart from CE, we also experiment with the logit loss, where \citet{zhao2020success} shows it behaves better for targeted attack. The adversarial perturbation $\epsilon$ is restricted by $\ell_{\infty}=16/255$. 
The step size $\alpha$ is set as $2/255$ and number of iteration $K$ is set as $400$ for all attacks.
In the following, we mainly show the results at $K=400$ and the results at different value of $K$ are shown in $\textit{Appendix}$.
% \citet{zhao2020success} shows the targeted attack performance is not sensitive to step size. Following \citet{zhao2020success}, step size $\alpha$ is set as $2/255$. 
% The more iterations can improve the performance of targeted attacks. We set the number of iteration $K$ as $400$ for all attacks. 
For RAP, we set $K_{LS}$ as $100$ and $\alpha_{n}$ as $2/255$. We set $\epsilon_{n}$ as $12/255$ for I and TI in untargeted attack and $16/255$ for other attacks in all other settings. The computational cost is shown in Section \ref{implementation details}  of \textit{Appendix}.

\textbf{Extra Experiments in Appendix.} Due to the space limitation, we put extra experiment results in \textit{Appendix}. The comparisons of RAP and EOT, VT, EMI, and Ghost Net methods are shown in Section \ref{appendix_attack} in \textit{Appendix}. The evaluation of RAP on stronger defense model, multi-step AT models, NRP, and feature denoising, is shown in Section \ref{appendix_defense}. The evaluation of ensemble-model attacks on these diverse network architectures is given in Section \ref{appendix_ensemble_attack}.

\vspace{-0.02in}
\subsection{The Evaluation of Untargeted Attacks}
\label{untarget attacks}
% \vspace{-0.03in}
% We evaluate the untargeted attack performance of the different baseline methods with our method on ResNet-50, DenseNet-121, VGG-16, and Inception-v3.

\textbf{Baseline Methods.} 
We first evaluate the performance of RAP and RAP-LS with different baseline attacks, including I, MI, DI, TI, SI, and Admix.
The results are shown in Table \ref{table_untarget_single}. 
For instance, the `MI/ +RAP/ +RAP-LS' denotes the methods of baseline MI, MI+RAP, and MI+RAP-LS, respectively.
RAP achieves the significant improvements for all methods on each target model. For average attack success rate of all target models, RAP outperforms the I and MI by $9.6\%$ and $16.3\%$, respectively. 
For TI, DI, SI, and Admix, RAP gets the improvements by $10.2\%$, $10.9\%$, $9.3\%$, and $6.3\%$. With late-start, RAP-LS further enhance the transfer attack performance for almost all methods.

\textbf{Combinational Methods.} 
Prior works demonstrate the combination of baseline methods could largely boost the adversarial transferability \citep{zhao2020success,wang2021admix}. 
We also investigate of behavior of RAP when incorporated with the combinational attacks.
The results are shown in Table \ref{table_untarget_combination}. 
As shown in the table, there exist the clear improvements of the combinational attacks over all baseline attacks shown in Table \ref{table_untarget_single}.
% Although three combinational methods have shown the good performance for untargeted attacks, RAP still gets the further improvements on all target models.
In addition, our RAP-LS further boosts the average attack success rate of the three combinational attacks by $6.9\%$, $2.6\%$, and $1.7\%$ respectively.
Combined with the three combinational attacks, RAP-LS achieves $95.4\%$, $97.6\%$, and $98.3\%$ average attack success rate, respectively. 
% While the RAP-LS obtains the best transfer attack success rate under most cases. 
These results demonstrate RAP can significantly enhance the transferability. 

% \begin{minipage}[t]
\vspace{-0.03in}
\begin{table}[t]
\small{\caption{
The \textbf{untargeted attack success rate ($\%$) of baseline attacks with RAP}. The results with $CE$ loss are reported. The best results are bold and the second best results are underlined.}
\label{table_untarget_single}
}
\vspace{-0.03in}
\begin{center}
\begin{small}
% \begin{sc}
\scalebox{0.67}{
\begin{tabular}{c|ccc|ccc}
\toprule
\multirow{2}{*}{\tabincell{c}{Attack}}  &  & \textbf{ResNet-50} $\Longrightarrow$ &  &  & \textbf{DenseNet-121}$\Longrightarrow$ & \\
& Dense-121 & VGG-16 & Inc-v3    & Res-50 & VGG-16 & Inc-v3  \\
\midrule 
I / +RAP / +RAP-LS 
% e = 16/255, k = 8
% & 79.2 / 81.8 / 79.5
% & 78.0 / 79.9 / 81.1
% & 34.6 / 45.9 / 46.2

% e = 12/255, k = 6
& 79.2 / \underline{91.5} / \textbf{91.9}
& 78.0 / \underline{91.1} / \textbf{92.9}
& 34.6 / \underline{57.0} / \textbf{57.2}

% e = 16/255, k = 8
% & 87.1 / 86.1 / 86.2
% & 85.1 / 84.6 / 84.6
% & 46.5 / 52.4 / 54.2

% e = 12/255, k = 6
& 87.4 / \underline{94.2} / \textbf{94.3}
& 85.1 / \underline{91.7} / \textbf{92.8}
& 46.5 / \underline{60.2} / \textbf{61.1}

\\ MI / +RAP / +RAP-LS
& 85.8 / \underline{95.0} / \textbf{96.1}
& 82.4 / \underline{93.9} / \textbf{94.5}
& 50.3 / \underline{75.9} / \textbf{77.4}
& 90.3 / \underline{97.6} / \textbf{97.9}
& 87.5 / \underline{96.0} / \textbf{97.6}
& 59.3 / \underline{80.4} / \textbf{82.8}

\\ TI / +RAP / +RAP-LS
% e = 16/255, k = 8
% & 82.0 / 82.9 / 81.6
% & 81.0 / 82.7 / 80.6
% & 45.5 / 54.3 / 51.9

% e = 12/255, k = 6
& 82.0 / \underline{94.1} / \textbf{95.1}
& 81.0 / \underline{93.1} / \textbf{93.3}
& 45.5 / \underline{66.1} / \textbf{67.0}

% e = 16/255, k = 8
% & 89.6 / 87.7 / 87.5
% & 87.0 / 85.3 / 85.9
% & 54.2 / 60.4 / 59.5

% e = 12/255, k = 6
& 89.6 / \underline{94.2} / \textbf{94.8}
& 87.0 / \underline{92.1} / \textbf{93.3}
& 54.2 / \underline{66.7} / \textbf{70.0}

\\ DI / +RAP / +RAP-LS
& 99.0 / \underline{99.6} / \textbf{99.7}
& 99.0 / \underline{99.6} / \textbf{99.7}
& 57.7 / \underline{82.9} / \textbf{85.0}
& 98.2 / \underline{99.6} / \textbf{99.7}
& 98.1 / \textbf{99.4} / \textbf{99.4}
& 67.6 / \underline{86.6} / \textbf{86.9}

\\ SI / +RAP / +RAP-LS
& 94.9 / \underline{98.9} / \textbf{99.7}
& 88.6 / \underline{95.7} / \textbf{97.2}
& 65.9 / \underline{79.7} / \textbf{84.4}
& 95.1 / \underline{96.9} / \textbf{98.8}
& 91.9 / \underline{95.0} / \textbf{97.5}
& 71.6 / \underline{83.2} / \textbf{87.4}

\\ Admix / +RAP / +RAP-LS
& 97.9 / \underline{99.6} / \textbf{99.9}
& 95.8 / \underline{97.7} / \textbf{99.0}
& 77.7 / \underline{87.4} / \textbf{92.6}
& 97.0 / \underline{99.0} / \textbf{99.2}
& 95.6 / \underline{97.7} / \textbf{98.6}
& 82.0 / \underline{89.8} / \textbf{93.8}

\\
% MI-DI-FGSM   \\
% MI-TI-DI-FGSM     \\
% MI-TI-DI-SI-FGSM    \\
% MI-TI-DI-Admix    \\
\midrule
\multirow{2}{*}{\tabincell{c}{Attack}}  &  & \textbf{VGG-16} $\Longrightarrow$ &  &  & \textbf{Inc-v3}$\Longrightarrow$ & \\  & Res-50 & Dense-121 & Inc-v3 & Res-50 & Dense-121 & VGG-16 \\
\midrule
I / +RAP / +RAP-LS
% e = 16/255, k = 8
% & 53.7 / 44.5 / 43.4
% & 49.1 / 41.8 / 40.6
% & 22.0 / 19.9 / 20.7

% e = 12/255, k = 6
& \underline{53.7} / 53.0 / \textbf{54.2}
& 49.1 / \underline{50.6} / \textbf{51.4}
& 22.0 / \underline{24.7} / \textbf{24.9}

% e = 16/255, k = 8
% & 51.5 / 55.3 / 55.8
% & 48.7 / 54.8 / 54.8
% & 55.1 / 60.4 / 60.9

% e = 12/255, k = 6
& 51.5 / \textbf{62.1} / \underline{62.0}
& 48.7 / \textbf{60.8} / \underline{60.0}
& 55.1 / \underline{65.9} / \textbf{68.0}

\\ MI / +RAP / +RAP-LS
& 62.5 / \underline{76.2} / \textbf{76.4}
& 60.5 / \underline{73.0} / \textbf{73.9}
& 30.0 / \textbf{42.7} / \underline{42.2}
& 62.0 / \textbf{85.8} / \underline{84.8}
& 56.7 / \textbf{84.6} / \textbf{84.6}
& 63.1 / \textbf{84.9} / \underline{84.6}

\\ TI / +RAP / +RAP-LS
% e = 16/255, k = 8
% & 62.8 / 52.9 / 52.1
% & 55.9 / 49.1 / 50.1
% & 29.1 / 31.9 / 31.5

% e = 12/255, k = 6
& 62.8 / \underline{64.8} / \textbf{65.8}
& 55.9 / \textbf{63.7} / \underline{62.1}
& 29.1 / \underline{36.2} / \textbf{37.1}

% e = 16/255, k = 8
% & 49.3 / 55.3 / 55.1
% & 49.4 / 58.5 / 56.9
% & 58.1 / 63.6 / 63.0

% e = 12/255, k = 6
& 49.3 / \textbf{63.4} / \underline{61.6}
& 49.4 / \underline{63.4} / \textbf{63.8}
& 58.1 / \underline{68.6} / \textbf{69.5}

\\ DI / +RAP / +RAP-LS
& 72.2 / \underline{86.0} / \textbf{88.8}
& 68.8 / \underline{85.0} / \textbf{87.4}
& 29.9 / \underline{46.6} / \textbf{51.6}
& 68.4 / \underline{81.7} / \textbf{81.8}
& 71.9 / \textbf{85.0} / \underline{84.0}
& 76.1 / \underline{85.2} / \textbf{86.4}

\\ SI / +RAP / +RAP-LS
& 80.0 / \underline{92.7} / \textbf{94.7}
& 82.1 / \underline{94.8} / \textbf{95.7}
& 45.8 / \underline{74.0} / \textbf{74.7}
& 66.2 / \underline{69.8} / \textbf{72.8}
& 65.9 / \underline{74.9} / \textbf{77.2}
& 66.0 / \underline{69.2} / \textbf{73.0}

\\ Admix / +RAP / +RAP-LS
& 87.3 / \underline{94.6} / \textbf{96.8}
& 88.2 / \underline{96.4} / \textbf{97.2}
& 55.5 / \underline{77.6} / \textbf{80.8}
& 75.9 / \underline{80.2} / \textbf{84.9}
& 78.5 / \underline{83.7} / \textbf{87.4}
& 74.5 / \underline{77.2} / \textbf{83.5}

\\
% MI-DI-FGSM    \\
% MI-TI-DI-FGSM   \\
% MI-TI-DI-SI-FGSM     \\
% MI-TI-DI-Admix   \\
\bottomrule
\end{tabular}
}
% \end{sc}
\end{small}
\end{center}
\vspace{-0.12in}
\end{table}
% CE Loss
% \end{minipage}

\vspace{-0.02in}
\begin{table}[t]
\small{\caption{
The \textbf{untargeted attack success rate ($\%$) of combinational methods with RAP}. The results with $CE$ loss are reported. The best results are bold and the second best results are underlined.} 
\label{table_untarget_combination}
}
\vspace{-0.03in}
\begin{center}
\begin{small}
% \begin{sc}
\scalebox{0.67}{
\begin{tabular}{c|ccc|ccc}
\toprule
\multirow{2}{*}{\tabincell{c}{Attack}}  &  & \textbf{ResNet-50} $\Longrightarrow$ &  &  & \textbf{DenseNet-121}$\Longrightarrow$ & \\
& Dense-121 & VGG-16 & Inc-v3    & Res-50 & VGG-16 & Inc-v3  \\
\midrule 
MTDI / +RAP / +RAP-LS
& 99.8 / \textbf{100} / \textbf{100}
& 99.8 / \textbf{100} / \underline{99.9}
& 85.7 / \underline{96.0} / \textbf{96.9}
& 99.4 / \underline{99.8} / \textbf{100}
& 99.2 / \underline{99.5} / \textbf{100}
& 89.1 / \textbf{97.1} / \textbf{97.1}

% \\ +RAP / +RAP-LS
% & \textbf{100} / \textbf{100}
% & \textbf{100} / 99.9
% & 96.0 / \textbf{96.9}
% & 99.8 / \textbf{100}
% & 99.5 / \textbf{100}
% & \textbf{97.1} / \textbf{97.1}

\\
MTDSI / +RAP / +RAP-LS
& \textbf{100} / \textbf{100} / \textbf{100}
& 99.7 / \textbf{99.9} / \underline{99.8}
& 97.0 / \textbf{99.1} / \textbf{99.1}
& 99.8 / \textbf{99.9} / \textbf{99.9}
& 99.2 / \underline{99.3} / \textbf{99.7}
& 95.1 / \underline{98.3} / \textbf{98.4}

% \\ +RAP / +RAP-LS
% & \textbf{100} / \textbf{100}
% & \textbf{99.9} / 99.8
% & \textbf{99.1} / \textbf{99.1}
% & \textbf{99.9} / \textbf{99.9}
% & 99.3 / \textbf{99.7}
% & 98.3 / \textbf{98.4}

\\
MTDAI / +RAP / +RAP-LS
& \textbf{100} / \textbf{100} / \textbf{100}
& 99.8 / \textbf{99.9} / \textbf{99.9}
& 98.3 / \underline{99.2} / \textbf{99.8}
& \underline{99.8} / \underline{99.8} / \textbf{99.9}
& 99.4 / \underline{99.6} / \textbf{99.8}
& 97.9 / \underline{98.8} / \textbf{98.9}

% \\ +RAP / +RAP-LS
% & \textbf{100} / \textbf{100}
% & \textbf{99.9} / \textbf{99.9}
% & 99.2 / \textbf{99.8}
% & 99.8 / \textbf{99.9}
% & 99.6 / \textbf{99.8}
% & 98.8 / \textbf{98.9}

\\

\midrule
\multirow{2}{*}{\tabincell{c}{Attack}}  &  & \textbf{VGG-16} $\Longrightarrow$ &  &  & \textbf{Inc-v3}$\Longrightarrow$ & \\  & Res-50 & Dense-121 & Inc-v3 & Res-50 & Dense-121 & VGG-16 \\
\midrule 
MTDI / +RAP / +RAP-LS
& 90.0 / \underline{97.2} / \textbf{97.7}
& 88.8 / \underline{97.0} / \textbf{97.3}
& 56.8 / \textbf{82.6} / \underline{81.4}
& 82.9 / \textbf{91.8} / \underline{90.6}
& 85.7 / \textbf{94.2} / \underline{93.3}
& 85.1 / \textbf{92.7} / \underline{91.0}

% \\ +RAP / +RAP-LS
% & 97.2 / \textbf{97.7}
% & 97.0 / \textbf{97.3}
% & \textbf{82.6} / 81.4
% & \textbf{91.8} / 90.6
% & \textbf{94.2} / 93.3
% & \textbf{92.7} / 91.0

\\
MTDSI / +RAP / +RAP-LS
& 97.6 / \underline{98.8} / \textbf{99.4}
& 98.1 / \underline{99.2} / \textbf{99.4}
& 85.0 / \underline{94.1} / \textbf{95.2}
& 89.0 / \underline{91.2} / \textbf{92.3}
& 92.0 / \underline{95.2} / \textbf{95.6}
& 87.6 / \underline{90.3} / \textbf{92.2}

% \\ +RAP / +RAP-LS
% & 98.8 / \textbf{99.4}
% & 99.2 / \textbf{99.4}
% & 94.1 / \textbf{95.2}
% & 91.2 / \textbf{92.3}
% & 95.2 / \textbf{95.6}
% & 90.3 / \textbf{92.2}

\\
MTDAI / +RAP / +RAP-LS
& 97.8 / \underline{99.2} / \textbf{99.6}
& 98.9 / \underline{99.5} / \textbf{99.6}
& 89.3 / \underline{95.0} / \textbf{95.5}
& 91.5 / \underline{94.1} / \textbf{94.7}
& 95.4 / \underline{96.2} / \textbf{97.6}
& 91.4 / \underline{93.2} / \textbf{94.1}

% \\ +RAP / +RAP-LS
% & 99.2 / \textbf{99.6}
% & 99.5 / \textbf{99.6}
% & 95.0 / \textbf{95.5}
% & 94.1 / \textbf{94.7}
% & 96.2 / \textbf{97.6}
% & 93.2 / \textbf{94.1}
\\

\bottomrule
\end{tabular}
}
% \end{sc}
\end{small}
\end{center}
\vskip -0.1in
\end{table}
% CE Loss

% \vspace{-0.05in}
\subsection{The Evaluation of Targeted Attacks}
\label{targeted attacks}
\vspace{-0.02in}
% As shown in the previous work \citep{zhao2020success, naseer2021generating},
% In this secion, we evaluate the targeted attack performance of the different baseline methods with RAP. We mainly show the results with using Logit loss in main submission. The results with using CE loss are shown in Appendix. 
We then evaluate the targeted attack performance of the different methods with RAP. The results with logit loss are presented and the results with CE loss are shown in {\it Appendix}. 

\textbf{Baseline Methods.}
The results of RAP with baseline attacks are shown in Table \ref{table_target_single}. 
From the results, RAP is also very effective in enhancing the transferability in targeted attacks.
% It significantly improves the attack performance of all methods on each target model.
Taking ResNet-50 and DenseNet-121 as surrogate models for example, the average performance improvements induced by RAP are $5.0\%$ (I), $8.1\%$ (MI), $4.6\%$ (TI), $10.4\%$ (DI), $18.5\%$ (SI), and $15.1\%$ (Admix), respectively.
Comparing to the ResNet-50 and DenseNet-121, the baseline attacks generally achieve lower transferability when using the VGG-16 or Inception-v3 as the surrogate models, which has also been verified in existing works \citep{xie2019improving,zhao2020success}. 
However, for Inception-v3 and VGG-16 as the surrogate models, RAP also consistently boosts the transferability under all cases.  
With late-start, RAP-LS could further improve the transferability of RAP for most attacks.
The average attack success rate under all attack cases of RAP-LS is $2.6\%$ higher than that of RAP.

\textbf{Combinational Methods.} 
As did in the untargeted attacks, we also evaluate the performance of combinational methods.
The results are shown in Table \ref{table_target_combination}.
Similar to the findings in untargeted attacks, the combinational methods obtain significantly improvements over baseline methods.
The RAP-LS outperforms all combinational methods by a significantly margin.
For example, taking the average attack success rate of all target models as evaluation metric, RAP-LS obtains $14.2\%$, $11.8\%$, $9.3\%$ improvements over the MTDI, MTDSI and MTDAI, respectively.

% For all combined baseline methods, we significantly improves the targeted attack performance on all models. For ResNet and DenseNet as surrogate models, compared with \citet{zhao2020success}, our method and - w get further boosts by $13.8\%$ and $21.8\%$ on average success rate. Especially for Inception as target model, our methods get more significant improvements by $16.4\%$ and $20.6\%$. Since Inception model is much more hard to attack, these significant boost sufficiently verify efficacy of our methods. As shown in Table \ref{table_target_combination}, combining MI-DI-TI with other augmentations also improves transferability. And, our method also further enhance the attack performance of these new combinations. MI-TI-DI-Admix with -W achieves the best targeted attack performance for all used models. When transferring from ResNet to DenseNet and VGG, its targeted attack performance is even close to that of untargeted attack, getting $93.6\%$ and $86.3\%$ success rate. 

\begin{table}[t]
\small{\caption{
The \textbf{targeted attack success rate ($\%$) of baseline methods with RAP}. The results with logit loss are reported. The best results are bold and the second best results are underlined.}
\label{table_target_single}
}
\vspace{-0.03in}
\begin{center}
\begin{small}
% \begin{sc}
\scalebox{0.67}{
\begin{tabular}{c|ccc|ccc}
\toprule
\multirow{2}{*}{\tabincell{c}{Attack}}  &  & \textbf{ResNet-50} $\Longrightarrow$ &  &  & \textbf{DenseNet-121}$\Longrightarrow$ & \\
& Dense-121 & VGG-16 & Inc-v3    & Res-50 & VGG-16 & Inc-v3  \\
\midrule 
I / +RAP / +RAP-LS
& 4.5 / \underline{9.5} / \textbf{14.3}
& 2.4 / \underline{9.8} / \textbf{11.8}
& \underline{0.1} / \underline{0.1} / \textbf{0.7}
& 5.0 / \underline{12.8} / \textbf{17.9}
& 2.9 / \underline{10.1} / \textbf{15.9}
& 0.0 / \underline{0.8} / \textbf{1.2}

\\MI / +RAP / +RAP-LS
& 6.3 / \underline{17.5} / \textbf{29.6}
& 2.2 / \underline{14.5} / \textbf{20.6}
& 0.1 / \underline{1.1} / \textbf{2.4}
& 4.6 / \underline{16.2} / \textbf{26.5}
& 3.1 / \underline{13.4} / \textbf{23.2}
& 0.3 / \underline{2.0} / \textbf{3.4}

\\ TI / +RAP / +RAP-LS
& 7.2 / \underline{11.0} / \textbf{17.3}
& 4.0 / \underline{12.9} / \textbf{15.3}
& 0.1 / \underline{0.8} / \textbf{1.2}
& 8.4 / \underline{13.5} / \textbf{20.8}
& 5.2 / \underline{12.4} / \textbf{16.4}
& 0.2 / \underline{2.1} / \textbf{3.0}

\\DI / +RAP / +RAP-LS
& 62.6 / \underline{64.9} / \textbf{73.9}
& 57.2 / \underline{63.4} / \textbf{69.3}
& 1.5 / \underline{7.9} / \textbf{10.1}
& 30.2 / \underline{52.6} / \textbf{60.4}
& 32.1 / \underline{49.5} / \textbf{58.9}
& 1.4 / \underline{8.8} / \textbf{10.0}

\\SI / +RAP / +RAP-LS
& 30.0 / \underline{53.2} / \textbf{61.1}
& 9.5 / \underline{32.8} / \textbf{36.0}
& 1.8 / \underline{9.3} / \textbf{10.5}
& 14.2 / \underline{41.5} / \textbf{43.4}
& 8.4 / \underline{31.0} / \textbf{35.2}
& 1.6 / \underline{8.5} / \textbf{10.4}

\\Admix / +RAP / +RAP-LS
& 54.6 / \underline{68.0} / \textbf{74.6}
& 26.0 / \underline{45.4} / \textbf{51.6}
& 5.8 / \underline{17.1} / \textbf{19.6}
& 29.3 / \underline{53.0} / \textbf{58.2}
& 21.5 / \underline{42.7} / \textbf{48.2}
& 5.0 / \underline{17.1} / \textbf{17.6}

\\
% MI-DI-FGSM   \\
% MI-TI-DI-FGSM     \\
% MI-TI-DI-SI-FGSM    \\
% MI-TI-DI-Admix    \\
\midrule
\multirow{2}{*}{\tabincell{c}{Attack}}  &  & \textbf{VGG-16} $\Longrightarrow$ &  &  & \textbf{Inc-v3}$\Longrightarrow$ & \\  & Res-50 & Dense-121 & Inc-v3 & Res-50 & Dense-121 & VGG-16 \\
\midrule
I / +RAP / +RAP-LS
& 0.1 / \underline{0.7} / \textbf{1.4}
& 0.2 / \underline{1.4} / \textbf{1.7}
& 0.0 / \underline{0.1} / \textbf{0.2}
& 0.2 / \textbf{0.9} / \underline{0.5}
& 0.2 / \textbf{0.6} / \underline{0.3}
& 0.1 / \textbf{0.5} / \textbf{0.5}

\\ MI / +RAP / +RAP-LS
& 0.5 / \underline{1.3} / \textbf{1.9}
& 0.5 / \underline{2.3} / \textbf{3.0}
& \underline{0.0} / \underline{0.0} / \textbf{0.3}
& 0.2 / \textbf{1.7} / \underline{1.5}
& 0.1 / \textbf{1.6} / \underline{1.5}
& 0.2 / \textbf{1.3} / \underline{1.0}

\\ TI / +RAP / +RAP-LS
& 0.7 / \underline{1.2} / \textbf{3.2}
& 0.8 / \underline{1.7} / \textbf{2.9}
& 0.0 / \underline{0.1} / \textbf{0.4}
& 0.2 / \underline{0.5} / \textbf{0.7}
& 0.1 / \textbf{0.7} / \underline{0.6}
& 0.2 / \textbf{0.8} / \underline{0.6}

\\ DI / +RAP / +RAP-LS
& 2.8 / \underline{7.3} / \textbf{9.7}
& 3.8 / \underline{8.4} / \textbf{12.7}
& 0.0 / \underline{0.4} / \textbf{1.1}
& 1.6 / \underline{4.6} / \textbf{6.4}
& 2.8 / \underline{5.8} / \textbf{7.5}
& 2.6 / \underline{6.3} / \textbf{8.1}

\\ SI / +RAP / +RAP-LS
& 3.3 / \textbf{9.8} / \textbf{9.8}
& 7.2 / \underline{16.8} / \textbf{17.8}
& 0.2 / \underline{1.7} / \textbf{1.8}
& 0.6 / \textbf{2.9} / \underline{2.5}
& 0.9 / \underline{2.7} / \textbf{3.2}
& 0.5 / \underline{1.5} / \textbf{2.3}

\\ Admix / +RAP / +RAP-LS
& 5.6 / \underline{11.1} / \textbf{11.9}
& 13.0 / \underline{20.2} / \textbf{23.6}
& 0.7 / \underline{2.4} / \textbf{2.8}
& 1.5 / \underline{4.9} / \textbf{5.2}
& 2.0 / \underline{6.9} / \textbf{7.5}
& 1.3 / \underline{3.3} / \textbf{4.4}

\\
% MI-DI-FGSM    \\
% MI-TI-DI-FGSM   \\
% MI-TI-DI-SI-FGSM     \\
% MI-TI-DI-Admix   \\
\bottomrule
\end{tabular}
}
% \end{sc}
\end{small}
\end{center}
\vskip -0.12in
\end{table}
% Loss Function: MaxLogit
% Checked

\begin{table}[t]
\small{\caption{
The \textbf{targeted attack success rate ($\%$) of combinational methods with RAP}. The results with logit loss are reported. The best results are bold and the second best results are underlined.} 
\label{table_target_combination}
}
\vspace{-0.03in}
\begin{center}
\begin{small}
% \begin{sc}
\scalebox{0.67}{
\begin{tabular}{c|ccc|ccc}
\toprule
\multirow{2}{*}{\tabincell{c}{Attack}}  &  & \textbf{ResNet-50} $\Longrightarrow$ &  &  & \textbf{DenseNet-121}$\Longrightarrow$ & \\
& Dense-121 & VGG-16 & Inc-v3    & Res-50 & VGG-16 & Inc-v3  \\
\midrule 
% MI-TI-DI \citep{zhao2020success}
MTDI / +RAP / +RAP-LS
& 74.9 / \underline{78.2} / \textbf{88.5}
& 62.8 / \underline{72.9} / \textbf{81.5}
& 10.9 / \underline{28.3} / \textbf{33.2}
& 44.9 / \underline{64.3} / \textbf{74.5}
& 38.5 / \underline{55.0} / \textbf{65.5}
& 7.7 / \underline{23.0} / \textbf{26.5}

% \\ +RAP / +RAP-LS 
% & 78.2 / \textbf{88.5}
% & 72.9 / \textbf{81.5}
% & 28.3 / \textbf{33.2}
% & 64.3 / \textbf{74.5}
% & 55.0 / \textbf{65.5}
% & 23.0 / \textbf{26.5}

\\
% MI-TI-DI-SI 
MTDSI / +RAP / +RAP-LS
& 86.3 / \underline{88.4} / \textbf{93.3}
& 70.1 / \underline{77.7} / \textbf{84.7}
& 38.1 / \underline{51.8} / \textbf{58.0}
& 55.0 / \underline{71.2} / \textbf{75.8}
& 42.0 / \underline{58.4} / \textbf{62.3}
& 19.8 / \underline{39.0} / \textbf{39.2}

% \\ +RAP / +RAP-LS 
% & 88.4 / \textbf{93.3}
% & 77.7 / \textbf{84.7}
% & 51.8 / \textbf{58.0}
% & 71.2 / \textbf{75.8}
% & 58.4 / \textbf{62.3}
% & 39.0 / \textbf{39.2}

\\
% MI-TI-DI-Admix 
MTDAI / +RAP / +RAP-LS
& \underline{91.4} / 89.4 / \textbf{93.6}
& \underline{79.9} / 79.0 / \textbf{86.3}
& 50.8 / \underline{57.1} / \textbf{64.1}
& 69.1 / \underline{74.2} / \textbf{82.1}
& 54.7 / \underline{63.1} / \textbf{69.3}
& 32.0 / \underline{43.5} / \textbf{49.3}

% \\ +RAP / +RAP-LS
% & 89.4 / \textbf{93.6}
% & 79.0 / \textbf{86.3}
% & 57.1 / \textbf{64.1}
% & 74.2 / \textbf{82.1}
% & 63.1 / \textbf{69.3}
% & 43.5 / \textbf{49.3}

\\

\midrule
\multirow{2}{*}{\tabincell{c}{Attack}}  &  & \textbf{VGG-16} $\Longrightarrow$ &  &  & \textbf{Inc-v3}$\Longrightarrow$ & \\  & Res-50 & Dense-121 & Inc-v3 & Res-50 & Dense-121 & VGG-16 \\
\midrule 
% MI-TI-DI \citep{zhao2020success}
MTDI / +RAP / +RAP-LS
& 11.8 / \underline{16.7} / \textbf{22.9}
& 13.7 / \underline{19.4} / \textbf{27.4}
& 0.7 / \underline{3.4} / \textbf{4.6}
& 1.8 / \textbf{8.3} / \underline{7.5}
& 4.1 / \textbf{14.8} / \underline{13.4}
& 2.9 / \underline{8.0} / \textbf{9.8}

% \\ +RAP / +RAP-LS  
% & 16.7 / \textbf{22.9}
% & 19.4 / \textbf{27.4}
% & 3.4 / \textbf{4.6}
% & \textbf{8.3} / 7.5
% & \textbf{14.8} / 13.4
% & 8.0 / \textbf{9.8}

\\
% MI-TI-DI-SI 
MTDSI / +RAP / +RAP-LS
& 31.0 / \underline{35.3} / \textbf{38.7}
& 41.7 / \underline{44.4} / \textbf{49.6}
& 9.6 / \textbf{15.2} / \underline{13.7}
& 5.6 / \textbf{11.9} / \underline{10.7}
& 10.4 / \textbf{21.2} / \underline{20.9}
& 4.2 / \textbf{8.9} / \underline{8.6}

% \\ +RAP / +RAP-LS
% & 35.3 / \textbf{38.7}
% & 44.4 / \textbf{49.6}
% & \textbf{15.2} / 13.7
% & \textbf{11.9} / 10.7
% & \textbf{21.2} / 20.9
% & \textbf{8.9} / 8.6

\\
% MI-TI-DI-Admix 
MTDAI / +RAP / +RAP-LS
& 36.2 / \underline{39.0} / \textbf{43.1}
& \underline{48.0} / 45.1 / \textbf{55.2}
& 11.6 / \underline{17.1} / \textbf{17.6}
& 9.6 / \underline{13.6} / \textbf{16.7}
& 17.9 / \underline{27.5} / \textbf{31.6}
& 8.4 / \underline{12.0} / \textbf{12.1}

% \\ +RAP / +RAP-LS
% & 39.0 / \textbf{43.1}
% & 45.1 / \textbf{55.2}
% & 17.1 / \textbf{17.6}
% & 13.6 / \textbf{16.7}
% & 27.5 / \textbf{31.6}
% & 12.0 / \textbf{12.1}

\\

\bottomrule
\end{tabular}
}
% \end{sc}
\end{small}
\end{center}
\vskip -0.12in
\end{table}

\vspace{-0.05in}
\subsection{The Comparison with Other Types of Attacks} 
\label{comparison}
% \vspace{-0.05in}
 \begin{wraptable}{r}{0.56\textwidth}
\vspace{-0.05in}
\centering
\small{\caption{
{The comparison with ILA and LinBP}. We use ResNet-50 as $\mathcal{M}^{S}$. The best results are bold.}
\label{table_comparison}
}
\vspace{-0.1in}
\scalebox{0.63}{
\begin{tabular}{c|ccc|ccc}
\toprule
\multirow{2}{*}{\tabincell{c}{Attack}}  &  & \textbf{Untarged} &  &  & \textbf{Targeted} & \\
& Dense-121 & VGG-16 & Inc-v3    & Dense-121 & VGG-16 & Inc-v3  \\
\midrule 
ILA 
& 95.0
& 94.2
& 77.7
& 2.8
& 1.5
& 0.5
\\ LinBP-ILA
& 99.5
& 99.2
& 89.8
& 9.4
& 4.9
& 2.0
\\ LinBP-ILA-SGM 
& 99.7
& 99.3
& 91.1
& 13.3
& 7.2
& 2.8
\\ LinBP-MI-DI 
& 99.5
& 99.2
& 89.3
& 26.1
& 16.5
& 3.2
\\ LinBP-MI-DI-SGM 
& 99.8
& 99.3
& 90.2
& 32.6
& 22.1
& 4.6
% \\ MI-TI-DI+RAP
% & 100
% & 100
% & 96.0
% & 78.3
% & 70.5
% & 21.3
% \\ MI-DI
% & 99.7
% & 99.8
% & 81.3
% & 41.3
% & 32.0
% & 2.9
\\ MI-DI+RAP
& \textbf{99.9}
& \textbf{100}
& \textbf{93.7}
& \textbf{75.1}
& \textbf{69.7}
& \textbf{13.9}
\\

\bottomrule
\end{tabular} }
\vspace{-0.1in}
\end{wraptable}
Apart from the baseline and the combinational methods, we also experiment with more diverse attack methods, including the model-specific attack LinBP \citep{GuoLinBP}, the feature-based attack ILA \citep{huang2019enhancing}, and the generative targeted attack TTP \citep{naseer2021generating}. 
The LinBP depends on the skip connection and the authors only provide the source code about ResNet-50. 
We use their released code and thus conduct experiments
with ResNet-50 as $\mathcal{M}^{S}$. 
The results of LibBP and ILA are shown in Table \ref{table_comparison}, where we also implement the variants of LinBP following \citet{GuoLinBP}, inlcuding LinBP-ILA, LinBP-ILA-SGM, LinBP-MI-DI, and LinBP-MI-DI-SGM.  
We observe that our MI-DI-RAP significantly outperforms the LinBP and ILA, especially for the targeted attacks. 
Compared with the second-best method (\ie, LinBP-MI-DI-SGM), we obtain a large improvement by $33.5\%$ on average ASR of targeted attacks. 

\begin{wraptable}{r}{0.4\textwidth}
\centering
\vspace{-0.05in}
\caption{
{The comparison with TTP on targeted attack.} The best results are bold.}
\label{table_TTP}
\vspace{-0.1in}
\scalebox{0.67}{
\begin{tabular}{c|ccc}
\toprule
Attack & Dense-121 & VGG-16 & Inc-v3\\
\midrule 
TTP 
& 79.6
& 78.6
& 40.3
\\ MTDI
& 78.6
& 74.6
& 12.7
\\ MTDI+RAP-LS
& 90.8
& 87.2
& 35.4
\\ MTDSI
& 93.2
& 80.0
& 41.3
\\ MTDSI+RAP-LS
& \textbf{95.7}
& \textbf{88.1}
& \textbf{59.3}
\\
\bottomrule
\end{tabular}
}
% \vspace{-0.1in}
\end{wraptable}
TTP \citep{naseer2021generating} is the state-of-the-art generative method to conduct targeted attack. 
To compare with it, we adopt the generators based on ResNet-50 provided by the authors. Since TTP needs to train the perturbation generator for each targeted class, we follow their ``10-Targets (all-source)" setting, as did in \citet{zhao2020success}. 
The results are shown in Table \ref{table_TTP}, where our MTDSI+RAP-LS behaves best and outperforms TTP and MTDI by large margins of $14.9\%$ and $25.7\%$, respectively.

\begin{table}[t]
\small{\caption{
The evaluation on diverse network architectures and defense models.}
%To attack models with diverse architectures, we use ResNet-50 as the surrogate model. To attack defense models, we use the ensemble-model as the surrogate model.
\label{table_ensemble}
}
\vspace{-0.02in}
\begin{center}
\begin{small}
% \begin{sc}
\scalebox{0.68}{
\begin{tabular}{c|ccc|ccc|cc|cc}
\toprule
\multirow{2}{*}{\tabincell{c}{Attack}} & \multicolumn{3}{c|}{\textbf{Untarged}} & \multicolumn{3}{c|}{\textbf{Targeted}} & \multicolumn{2}{c|}{\textbf{Untarged}} & \multicolumn{2}{c}{\textbf{Targeted}} \\
& IncRes-v2 & NASNet-L & ViT-B/16 & IncRes-v2 & NASNet-L & ViT-B/16 & Inc-v3$_{adv}$ &  IncRes-v2$_{ens}$ & Inc-v3$_{adv}$ &  IncRes-v2$_{ens}$ \\
\midrule 
MTDI 
% % Ensemble
% & 98.6
% & 99.3
% & 46.2
% & 65.7
% & 80.1
% & 2.8

% ResNet-50
& 83.4
& 89.0
& 27.9
& 14.8
& 32.1
& 0.4

& 68.1
& 50.9
& 0.8
& 0.0
\\ MTDI+RAP-LS
% % Ensemble
% & 100
% & 100
% & 73.2
% & 84.4
% & 89.7
% & 12.7

% ResNet-50
& \textbf{95.6}
& \textbf{97.5}
& \textbf{42.7}
& \textbf{43.0}
& \textbf{62.5}
& \textbf{1.7}

& \textbf{86.5}
& \textbf{72.3}
& \textbf{9.7}
& \textbf{4.1}
\\ MTDSI 
% % Ensemble
% & 99.8
% & 100
% & 68.3
% & 81.7
% & 89.4
% & 15.0

% ResNet-50
& 95.7
& 98.0
& 43.0
& 45.5
& 67.9
& 2.6

& 90.0
& 79.6
& 12.7
& 6.7
\\ MTDSI+RAP-LS
% % Ensemble
% & 100
% & 100
% & 85.0
% & 89.8
% & 92.3
% & 25.1

% ResNet-50
& \textbf{98.6}
& \textbf{99.7}
& \textbf{57.4}
& \textbf{64.0}
& \textbf{80.4}
& \textbf{5.3}

& \textbf{96.5}
& \textbf{91.5}
& \textbf{31.0}
& \textbf{22.0}
\\ MTDAI
% % Ensemble
% & 100
% & 100
% & 70.7
% & 88.8
% & 91.2
% & 16.8

% ResNet-50
& 97.3
& 98.8
& 45.5
& 58.4
& 75.3
& 3.3

& 92.1
& 82.7
& 17.2
& 12.2
\\ MTDAI+RAP-LS
% % Ensemble
% & 100
% & 100
% & 84.6
% & 90.4
% & 91.8
% & 27.8

% ResNet-50
& \textbf{99.2}
& \textbf{99.8}
& \textbf{60.2}
& \textbf{70.4}
& \textbf{82.6}
& \textbf{7.4}

& \textbf{96.7}
& \textbf{91.6}
& \textbf{34.4}
& \textbf{26.0}
\\

\bottomrule
\end{tabular}
}
% \end{sc}
\end{small}
\end{center}
\vskip -0.1in
\end{table}

% \vspace{-0.05in}
\subsection{The Evaluation on Diverse Network Architectures and Defense Models} 
\label{more models}
% \vspace{-0.03in}

To further demonstrate the efficacy of RAP, we evaluate our method on more diverse network  architectures, including Inception-ResNet-v2, NASNet-Large and ViT-Base/16.
We adopt ResNet-50 as the surrogate model and the results are shown in Table \ref{table_ensemble}, {\it col 2-7}.
As shown in the table, the proposed RAP-LS achieves significant improvements for all three combinational methods on all target models,
and MTDAI+RAP-LS achieves the best performance for diverse models. 
For MTDAI, the average performance improvements induced by RAP-LS is 5.9\% and 7.8\% for untargeted and targeted attacks, respectively.
Since ViT is based on the transformer architecture that totally being different from convolution models, the transfer attacks based on Resnet-50 behave relatively poor on it, especially on targeted attacks. 
Yet our RAP-LS still obtains consistent improvements for all compared methods.
We also consider the ensemble-model attack on these diverse network architectures and the results are given in {\it Appendix.}

Furthermore, we evaluate RAP on attacking defense models. We choose ensemble adversarial training (AT) model, adv-Inc-v3 and ens-adv-Inc-Res-v2 \citep{tramer2018ensemble}, multi-step AT model \citep{NEURIPS2020_24357dd0}, imput transformation defense \citep{xie2018mitigating}, feature denoising \citep{xie2019feature}, and purification defense (NRP) \citep{naseer2020self}. We only demonstrate the results of ensemble AT in main submission. The evaluations of other defense models are shown in \textit{Appendix}.
Following prior works \citep{wang2021admix,xie2019improving}, we adopt the ensemble-model attack by averaging the logits of different surrogate models, including ResNet-50, ResNet-101, Inception-v3, and Inception-ResNet-v2. 
The transfer attack success rate on defense models are shown in Table \ref{table_ensemble}, {\it col 8-11}. 
We can observe that our RAP-LS further boosts transferability of the baseline methods on both targeted and untargeted attacks. 
For untargeted attacks, RAP-LS achieves average performance improvements of $9.8\%$ and $14.1\%$ on Inc-v3$_{adv}$ and IncRes-v2$_{ens}$, respectively. 
For targeted attacks, the average performance improvements of RAP-LS are $14.8\%$ and $11.1\%$, respectively.

% CE Loss for untargted attack
% MaxLogit Loss for targted attack
% \vspace{-0.05in}

% \vspace{-0.06in}
\subsection{Ablation Study}
\label{ablation study}
% \vspace{-0.04in}

We conduct ablation study on the hyper-parameters of the proposed RAP, including the size of neighborhoods $\epsilon_{n}$, the iteration number of inner optimization $T$ and late-start $K_{LS}$.
We adopt targeted attacks with ResNet-50 as the surrogate model.

\begin{figure}[t]
    \vspace{-0.05in}
    \centering
    \includegraphics[width=1\textwidth]{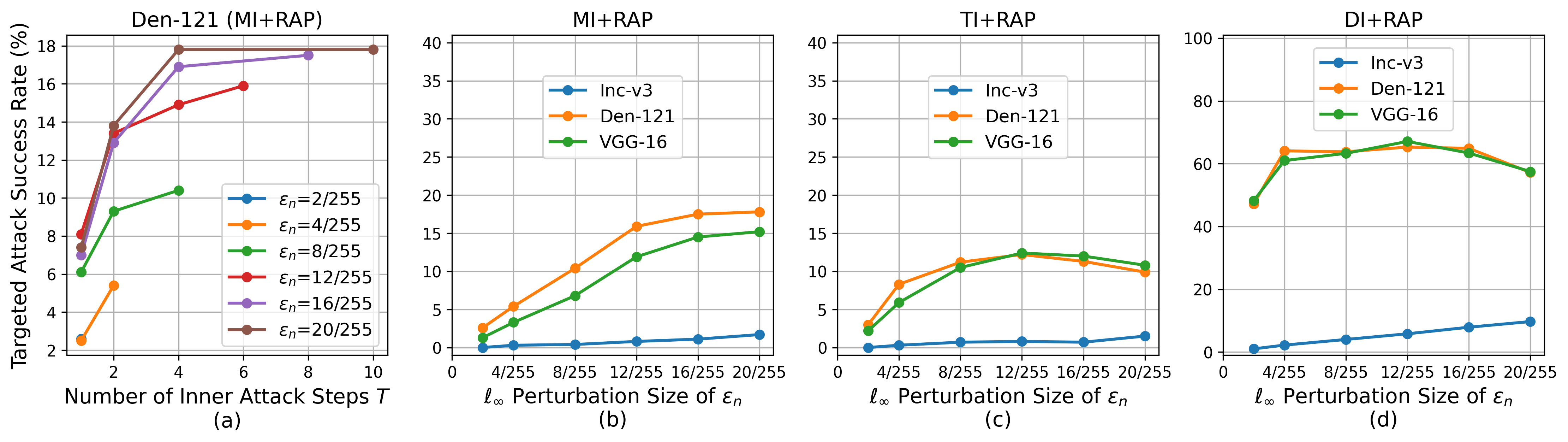}
    \vspace{-0.15in}
    \small{\caption{Targeted attack success rate ($\%$) with various $T$ and $\epsilon_{n}$. Res-50 is set as surrogate model.
    }
    \vspace{-0.02in}
    \label{fig-inner}}
\end{figure}

%and set the minimum acceptable value of $\alpha_{n}$ to $2/255$
We first evaluate the effect of $\epsilon_{n}$ and $T$.
We consider different values of $\epsilon_{n}$, including $2/255$, $4/255$, $8/255$, $12/255$, $16/255$, and $20/255$. 
In Figure \ref{fig-inner} (a), we plot the tendency curves of the targeted attack success rate under different values of $\epsilon_{n}$ and $T$.
Note that in Sec.~\ref{method}, we set $\alpha_{n} = \epsilon_{n}/T$. 
Thus for a fixed $\epsilon_{n}$, larger $T$ indicating lower stepsize $\alpha_{n}$.
The minimum stepsize of $\alpha_{n}$ is set to $2/255$.
We have the following observation from the plot: for a fixed $\epsilon_{n}$, the more iterations $T$, the better attack performance. 
Thus, we adopt a relatively smaller $\alpha_{n}=2/255$ in our experiments.
% 2) larger $\epsilon_{n}$ generally leads to better attack performance.
%
In Figure~\ref{fig-inner} (b-d), we further plot the results of different attack methods and target models \wrt $\epsilon_{n}$, where $\alpha_{n} = 2/255$.
As shown in the plots, the larger $\epsilon_{n}$ generally improves the attack performance. 
For Inception-v3 and DenseNet-121, the improvements become mild for even larger $\epsilon_{n}$. 
Overall, the value of 12 or 16 could lead to satisfactory result under most cases.
% For I and TI in untargeted attack, we use the $12/255$. For all other settings, we adopt $16/255$. 

\begin{figure}[t]
    \vspace{-0.07in}
    \centering
    \includegraphics[width=1\textwidth]{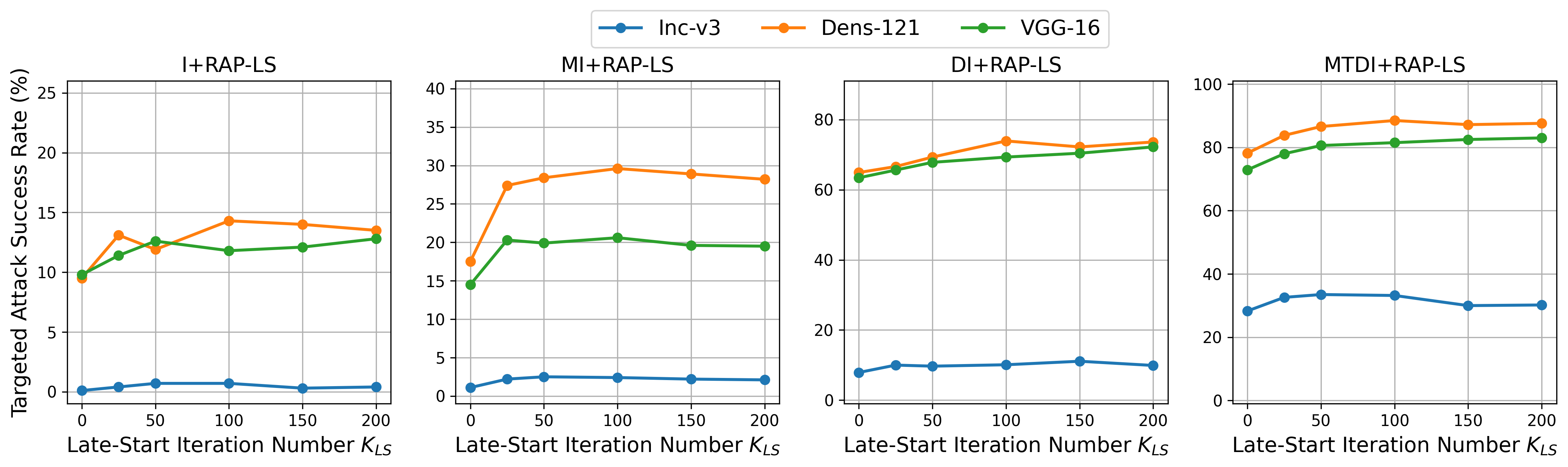}
    \vspace{-0.15in}
    \small{\caption{
    Targeted attack success rate ($\%$) with various $K_{LS}$. Res-50 is set as surrogate model.}
    \label{fig-late-start}
    }
    \vspace{-0.05in}
\end{figure}

Then we conduct the ablation study of $K_{LS}$. In Figure~\ref{fig-late-start}, we report the targeted attack success rate of I, MI, DI, and MI-TI-DI combined with RAP-LS with $K_{LS} = 0, 25,50,100,150,200$. 
Note that the RAP-LS with $K_{LS} = 0$ reduces to RAP.
As shown in the plots, the proposed late-start strategy can further boost attack performance of RAP for most cases. 
In general, the performance improvements increase as $K_{LS}$ increases, and then become mild when $K_{LS}$ is larger than 100.
The suitable value of $K_{LS}$ is relatively consistent among different methods and target models.
% For each method with RAP-LS, it behaves the almost same trend on different target models. Therefore, for each method, we could choose the same $K_{LS}$ across the different target models. 

% \begin{figure*}[t]
% \vskip -0.15in
% \centering
% \scalebox{0.94}{
% \subfigure{
% \begin{minipage}[htp]{0.25\linewidth}
% \centering
% \includegraphics[width=1.4in]{figures/InnerAttack_All/MI/Den.png}
% \end{minipage}%
% }%
% \subfigure{
% \begin{minipage}[htp]{0.25\linewidth}
% \centering
% \includegraphics[width=1.4in]{figures/InnerAttack/Method/MI.png}
% \end{minipage}%
% }%
% \subfigure{
% \begin{minipage}[htp]{0.25\linewidth}
% \centering
% \includegraphics[width=1.4in]{figures/InnerAttack/Method/TI.png}
% \end{minipage}
% }%
% % \rulesep
% \subfigure{
% \begin{minipage}[htp]{0.25=\linewidth}
% \centering
% \includegraphics[width=1.4in]{figures/InnerAttack/Method/DI.png}
% \end{minipage}
% }}
% \centering
% \vskip -0.2in
% \small{\caption{Targeted attack success rate ($\%$) with various $T$ and $\epsilon_{n}$. Res-50 is set as surrogate model.
% }
% \label{fig-inner}}
% \vskip -0.05in
% \end{figure*}

% \begin{figure*}[t]
% \vskip -0.1in
% \centering
% \scalebox{0.94}{
% \subfigure{
% \begin{minipage}[htp]{0.25\linewidth}
% \centering
% \includegraphics[width=1.4in]{figures/LateStart/new/I.png}
% \end{minipage}%
% }%
% \subfigure{
% \begin{minipage}[htp]{0.25\linewidth}
% \centering
% \includegraphics[width=1.4in]{figures/LateStart/new/MI.png}
% \end{minipage}%
% }%
% \subfigure{
% \begin{minipage}[htp]{0.25\linewidth}
% \centering
% \includegraphics[width=1.4in]{figures/LateStart/new/DI.png}

% \end{minipage}
% }%
% % \rulesep
% \subfigure{
% \begin{minipage}[htp]{0.25\linewidth}
% \centering
% \includegraphics[width=1.4in]{figures/LateStart/new/MI-TI-DI.png}
% \end{minipage}
% }}
% \centering
% \vskip -0.1in
% \small{\caption{
% Targeted attack success rate ($\%$) with various $K_{LS}$. Res-50 is set as surrogate model.}
% \label{fig-late-start}
% }
% \vskip -0.1in
% \end{figure*}

% \vspace{-0.05in}
\subsection{The Targeted Attack Against Google Cloud Vision API}
\label{api}
% \vspace{-0.02in}
Finally, we conduct the transfer attacks to attack a practical and widely used image recognition system, Google Cloud Vision API, and in the more challenging targeted attack scenario. 
MTDAI-RAP-LS behaves the best performance in above experiments, so we choose it to conduct the attack. 
We take the evaluation on randomly selected 500 images and use ResNet-50 as surrogate model. 
As the API returns 10 predicted labels for each query, to evaluate the attacking performance, we test whether or not the target class appears in the returned predictions. 
Since the predicted label space of Google Cloud Vision API do not fully correspond to the 1000 ImageNet classes, we manually treat classes with similar semantics to be the same classes. 
In comparison, the baseline MTDAI successfully attacks $232$ images against the Google API. 
Our RAP-LS achieves a large improvement, successfully attacking $342$ images, leading to a 22.0\% performance improvements. 
These demonstrates the high efficacy of our method to improve transferability on real-world system.

% \subsection{The Discussion and Limitation}

% {\color{red}Do we need to talk about the computation cost of our method. (see the rebuttal of ICLR). and the discussion of flatness and transferability}

% \vspace{-0.07in}
\section{Conclusion}
\label{conclusion}
% \vspace{-0.07in}
In this work, we study the transferability of adversarial examples that is significant for black-box attacks.
The transferability of adversarial examples is generally influenced by the overfitting of surrogate models.
To alleviate this, we propose to seeking adversarial examples that locate at flatter local regions.
That is, instead of optimizing the pinpoint attack loss, we aim to obtain a consistently low loss at the  neighbor regions of the adversarial examples.
We formulate this as a min-max bi-level optimization problem, where the inner maximization aims to inject the worse-case perturbation for the adversarial examples.
We conduct a rigorous experimental study, covering  untargeted attack and targeted attack, standard and defense models, and a real-world Google Clould Vision API.
The experimental results demonstrate that RAP can significantly boost the transferability of adversarial examples, which also demonstrates that transfer attacks have become serious threats. We need to consider how to effectively defense against them.

\subsubsection*{Acknowledgments}
% \vspace{-0.05in}
We want to thank the anonymous reviewers for their valuable suggestions and comments.
This work is supported by the National Natural Science Foundation of China under grant No.62076213, Shenzhen
Science and Technology Program under grant No.RCYX20210609103057050, and the university
development fund of the Chinese University of Hong Kong, Shenzhen under grant No.01001810, and
sponsored by CCF-Tencent Open Fund.

\bibliographystyle{plainnat}
\bibliography{reference}

\newpage

\appendix

\section{Social Impact}
\label{impact}
Deep neural networks (DNNs) have been successfully applied in many safety-critical tasks, such as autonomous driving, face recognition and verification, \etc. And adversarial samples have posed a serious threat to machine learning systems. For real-world applications, the DNN model as well as the training dataset, are often hidden from users. Therefore, the attackers need to generate the adversarial examples under black-box setting where they do not know any information of the target model. For black-box setting, the adversarial transferability matters since it can allow the attackers to attack target models by using adversarial examples generated on the surrogate models. This work can potentially contribute to understanding of transferability of adversarial examples. Besides, the better transferability of adversarial examples calls the machine learning and security communities into action to create stronger defenses and robust models against black-box attacks.

\section{Implementation Details}
\label{implementation details}

We conducted all experiments in an Nvidia-V100 GPU. And we run all experiments $3$ times and average
all results over $3$ random seeds.

\paragraph{Dataset} The used two datasets are licensed under MIT. Imagenet is licensed under Custom (non-commercial).

\paragraph{Implementation Details of Evaluated Models.} For ResNet-50, DenseNet-121, VGG-16, Inception-v3, we adopt the pre-trained models provided by torchvision package. For Inception-ResNet-v2, NASNet-Large, ViT-Base/16, adv-Inc-v3, and ens-adv-Inc-Res-v2, we adopt the provided pre-trained models\footnote{\url{https://github.com/rwightman/pytorch-image-models}}. 

\paragraph{Implementation Details of Baseline Attack Methods.} We adopt the source code \footnote{\url{https://github.com/ZhengyuZhao/Targeted-Tansfer}} provided by \citet{zhao2020success} to implement I, MI, TI, and DI attacks. The decay factor for MI is set as $1.0$. The kernel size is set as $5$ for TI attack, following \citet{gao2020patch}. The transformation probability is set as $0.7$ for DI. For SI and Admix, we adopt the parameters suggested in \citet{wang2021admix}. The number of copies for SI is set as $5$. The number of randomly sample $m_2$ and $\eta$ of Admix are set as $3$ and $0.2$ respectively. For implementation of ILA and LinBP, we utilize the source code \footnote{\url{https://github.com/qizhangli/linbp-attack}} provided by \citet{GuoLinBP}. For implementation of TTP, we use the pre-trained generator \footnote{\url{https://github.com/Muzammal-Naseer/TTP}} based on ResNet-50 provided by \cite{naseer2021generating}. 

\paragraph{Computational Cost.} Here, we analyze the computational cost of our method. In Algorithm 1 with global iteration number $K$, late-start iteration number $K_{LS}$ and inner iteration number $T$, our RAP-LS requires $K+\left(K-K_{L S}\right) * T$ forward and backward calculation. While the original attack algorithm requires $K$ forward and backward calculation. The extra computation cost of RAP-LS is $\left(K-K_{L S}\right) * T$ times forward and backward calculation. 

The adversarial example generation process is conducted based on the offline surrogate models. Compared with this offline time cost, the attacking performance is much more important for black-box attacks. Besides, our late-start strategy could alleviate the time cost.

\paragraph{Implementation Details of Visualization.}  We visualize the flatness of the loss landscape around $\bm{x}^{adv}$ on surrogate model by plotting the loss change when moving $\bm{x}^{adv}$ along a random direction with different magnitudes. Specially, we first sample $\bm{d}$ from a Gaussian distribution and normalize it on a $\ell_{2}$ unit norm ball, $\bm{d} \leftarrow  \frac{\bm{d}}{\|\bm{d}\|_{F}}$. Then, we calculate the loss change (flatness) $f(a)$ with different magnitudes $a$,   
\begin{equation}
\begin{aligned}
f(a) = \mathcal{L} (\mathcal{M}^s (\mathcal{G}(\bm{x}^{adv}+a\cdot\bm{d});\bm{\theta}), y_t) - \mathcal{L} (\mathcal{M}^s (\mathcal{G}(\bm{x}^{adv});\bm{\theta}), y_t).
\end{aligned}
\end{equation}
Considering $\bm{d}$ is randomly selected, we repeat the above calculation 20 times with different $\bm{d}$ and take the averaged value to conduct the visualization.

We also add the visualization results about targeted attacks. 

\begin{figure}[H]
    \vspace{-0.05in}
    \centering
    \includegraphics[width=0.94\textwidth]{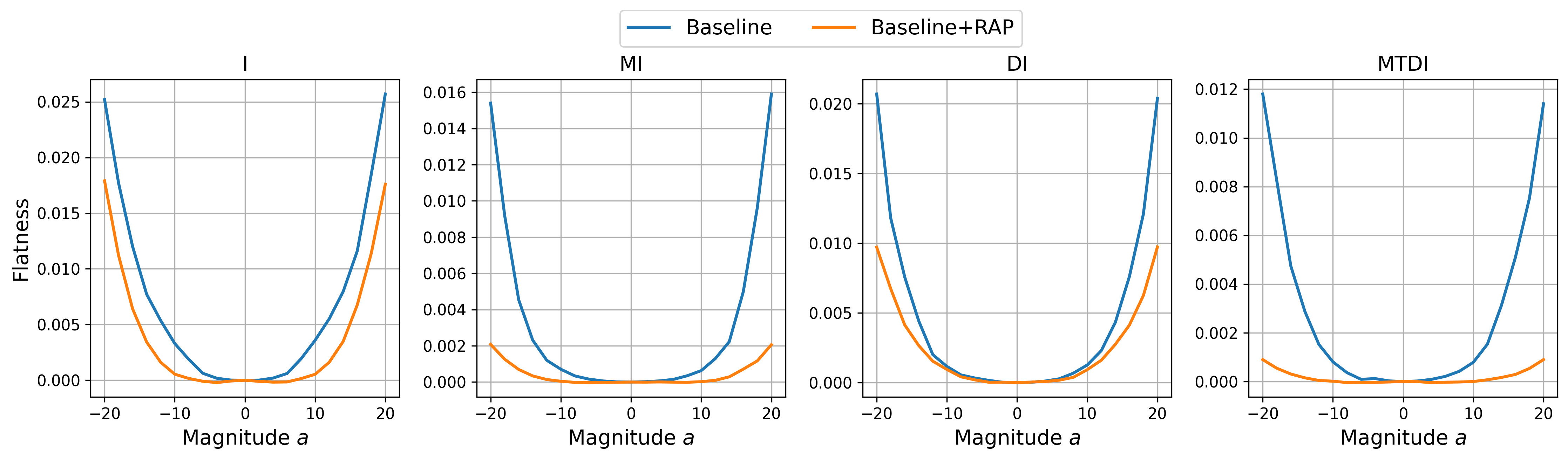}
    \vspace{-0.05in}
    \small{\caption{The flatness visualization of targeted adversarial examples.}
    \label{rebuttal_targeted_flatness}
}
\vspace{-0.2in}
\end{figure}

\section{Experimental Results about More baseline attacks}
\label{appendix_attack}

In this section, we show the comparison of our method RAP and EOT \citep{athalye2018synthesizing}, VT \citep{Wang_2021_CVPR}, EMI \citep{wang2021boosting}, and Ghost Net \citep{li2020learning} attack methods. 

\subsection{Experimental Results about VT and EMI}
\label{appendix-vt-emi}

In the below Table \ref{table_vt-emi} and \ref{table_vt-emi-target}, we demonstrate the untargeted and targeted attack performance of VT, EMI, and our methods. We choose MI-TI-DI as the baseline method and follow experimental settings in Section \ref{evaluation settings}. As shown in experimental results, our RAP-LS achieves better performance, especially for targeted attacks. Compared with VT, RAP-LS gets an increase of $6.7\%$ for targeted attacks in terms of average success rate. This demonstrates the effectiveness of our methods. 

\begin{table}[H]
\small{\caption{
The untargeted attack success rate ($\%$) of VT, EMI, and RAP-LS with the MI-TI-DI baseline.}
\label{table_vt-emi}
}
\vspace{-0.03in}
\begin{center}
\begin{small}
% \begin{sc}
\scalebox{0.8}{
\begin{tabular}{c|ccc|ccc}
\toprule
\multirow{2}{*}{\tabincell{c}{Attack}}  &  & \textbf{ResNet-50} $\Longrightarrow$ &  &  & \textbf{DenseNet-121}$\Longrightarrow$ & \\
& Dense-121 & VGG-16 & Inc-v3    & Res-50 & VGG-16 & Inc-v3  \\
\midrule 
MI-TI-DI
% e = 16/255, k = 8
% & 79.2 / 81.8 / 79.5
% & 78.0 / 79.9 / 81.1
% & 34.6 / 45.9 / 46.2

% e = 12/255, k = 6
&99.8	&99.8 &85.7

% e = 16/255, k = 8
% & 87.1 / 86.1 / 86.2
% & 85.1 / 84.6 / 84.6
% & 46.5 / 52.4 / 54.2

% e = 12/255, k = 6
&99.4	&99.2	&89.1

\\ MI-TI-DI-VT
&100	&100  &95.8	

& 100 
& 100 
& 96.0 

\\ EMI-TI-DI 
% e = 16/255, k = 8
% & 82.0 / 82.9 / 81.6
% & 81.0 / 82.7 / 80.6
% & 45.5 / 54.3 / 51.9

% e = 12/255, k = 6
&100	&100 &93.6	

% e = 16/255, k = 8
% & 89.6 / 87.7 / 87.5
% & 87.0 / 85.3 / 85.9
% & 54.2 / 60.4 / 59.5

% e = 12/255, k = 6
&100	&100 &94.2

\\ MI-TI-DI+RAP-LS
&100	&100 &96.9	
 
&100	&100 &97.1	 

\\
% MI-DI-FGSM   \\
% MI-TI-DI-FGSM     \\
% MI-TI-DI-SI-FGSM    \\
% MI-TI-DI-Admix    \\
\midrule
\multirow{2}{*}{\tabincell{c}{Attack}}  &  & \textbf{VGG-16} $\Longrightarrow$ &  &  & \textbf{Inc-v3}$\Longrightarrow$ & \\  & Res-50 & Dense-121 & Inc-v3 & Res-50 & Dense-121 & VGG-16 \\
\midrule
MI-TI-DI
% e = 16/255, k = 8
% & 79.2 / 81.8 / 79.5
% & 78.0 / 79.9 / 81.1
% & 34.6 / 45.9 / 46.2

% e = 12/255, k = 6
&90 	&88.8 	&56.8

% e = 16/255, k = 8
% & 87.1 / 86.1 / 86.2
% & 85.1 / 84.6 / 84.6
% & 46.5 / 52.4 / 54.2

% e = 12/255, k = 6
& 82.9 
& 85.7 
& 85.1 

\\ MI-TI-DI-VT
&93.9 	&93	&76.5

% e = 16/255, k = 8
% & 87.1 / 86.1 / 86.2
% & 85.1 / 84.6 / 84.6
% & 46.5 / 52.4 / 54.2

% e = 12/255, k = 6
& 87.1
& 90.3 
& 87.5

\\ EMI-TI-DI 
% e = 16/255, k = 8
% & 82.0 / 82.9 / 81.6
% & 81.0 / 82.7 / 80.6
% & 45.5 / 54.3 / 51.9

% e = 12/255, k = 6
&91.7 	&91.5	&74.3

% e = 16/255, k = 8
% & 87.1 / 86.1 / 86.2
% & 85.1 / 84.6 / 84.6
% & 46.5 / 52.4 / 54.2

% e = 12/255, k = 6
& 86
& 88.4 
& 86.2 

\\ MI-TI-DI+RAP-LS
& 97.7 
& 97.3 
& 81.4 
& 90.6 
& 93.3 
& 91.0

\\
% MI-DI-FGSM    \\
% MI-TI-DI-FGSM   \\
% MI-TI-DI-SI-FGSM     \\
% MI-TI-DI-Admix   \\
\bottomrule
\end{tabular}
}
% \end{sc}
\end{small}
\end{center}
\vspace{-0.1in}
\end{table}

\begin{table}[H]
\small{\caption{
The targeted attack success rate ($\%$) of VT, EMI, and RAP-LS with the MI-TI-DI baseline}.
\label{table_vt-emi-target}
}
\vspace{-0.03in}
\begin{center}
\begin{small}
% \begin{sc}
\scalebox{0.8}{
\begin{tabular}{c|ccc|ccc}
\toprule
\multirow{2}{*}{\tabincell{c}{Attack}}  &  & \textbf{ResNet-50} $\Longrightarrow$ &  &  & \textbf{DenseNet-121}$\Longrightarrow$ & \\
& Dense-121 & VGG-16 & Inc-v3    & Res-50 & VGG-16 & Inc-v3  \\
\midrule 
MI-TI-DI
% e = 16/255, k = 8
% & 79.2 / 81.8 / 79.5
% & 78.0 / 79.9 / 81.1
% & 34.6 / 45.9 / 46.2

% e = 12/255, k = 6
&74.9	&62.8 &10.9

% e = 16/255, k = 8
% & 87.1 / 86.1 / 86.2
% & 85.1 / 84.6 / 84.6
% & 46.5 / 52.4 / 54.2

% e = 12/255, k = 6
&44.9	&38.5	&7.7

\\ MI-TI-DI-VT
&82.5	&71.9 &21.6	

&59.2	&53.6	&21.3

\\ EMI-TI-DI 
% e = 16/255, k = 8
% & 82.0 / 82.9 / 81.6
% & 81.0 / 82.7 / 80.6
% & 45.5 / 54.3 / 51.9

% e = 12/255, k = 6
&79.1	&67.8 &19.2	

&56.3	&50.4	&19.8

\\ MI-TI-DI+RAP-LS
&88.5	&81.5 &33.2	
 
&74.5	&65.5 &26.5

\\
% MI-DI-FGSM   \\
% MI-TI-DI-FGSM     \\
% MI-TI-DI-SI-FGSM    \\
% MI-TI-DI-Admix    \\
\midrule
\multirow{2}{*}{\tabincell{c}{Attack}}  &  & \textbf{VGG-16} $\Longrightarrow$ &  &  & \textbf{Inc-v3}$\Longrightarrow$ & \\  & Res-50 & Dense-121 & Inc-v3 & Res-50 & Dense-121 & VGG-16 \\
\midrule
MI-TI-DI
% e = 16/255, k = 8
% & 79.2 / 81.8 / 79.5
% & 78.0 / 79.9 / 81.1
% & 34.6 / 45.9 / 46.2

% e = 12/255, k = 6
&11.8 	&13.7 	&0.7

% e = 16/255, k = 8
% & 87.1 / 86.1 / 86.2
% & 85.1 / 84.6 / 84.6
% & 46.5 / 52.4 / 54.2

% e = 12/255, k = 6
& 1.8 
& 4.1
& 2.9 

\\ MI-TI-DI-VT
&19.3 	&22.5	&2.5

% e = 16/255, k = 8
% & 87.1 / 86.1 / 86.2
% & 85.1 / 84.6 / 84.6
% & 46.5 / 52.4 / 54.2

% e = 12/255, k = 6
& 5.6
& 9.8 
& 6.4

\\ EMI-TI-DI 
% e = 16/255, k = 8
% & 82.0 / 82.9 / 81.6
% & 81.0 / 82.7 / 80.6
% & 45.5 / 54.3 / 51.9

% e = 12/255, k = 6
&14.1	&19.7	&2.0

% e = 16/255, k = 8
% & 87.1 / 86.1 / 86.2
% & 85.1 / 84.6 / 84.6
% & 46.5 / 52.4 / 54.2

% e = 12/255, k = 6
& 4.3
& 8.0
& 5.2 

\\ MI-TI-DI+RAP-LS
& 22.9 
& 27.4 
& 4.6 
& 7.5
& 13.4
& 9.8

\\
% MI-DI-FGSM    \\
% MI-TI-DI-FGSM   \\
% MI-TI-DI-SI-FGSM     \\
% MI-TI-DI-Admix   \\
\bottomrule
\end{tabular}
}
% \end{sc}
\end{small}
\end{center}
\vspace{-0.1in}
\end{table}

\subsection{Experimental Results about Ghost Net attack}
\label{appendix-ghost}

For combining Ghost net with RAP, we conduct the experiments on our PyTorch codes following the original TensorFlow codes provided by the authors. The main idea of ghost network is to perturb skip connections of ResNet to generate ensemble networks. To achieve this goal, the authors multiply skip connection by the random scalar $r$ sampled from a uniform distribution. We reimplement this procedure with the hyperparameter about $r$ recommended in the orginal paper on the PyTorch ResNet-50 model.

The targeted attack results ($\%$) are shown in the below Table \ref{append-gn}. We also follow the experimental settings in Section \ref{evaluation settings}. Here, we use GN to represent Ghost Net method. The results show that ghost network method can improve the transfer attack performance. Combined with RAP-LS, the adversarial transferability can be further improved especially on the inception-v3 model.

\begin{table}[H]
\small{\caption{The targeted attack success rate ($\%$) of GN, and RAP-LS with the MI-TI-DI baseline}
\label{append-gn}
}
\vskip -0.25in
\begin{center}
\begin{small}
% \begin{sc}
\scalebox{0.9}{
\begin{tabular}{c|ccc}
\toprule
\multirow{2}{*}{\tabincell{c}{Attack}} & \multicolumn{3}{c}{\textbf{ResNet-50}~$\Longrightarrow$}   \\
& Dense-121 & VGG-16 & Inc-v3 \\
\midrule 
MTDI 
% Ensemble
& 74.9	
& 62.8
& 10.9

\\ MTDI-GN
% Ensemble

&85.9	

&80.7

&24.9

\\ MTDI-GN+RAP-LS 
% Ensemble
&89.6	&87.7

&49.7	

\\

\bottomrule
\end{tabular}
}
% \end{sc}
\end{small}
\end{center}
\vskip -0.1in
\end{table}

\subsection{Experimental Results about EOT attack}
\label{appendix-eot}

We conducted the experiment of the EOT baseline. We choose the ResNet-50 as the source model and MI-TI-DI as the baseline method. Instead of adding input transformation once like DI, we sample random transformation (resizing and padding) multiple times in each iteration. Then, we add them to  following the expectation of transformation (EOT) \citep{athalye2018synthesizing}. We set the number of sampling as 10. Our RAP can be also naturally combined with EOT attack. 

The targeted attack results ($\%$) are shown in the below Table \ref{append-eot}.  We also follow the experimental settings in Section \ref{evaluation settings}.

As shown in the below table, \textbf{1)} the EOT attack gets a moderate increase on attack performance compared with the baseline MI-TI-DI attack, which demonstrates that EOT could improve adversarial transferability. \textbf{2)} Our RAP attack achieves better performance and surpasses the EOT attack by a large margin, especially for Inc-v3 and VGG-16 target models. \textbf{3)} Combining RAP with EOT can further boost EOT attack performance. \textbf{These results demonstrate that RAP could achieve better adversarial transferability and help find better flat local minima. Besides, the combination of RAP and EOT achieves the best performance among them, which demonstrates that these two methods could complement each other. }

\begin{table}[H]
\small{\caption{The targeted attack success rate ($\%$) of EOT, and RAP-LS with the MI-TI-DI baseline}
\label{append-eot}
}
\vskip -0.25in
\begin{center}
\begin{small}
% \begin{sc}
\scalebox{0.9}{
\begin{tabular}{c|ccc}
\toprule
\multirow{2}{*}{\tabincell{c}{Attack}} & \multicolumn{3}{c}{\textbf{ResNet-50}~$\Longrightarrow$}  \\
& Dense-121 & VGG-16 & Inc-v3 \\
\midrule 
MTDI 
% Ensemble
& 74.9	
& 62.8
& 10.9

\\ MTDI-EOT
% Ensemble

&76.9	&66.9

&11.2	

\\ MTDI+RAP 
% Ensemble
&78.2	&72.9

&28.3	

\\ MTDI-EOT+RAP 
% Ensemble
&86.1	&79.5

&32.8	

\\

\bottomrule
\end{tabular}
}
% \end{sc}
\end{small}
\end{center}
\vskip -0.1in
\end{table}

\section{Experimental Results about More Defense Models}
\label{appendix_defense}

\paragraph{The Evaluation on More Defense Models} Here, we show the evaluation of more defense models containing  multi-step Adversarial training models in ImageNet \citep{NEURIPS2020_24357dd0}, Feature Denoising \citep{xie2019feature}, NRP \citep{naseer2020self}, input transformation defense (R$\&$P) \citep{xie2018mitigating}.

For Feature Denoising, we utilize the pre-trained ResNet-152 model provided by the authors \footnote{\url{https://github.com/facebookresearch/ImageNet-Adversarial-Training}}. For AT models on ImageNet, we adopt the pre-trained ResNet-50 AT models provided by the authors \footnote{\url{https://github.com/microsoft/robust-models-transfer}}. For $\ell_{\infty}$ norm, we adopt the ResNet-50 AT model with budget $4/255$, which ranks first in the RobustBench leaderboard \footnote{\url{https://robustbench.github.io}}. For $\ell_{2}$ norm, we adopt the ResNet-50 AT model with budget $0.5$. The untargeted attack performance is shown in Table \ref{rebuttal-adv}. We follow the experimental settings in Section \ref{more models} of the main submission.  We can observe that our RAP-LS further boosts the transferability of baseline methods on these new defense models, getting a $5.5\%$ boost for the average attack success rate.

For NRP, we adopt the pre-trained purifiers provided by the authors \footnote{\url{https://github.com/Muzammal-Naseer/NRP}}. Since NRP is an offline defense module, we combine it with the two used ensemble AT models and the above two AT models. The untargeted attack performance is shown in Table \ref{rebuttal-nrp}. We also follow the experimental settings in Section \ref{more models} of the main submission. Combining NRP with AT models is a much stronger defense mechanism, but RAP-LS still achieves an improvement by $0.8\%$.

For R$\&$P, we adopt the source code provided by \citep{dong2020benchmarking} to implement it. We also combine R$\&$P with the two used ensemble AT models and the two new AT models above. The untargeted attack performance is shown in Table \ref{rebuttal-rp}. We also follow the experimental settings in Section \ref{more models} of the main submission. For R$\&$P, RAP-LS achieves an $9.1\%$ increase in terms of average attack success rate.

\begin{table}[H]
\small{\caption{{
The evaluation of ensemble attacks on \textbf{two AT models} and \textbf{Feature Noising}.}}
\label{rebuttal-adv}
}
\vskip -0.25in
\begin{center}
\begin{small}
% \begin{sc}
\scalebox{0.9}{
\begin{tabular}{c|ccc}
\toprule
\multirow{2}{*}{\tabincell{c}{Attack}} & \multicolumn{3}{c}{\textbf{Untarged}}  \\
& Res-50 AT ($\ell_{2}$) & Res-50 AT ($\ell_{\infty}$) & Feature Denoising \\
\midrule 
MTDI 
% Ensemble
& 42.5
& 32.4
& 44.1

\\ MTDI+RAP-LS
% Ensemble
& \textbf{59.5}
& \textbf{34.4}
& \textbf{44.4}

\\ MTDSI 
% Ensemble
& 56.6
& 35.8
& 45.0

\\ MTDSI+RAP-LS
% Ensemble
& \textbf{70.3}
& \textbf{36.6}
& \textbf{45.7}

\\ MTDAI
% Ensemble
& 62.1
& 35.6
& 44.2

\\ MTDAI+RAP-LS
% Ensemble
& \textbf{73.7}
& \textbf{37.7}
& \textbf{45.2}

\\

\bottomrule
\end{tabular}
}
% \end{sc}
\end{small}
\end{center}
\vskip -0.1in
\end{table}

\begin{table}[H]
\small{\caption{The evaluation of ensemble attacks on defense models with \textbf{NRP}.}
\label{rebuttal-nrp}
}
\vskip -0.25in
\begin{center}
\begin{small}
% \begin{sc}
\scalebox{0.9}{
\begin{tabular}{c|cccc}
\toprule
\multirow{2}{*}{\tabincell{c}{Attack}} & \multicolumn{4}{c}{\textbf{Untarged}}  \\
& Inc-v3$_{adv}$ & IncRes-v2$_{ens}$ & Res-50 AT ($\ell_{2}$) & Res-50 AT ($\ell_{\infty}$)\\
\midrule 
MTDI 
% Ensemble
& \textbf{23.1}
& 13.5
& 14.2
& 25.7

\\ MTDI+RAP-LS
% Ensemble
& 22.7
& \textbf{14.8}
& \textbf{14.9}
& \textbf{26.3}

\\ MTDSI 
% Ensemble
& 22.5
& 14.2
& 15.0
& 26.1

\\ MTDSI+RAP-LS
% Ensemble
& \textbf{24.5}
& \textbf{15.3}
& \textbf{15.4}
& \textbf{26.2}

\\ MTDAI
% Ensemble
& 24.1
& 14.7
& 14.2
& 25.9

\\ MTDAI+RAP-LS
% Ensemble
& \textbf{24.9}
& \textbf{15.6}
& \textbf{15.3}
& \textbf{26.1}

\\

\bottomrule
\end{tabular}
}
% \end{sc}
\end{small}
\end{center}
\vskip -0.1in
\end{table}

\begin{table}[H]
\small{\caption{The evaluation of ensemble attacks on defense models with \textbf{R\&P}.}
\label{rebuttal-rp}
}
\vskip -0.25in
\begin{center}
\begin{small}
% \begin{sc}
\scalebox{0.9}{
\begin{tabular}{c|cccc}
\toprule
\multirow{2}{*}{\tabincell{c}{Attack}} & \multicolumn{4}{c}{\textbf{Untarged}}  \\
& Inc-v3$_{adv}$ & IncRes-v2$_{ens}$ & Res-50 AT ($\ell_{2}$) & Res-50 AT ($\ell_{\infty}$)\\
\midrule 
MTDI 
% Ensemble
& 65.0
& 46.2
& 52.5
& 43.7

\\ MTDI+RAP-LS
% Ensemble
& \textbf{82.1}
& \textbf{63.2}
& \textbf{65.3}
& \textbf{45.8}

\\ MTDSI 
% Ensemble
& 86.5
& 69.6
& 64.1
& 45.9

\\ MTDSI+RAP-LS
% Ensemble
& \textbf{93.4}
& \textbf{84.9}
& \textbf{74.0}
& \textbf{46.2}

\\ MTDAI
% Ensemble
& 88.9
& 76.5
& 68.4
& 46.2

\\ MTDAI+RAP-LS
% Ensemble
& \textbf{94.8}
& \textbf{87.0}
& \textbf{77.7}
& \textbf{47.7}

\\

\bottomrule
\end{tabular}
}
% \end{sc}
\end{small}
\end{center}
\vskip -0.1in
\end{table}

The above experimental results also show that RAP is less effective when attacking Feature Denoising. We think this is mainly due to the specially designed feature denoising block (\ie the non-local block), and the different settings of maximum perturbation size during adversarial training, as follows. 

\begin{itemize}
    \item Feature Denoising \citep{xie2019feature} inserts several non-local blocks into network to eliminate the adversarial noise at the feature level. According to \citep{xie2019feature}, for input feature map $F_{i}$, the non-local block computes a denoised output feature map $F_0$ by taking a weighted average of input features in all spatial locations. Through this, the non-local block would model the global relationship between features in all spatial locations, which may smooth the learned decision boundary. Recalling that our RAP is to boost the transferability by seeking for a flat local minimum. The smoothness of decision boundary could make it harder to escape from certain local minimum, especially for small attack perturbation size, so as to limit the performance improvement of RAP.
    \item In our experiment, for Feature Denoising, the maximum perturbation size during their training is set to 16/255. In Table 12, the maximum perturbation size of attack is also set to 16/255. The attack size of 16/255 may not be large enough for escaping from local minima for the Feature Denoising model trained with 16/255 perturbation size. In contrast, the maximum perturbation size of AT-$\ell_{\infty}$ during training is 4/255. To verify this, we conduct an ablation study of increasing the maximum perturbation size to 20/255. Using a larger perturbation size of 20/255, the attacking performance against Feature Denoising is $48.3\%$ for MI-TI-DI and $50.7\%$ for MI-TI-DI+RAP-LS. The relative performance improvement of RAP-LS is $2.4\%$, which is much larger than the relative performance improvement of $0.3\%$ in Table 12 with perturbation size 16/255, which may partially explain the phenomenon.
\end{itemize}

\vskip 0.1in
\section{Additional Experimental Results}

In this section, we first show the evaluation of targeted attacks with CE loss in Section \ref{target-CE}. Then we show the results of ensemble attacks on more diverse network architectures in Section
\ref{appendix_ensemble_attack}. In Section \ref{whole results}, we report the experimental results \wrt different value of iterations.

\subsection{The Results of Targeted Attacks with CE Loss}
\label{target-CE}

Following the settings in main submission, we evaluate the targeted attack performance of the different baseline methods with our method on ResNet-50, DenseNet-121, VGG-16, and Inception-v3.
The results of combinational methods are shown in Table \ref{table_target_combination-CE}. 
The RAP-LS outperforms all combinational methods by a significantly margin. 
Taking the average attack success rate of all target models as the evaluation metric, RAP-LS achieves $20.9\%$, $18.4\%$, and $15.1\%$ improvements over the MTDI, MTDSI and MTDAI, respectively.

\begin{table}[H]
\small{\caption{
The \textbf{targeted attack success rate ($\%$) of combinational methods with RAP}. The results with $CE$ loss and 400 iterations are reported. The best results are bold and the second best results are underlined.} 
\label{table_target_combination-CE}
}
\vskip -0.15in
\begin{center}
\begin{small}
% \begin{sc}
\scalebox{0.7}{
\begin{tabular}{c|ccc|ccc}
\toprule
\multirow{2}{*}{\tabincell{c}{Attack}}  &  & \textbf{ResNet-50} $\Longrightarrow$ &  &  & \textbf{DenseNet-121}$\Longrightarrow$ & \\
& Dense-121 & VGG-16 & Inc-v3    & Res-50 & VGG-16 & Inc-v3  \\
\midrule 
% MI-TI-DI \citep{zhao2020success}
MTDI / +RAP / +RAP-LS
& 45.5 / \underline{78.3} / \textbf{85.9}
& 29.8 / \underline{70.5} / \textbf{76.7}
& 4.5 / \underline{21.3} / \textbf{25.3}
& 20.0 / \underline{54.0} / \textbf{62.7}
& 9.9 / \underline{41.7} / \textbf{48.7}
& 2.6 / \underline{17.5} / \textbf{18.5}

% \\ +RAP / +RAP-LS  
% & 78.3 / \textbf{85.9}
% & 70.5 / \textbf{76.7}
% & 21.3 / \textbf{25.3}
% & 54.0 / \textbf{62.7}
% & 41.7 / \textbf{48.7}
% & 17.5 / \textbf{18.5}

\\
% MI-TI-DI-SI
MTDSI / +RAP / +RAP-LS
& 77.7 / \underline{89.0} / \textbf{93.7}
& 39.9 / \underline{69.4} / \textbf{76.7}
& 26.9 / \underline{45.3} / \textbf{50.8}
& 30.5 / \underline{60.4} / \textbf{69.5}
& 14.9 / \underline{42.8} / \textbf{49.7}
& 12.7 / \underline{26.6} / \textbf{32.5}

% \\ +RAP / +RAP-LS
% & 89.0 / \textbf{93.7}
% & 69.4 / \textbf{76.7}
% & 45.3 / \textbf{50.8}
% & 60.4 / \textbf{69.5}
% & 42.8 / \textbf{49.7}
% & 26.6 / \textbf{32.5}

\\
% MI-TI-DI-Admix 
MTDAI / +RAP / +RAP-LS
& 90.2 / \underline{91.4} / \textbf{96.1}
& 61.8 / \underline{73.7} / \textbf{83.4}
& 44.5 / \underline{47.9} / \textbf{59.0}
& 55.8 / \underline{68.4} / \textbf{79.3}
& 35.1 / \underline{51.8} / \textbf{64.1}
& 26.3 / \underline{32.4} / \textbf{40.4}

% \\ +RAP / +RAP-LS
% & 91.4 / \textbf{96.1}
% & 73.7 / \textbf{83.4}
% & 47.9 / \textbf{59.0}
% & 68.4 / \textbf{79.3}
% & 51.8 / \textbf{64.1}
% & 32.4 / \textbf{40.4}

\\

\midrule
\multirow{2}{*}{\tabincell{c}{Attack}}  &  & \textbf{VGG-16} $\Longrightarrow$ &  &  & \textbf{Inc-v3}$\Longrightarrow$ & \\  & Res-50 & Dense-121 & Inc-v3 & Res-50 & Dense-121 & VGG-16 \\
\midrule 
% MI-TI-DI \citep{zhao2020success}
MTDI / +RAP / +RAP-LS
& 0.5 / \underline{10.4} / \textbf{12.1}
& 0.1 / \underline{11.0} / \textbf{13.5}
& 0.0 / \underline{1.7} / \textbf{2.0}
& 2.2 / \underline{4.9} / \textbf{5.9}
& 2.2 / \underline{9.8} / \textbf{11.0}
& 1.2 / \underline{4.9} / \textbf{6.7}

% \\ +RAP / +RAP-LS 
% & 10.4 / \textbf{12.1}
% & 11.0 / \textbf{13.5}

\\
% MI-TI-DI-SI 
MTDSI / +RAP / +RAP-LS
& 5.4 / \textbf{17.4} / \underline{16.8}
& 9.5 / \textbf{28.4} / \underline{25.2}
& 2.2 / \textbf{7.1} / \underline{5.1}
& 4.4 / \underline{8.6} / \textbf{8.9}
& 7.9 / \underline{16.3} / \textbf{19.3}
& 2.0 / \textbf{6.4} / \textbf{6.4}

% \\ +RAP / +RAP-LS
% & \textbf{17.4} / 16.8
% & \textbf{28.4} / 25.2

\\
% MI-TI-DI-Admix 
MTDAI / +RAP / +RAP-LS
& 11.6 / \underline{22.6} / \textbf{26.6}
& 20.6 / \underline{32.1} / \textbf{39.1}
& 5.1 / \underline{9.2} / \textbf{9.5}
& 6.7 / \underline{12.3} / \textbf{17.0}
& 14.0 / \underline{22.9} / \textbf{29.2}
& 4.5 / \underline{9.4} / \textbf{13.2}

% \\ +RAP / +RAP-LS
% & 22.6 / \textbf{26.6}
% & 32.1 / \textbf{39.1}

\\

\bottomrule
\end{tabular}
}
% \end{sc}
\end{small}
\end{center}
\vskip -0.1in
\end{table}

\subsection{The Results of Ensemble Attacks on Diverse Network Architectures} 
\label{appendix_ensemble_attack}

We also take the evaluation of the ensemble attacks on diverse network architecture (Sec.\ref{more models}). We adopt the ensemble-model attack by averaging the logits of different surrogate models, including ResNet-50, DenseNet-121, VGG-16, and Inception-v3. The transfer attack success rate on diverse models are shown in Table \ref{table_diverse-ensemble}. Compared with results of single model attack in Table \ref{table_ensemble}, the ensemble attack achieve the better performance. We can observe that our RAP-LS further boosts transferability of the baseline methods on both targeted and untargeted attacks. We take ViT as target model for example. For untargeted attacks, RAP-LS achieves average performance improvements of $19.2\%$. For targeted attacks, RAP-LS achieves average performance improvements of $10.4\%$.

\begin{table}[H]
\small{\caption{
The evaluation of ensemble attacks on diverse network architectures.}
\label{table_diverse-ensemble}
}
\vskip -0.35in
\begin{center}
\begin{small}
% \begin{sc}
\scalebox{0.9}{
\begin{tabular}{c|ccc|ccc}
\toprule
\multirow{2}{*}{\tabincell{c}{Attack}} & \multicolumn{3}{c|}{\textbf{Untarged}} & \multicolumn{3}{c}{\textbf{Targeted}} \\
& IncRes-v2 & NASNet-L & ViT-B/16 & IncRes-v2 & NASNet-L & ViT-B/16 \\
\midrule 
MTDI 
% Ensemble
& 98.6
& 99.3
& 46.2
& 65.7
& 80.1
& 2.8

\\ MTDI+RAP-LS
% Ensemble
& \textbf{100}
& \textbf{100}
& \textbf{73.2}
& \textbf{84.4}
& \textbf{89.7}
& \textbf{12.7}

\\ MTDSI 
% Ensemble
& 99.8
& \textbf{100}
& 68.3
& 81.7
& 89.4
& 15.0

\\ MTDSI+RAP-LS
% Ensemble
& \textbf{100}
& \textbf{100}
& \textbf{85.0}
& \textbf{89.8}
& \textbf{92.3}
& \textbf{25.1}

\\ MTDAI
% Ensemble
& \textbf{100}
& \textbf{100}
& 70.7
& 88.8
& 91.2
& 16.8

\\ MTDAI+RAP-LS
% Ensemble
& \textbf{100}
& \textbf{100}
& \textbf{84.6}
& \textbf{90.4}
& \textbf{91.8}
& \textbf{27.8}

\\

\bottomrule
\end{tabular}
}
% \end{sc}
\end{small}
\end{center}
\vskip -0.1in
\end{table}
% CE Loss for untargted attack
% MaxLogit Loss for targted attack

% \newpage

\subsection{The Experimental Results w.r.t Different Value of Iterations}
\label{whole results}

In the main submission, we report the evaluations of $K=400$.
Here, we further report the performance with different values of $K$ for completeness in Table \ref{targeted_Logit_R_I} (targeted attack) and Table \ref{untargeted_CE_R_I} (untargeted attack).
From the results, we observe that the attacking performance generally increase as $K$ increases for most cases, this is also aligned with prior works \citep{zhao2020success}.
% Yet the larger $K$ also induces the larger computation cost, and we adopt $K=400$ in the main comparison.

\begin{table}[H]
\small{\caption{
The targeted attack success rate ($\%$) of all baseline attacks with our method. The results with logit loss and 10/100/200/300/400 iterations are reported. We highlight the results with $K=400$ in bold.}
\label{targeted_Logit_R_I}}
\vskip -0.15in
\begin{center}
\begin{small}
% \begin{sc}
\scalebox{0.9}{
\begin{tabular}{c|ccc}
\toprule
\multirow{2}{*}{\tabincell{c}{}} & \multicolumn{3}{c}{\textbf{ResNet-50 $\boldsymbol{\rightarrow}$ Inception-v3}}  \\
& Baseline &  +RAP  & +RAP-LS \\
\midrule 
I
& 0.0 / 0.1 / 0.2 / 0.1 / \textbf{0.1}
& 0.0 / 0.2 / 0.3 / 0.3 / \textbf{0.1}
& 0.0 / 0.1 / 0.4 / 0.6 / \textbf{0.7}

\\ MI
& 0.1 / 0.1 / 0.2 / 0.1 / \textbf{0.1}
& 0.0 / 0.6 / 1.0 / 1.0 / \textbf{1.1}
& 0.1 / 0.1 / 1.4 / 1.6 / \textbf{2.4}

\\ TI
& 0.0 / 0.3 / 0.2 / 0.2 / \textbf{0.1}
& 0.0 / 0.7 / 0.9 / 1.2 / \textbf{0.8}
& 0.0 / 0.3 / 1.3 / 1.3 / \textbf{1.2}

\\ DI
& 0.2 / 1.2 / 1.7 / 1.5 / \textbf{1.5}
& 0.0 / 3.8 / 6.6 / 7.7 / \textbf{7.9}
& 0.2 / 1.2 / 10.2 / 9.4 / \textbf{10.1}

\\ SI
& 0.3 / 2.6 / 2.4 / 2.0 / \textbf{1.8}
& 0.2 / 6.6 / 8.2 / 8.6 / \textbf{9.3}
& 0.3 / 2.6 / 9.6 / 9.3 / \textbf{10.5}

\\ Admix
& 1.4 / 5.7 / 5.9 / 6.0 / \textbf{5.8}
& 0.6 / 14.6 / 16.6 / 16.5 / \textbf{17.1}
& 1.4 / 5.7 / 18.5 / 19.2 / \textbf{19.6}

\\ MI-TI-DI
& 1.5 / 7.9 / 9.8 / 10.5 / \textbf{10.9}
& 0.1 / 12.7 / 22.3 / 26.3 / \textbf{28.3}
& 1.5 / 7.9 / 26.8 / 30.0 / \textbf{33.2}

\\ MI-TI-DI-SI
& 8.9 / 34.1 / 36.7 / 38.1 / \textbf{38.1}
& 3.3 / 43.3 / 47.9 / 49.9 / \textbf{51.8}
& 8.9 / 34.8 / 54.8 / 55.8 / \textbf{58.0}

\\ MI-TI-DI-Admix
& 13.5 / 45.7 / 49.2 / 50.5 / \textbf{50.8}
& 5.0 / 48.1 / 53.4 / 56.2 / \textbf{57.1}
& 13.5 / 45.1 / 61.4 / 63.0 / \textbf{64.1}

\\
\bottomrule
\end{tabular} 
}
% \end{sc}
\end{small}
\end{center}
% \vskip -0.15in
\end{table}

\begin{table}[H]
% \vskip -0.15in
\begin{center}
\begin{small}
% \begin{sc}
\scalebox{0.9}{
\begin{tabular}{c|ccc}
\toprule
\multirow{2}{*}{\tabincell{c}{}} & \multicolumn{3}{c}{\textbf{ResNet-50 $\boldsymbol{\rightarrow}$ DenseNet-121}}  \\
& Baseline &  +RAP  & +RAP-LS \\
\midrule 
I
& 0.9 / 5.3 / 5.0 / 5.5 / \textbf{4.5}
& 0.0 / 4.8 / 7.9 / 8.8 / \textbf{9.5}
& 0.9 / 5.3 / 14.0 / 14.0 / \textbf{14.3}

\\ MI
& 3.4 / 6.3 / 6.3 / 6.0 / \textbf{6.3}
& 0.2 / 9.0 / 14.1 / 15.8 / \textbf{17.5}
& 3.4 / 6.3 / 25.9 / 28.9 / \textbf{29.6}

\\ TI
& 2.5 / 8.6 / 8.9 / 9.0 / \textbf{7.2}
& 0.0 / 7.1 / 10.1 / 11.2 / \textbf{11.0}
& 2.5 / 8.6 / 16.1 / 16.4 / \textbf{17.3}

\\ DI
& 8.4 / 54.8 / 60.4 / 61.2 / \textbf{62.6}
& 0.1 / 40.6 / 53.2 / 59.4 / \textbf{64.9}
& 8.4 / 54.6 / 70.9 / 72.5 / \textbf{73.9}

\\ SI
& 9.7 / 29.6 / 30.4 / 30.4 / \textbf{30.0}
& 2.2 / 45.8 / 50.9 / 52.5 / \textbf{53.2}
& 9.7 / 29.6 / 60.0 / 61.1 / \textbf{61.1}

\\ Admix
& 23.6 / 55.6 / 55.5 / 55.6 / \textbf{54.6}
& 5.3 / 61.2 / 66.0 / 66.9 / \textbf{68.0}
& 23.6 / 55.6 / 74.4 / 74.7 / \textbf{74.6}

\\ MI-TI-DI
& 16.3 / 66.9 / 71.4 / 73.4 / \textbf{74.9}
& 1.8 / 56.7 / 71.2 / 76.4 / \textbf{78.2}
& 16.3 / 66.7 / 85.2 / 85.7 / \textbf{88.5}

\\ MI-TI-DI-SI
& 41.0 / 82.8 / 84.5 / 86.2 / \textbf{86.3}
& 12.9 / 80.2 / 85.7 / 87.8 / \textbf{88.4}
& 41.0 / 82.5 / 91.9 / 92.4 / \textbf{93.3}

\\ MI-TI-DI-Admix
& 48.0 / 88.7 / 90.9 / 91.1 / \textbf{91.4}
& 20.3 / 83.2 / 87.2 / 88.4 / \textbf{89.4}
& 47.9 / 88.5 / 93.5 / 93.8 / \textbf{93.6}

\\
\bottomrule
\end{tabular} 
}
\label{targeted_Logit_R_D}
% \end{sc}
\end{small}
\end{center}
% \vskip -0.15in
\end{table}
\begin{table}[H]
% \vskip -0.15in
\begin{center}
\begin{small}
% \begin{sc}
\scalebox{0.9}{
\begin{tabular}{c|ccc}
\toprule
\multirow{2}{*}{\tabincell{c}{}} & \multicolumn{3}{c}{\textbf{ResNet-50 $\boldsymbol{\rightarrow}$ VGG-16}}  \\
& Baseline &  +RAP  & +RAP-LS \\
\midrule 
I
& 1.0 / 2.7 / 2.6 / 2.3 / \textbf{2.4}
& 0.0 / 5.6 / 8.3 / 9.8 / \textbf{9.8}
& 1.0 / 2.7 / 11.4 / 12.8 / \textbf{11.8}

\\ MI
& 1.2 / 2.1 / 2.4 / 2.2 / \textbf{2.2}
& 0.1 / 8.6 / 12.3 / 14.1 / \textbf{14.5}
& 1.2 / 2.1 / 18.2 / 20.0 / \textbf{20.6}

\\ TI
& 1.1 / 4.8 / 4.8 / 4.5 / \textbf{4.0}
& 0.1 / 6.0 / 9.3 / 9.9 / \textbf{12.9}
& 1.1 / 4.8 / 14.2 / 15.3 / \textbf{15.3}

\\ DI
& 7.6 / 51.0 / 56.9 / 56.6 / \textbf{57.2}
& 0.4 / 42.4 / 55.0 / 61.5 / \textbf{63.4}
& 7.6 / 51.0 / 69.3 / 69.8 / \textbf{69.3}

\\ SI
& 4.4 / 10.4 / 8.9 / 8.8 / \textbf{9.5}
& 1.1 / 27.8 / 31.1 / 30.7 / \textbf{32.8}
& 4.4 / 10.4 / 35.8 / 35.1 / \textbf{36.0}

\\ Admix
& 10.6 / 24.9 / 25.0 / 26.2 / \textbf{26.0}
& 3.6 / 41.4 / 45.2 / 43.8 / \textbf{45.4}
& 10.6 / 24.9 / 51.7 / 51.9 / \textbf{51.6}

\\ MI-TI-DI
& 12.1 / 55.9 / 61.0 / 63.9 / \textbf{62.8}
& 1.5 / 53.0 / 64.7 / 70.9 / \textbf{72.9}
& 12.1 / 55.8 / 78.5 / 81.7 / \textbf{81.5}

\\ MI-TI-DI-SI
& 24.6 / 67.4 / 68.5 / 69.7 / \textbf{70.1}
& 8.2 / 66.4 / 73.7 / 75.2 / \textbf{77.7}
& 24.5 / 66.4 / 82.4 / 83.7 / \textbf{84.7}

\\ MI-TI-DI-Admix
& 33.4 / 75.3 / 77.5 / 78.7 / \textbf{79.9}
& 14.6 / 70.4 / 76.7 / 78.3 / \textbf{79.0}
& 33.3 / 75.2 / 85.4 / 86.4 / \textbf{86.3}

\\
\bottomrule
\end{tabular} 
}
\label{targeted_Logit_R_V}
% \end{sc}
\end{small}
\end{center}
% \vskip -0.15in
\end{table}

\begin{table}[H]
% \vskip -0.15in
\begin{center}
\begin{small}
% \begin{sc}
\scalebox{0.9}{
\begin{tabular}{c|ccc}
\toprule
\multirow{2}{*}{\tabincell{c}{}} & \multicolumn{3}{c}{\textbf{DenseNet121 $\boldsymbol{\rightarrow}$ Inception-v3}}  \\
& Baseline &  +RAP  & +RAP-LS \\
\midrule 
I
& 0.0 / 0.1 / 0.2 / 0.1 / \textbf{0.0}
& 0.0 / 0.6 / 0.9 / 0.7 / \textbf{0.8}
& 0.0 / 0.1 / 1.0 / 1.3 / \textbf{1.2}

\\ MI
& 0.2 / 0.2 / 0.3 / 0.3 / \textbf{0.3}
& 0.0 / 1.2 / 2.1 / 2.1 / \textbf{2.0}
& 0.2 / 0.2 / 2.5 / 3.7 / \textbf{3.4}

\\ TI
& 0.0 / 0.4 / 0.3 / 0.5 / \textbf{0.2}
& 0.0 / 1.2 / 1.5 / 1.6 / \textbf{2.1}
& 0.0 / 0.4 / 2.6 / 3.1 / \textbf{3.0}

\\ DI
& 0.3 / 1.9 / 1.4 / 1.7 / \textbf{1.4}
& 0.0 / 4.1 / 7.0 / 7.6 / \textbf{8.8}
& 0.3 / 1.9 / 9.3 / 9.9 / \textbf{10.0}

\\ SI
& 0.3 / 1.5 / 1.8 / 1.6 / \textbf{1.6}
& 0.1 / 7.6 / 9.2 / 10.0 / \textbf{8.5}
& 0.3 / 1.5 / 9.2 / 10.7 / \textbf{10.4}

\\ Admix
& 1.7 / 5.0 / 5.4 / 5.5 / \textbf{5.0}
& 0.2 / 15.8 / 17.0 / 17.7 / \textbf{17.1}
& 1.7 / 5.0 / 18.5 / 18.2 / \textbf{17.6}

\\ MI-TI-DI
& 1.2 / 6.8 / 7.9 / 8.7 / \textbf{7.7}
& 0.1 / 13.0 / 19.7 / 22.2 / \textbf{23.0}
& 1.2 / 6.7 / 21.9 / 26.2 / \textbf{26.5}

\\ MI-TI-DI-SI
& 5.1 / 17.6 / 18.9 / 19.3 / \textbf{19.8}
& 2.0 / 30.4 / 35.1 / 37.0 / \textbf{39.0}
& 5.2 / 17.7 / 36.8 / 38.9 / \textbf{39.2}

\\ MI-TI-DI-Admix
& 11.4 / 30.5 / 32.2 / 31.4 / \textbf{32.0}
& 3.9 / 36.7 / 41.3 / 42.2 / \textbf{43.5}
& 11.2 / 31.2 / 47.2 / 49.2 / \textbf{49.3}

\\
\bottomrule
\end{tabular} 
}
\label{targeted_Logit_D_I}
% \end{sc}
\end{small}
\end{center}
% \vskip -0.15in
\end{table}

\begin{table}[H]
% \vskip -0.15in
\begin{center}
\begin{small}
% \begin{sc}
\scalebox{0.9}{
\begin{tabular}{c|ccc}
\toprule
\multirow{2}{*}{\tabincell{c}{}} & \multicolumn{3}{c}{\textbf{DenseNet121 $\boldsymbol{\rightarrow}$ ResNet-50}}  \\
& Baseline &  +RAP  & +RAP-LS \\
\midrule 
I
& 1.8 / 6.5 / 5.6 / 5.5 / \textbf{5.0}
& 0.2 / 7.7 / 11.2 / 12.4 / \textbf{12.8}
& 1.8 / 6.5 / 18.7 / 19.0 / \textbf{17.9}

\\ MI
& 3.4 / 5.4 / 5.2 / 4.9 / \textbf{4.6}
& 0.3 / 10.2 / 14.3 / 16.3 / \textbf{16.2}
& 3.4 / 5.4 / 23.6 / 26.3 / \textbf{26.5}

\\ TI
& 2.6 / 8.1 / 7.9 / 8.4 / \textbf{8.4}
& 0.2 / 7.8 / 10.9 / 12.1 / \textbf{13.5}
& 2.6 / 8.1 / 19.2 / 20.2 / \textbf{20.8}

\\ DI
& 6.3 / 30.4 / 33.1 / 32.0 / \textbf{30.2}
& 0.4 / 33.6 / 44.1 / 48.7 / \textbf{52.6}
& 6.3 / 30.8 / 58.8 / 60.4 / \textbf{60.4}

\\ SI
& 7.3 / 16.5 / 15.9 / 14.8 / \textbf{14.2}
& 1.5 / 33.8 / 39.5 / 41.4 / \textbf{41.5}
& 7.3 / 16.5 / 44.7 / 44.8 / \textbf{43.4}

\\ Admix
& 16.4 / 32.6 / 30.3 / 28.8 / \textbf{29.3}
& 3.7 / 48.3 / 52.9 / 53.4 / \textbf{53.0}
& 16.4 / 32.6 / 60.1 / 58.8 / \textbf{58.2}

\\ MI-TI-DI
& 8.3 / 40.3 / 44.6 / 46.3 / \textbf{44.9}
& 0.9 / 42.0 / 56.4 / 62.4 / \textbf{64.3}
& 8.3 / 40.1 / 69.5 / 72.8 / \textbf{74.5}

\\ MI-TI-DI-SI
& 18.6 / 52.3 / 54.1 / 56.2 / \textbf{55.0}
& 6.6 / 60.3 / 67.5 / 70.6 / \textbf{71.2}
& 18.6 / 52.5 / 73.8 / 75.5 / \textbf{75.8}

\\ MI-TI-DI-Admix
& 27.6 / 66.3 / 69.7 / 69.8 / \textbf{69.1}
& 12.1 / 66.4 / 70.8 / 73.2 / \textbf{74.2}
& 27.6 / 66.4 / 81.4 / 82.0 / \textbf{82.1}

\\
\bottomrule
\end{tabular} 
}
\label{targeted_Logit_D_R}
% \end{sc}
\end{small}
\end{center}
% \vskip -0.15in
\end{table}

\begin{table}[H]
% \vskip -0.15in
\begin{center}
\begin{small}
% \begin{sc}
\scalebox{0.9}{
\begin{tabular}{c|ccc}
\toprule
\multirow{2}{*}{\tabincell{c}{}} & \multicolumn{3}{c}{\textbf{DenseNet121 $\boldsymbol{\rightarrow}$ VGG-16}}  \\
& Baseline &  +RAP  & +RAP-LS \\
\midrule 
I
& 0.6 / 3.8 / 3.5 / 3.5 / \textbf{2.9}
& 0.1 / 6.2 / 9.3 / 10.5 / \textbf{10.1}
& 0.6 / 3.8 / 14.5 / 15.7 / \textbf{15.9}

\\ MI
& 1.6 / 2.4 / 2.6 / 2.7 / \textbf{3.1}
& 0.2 / 8.6 / 12.2 / 13.0 / \textbf{13.4}
& 1.6 / 2.4 / 19.5 / 21.7 / \textbf{23.2}

\\ TI
& 1.1 / 5.6 / 5.8 / 4.8 / \textbf{5.2}
& 0.1 / 6.3 / 9.1 / 11.0 / \textbf{12.4}
& 1.1 / 5.6 / 16.5 / 17.0 / \textbf{16.4}

\\ DI
& 4.1 / 29.8 / 32.7 / 33.1 / \textbf{32.1}
& 0.2 / 31.5 / 44.7 / 48.7 / \textbf{49.5}
& 4.1 / 29.9 / 57.2 / 56.5 / \textbf{58.9}

\\ SI
& 2.8 / 9.8 / 8.8 / 8.5 / \textbf{8.4}
& 0.6 / 25.8 / 28.2 / 31.4 / \textbf{31.0}
& 2.8 / 9.8 / 33.5 / 35.3 / \textbf{35.2}

\\ Admix
& 10.2 / 23.3 / 22.1 / 21.3 / \textbf{21.5}
& 1.7 / 39.4 / 42.2 / 43.0 / \textbf{42.7}
& 10.2 / 23.3 / 49.7 / 49.3 / \textbf{48.2}

\\ MI-TI-DI
& 6.1 / 32.4 / 36.3 / 39.0 / \textbf{38.5}
& 0.7 / 36.2 / 49.9 / 53.2 / \textbf{55.0}
& 6.1 / 32.6 / 61.8 / 64.3 / \textbf{65.5}

\\ MI-TI-DI-SI
& 12.4 / 40.2 / 41.9 / 42.2 / \textbf{42.0}
& 4.6 / 46.9 / 54.0 / 57.0 / \textbf{58.4}
& 12.4 / 40.0 / 61.3 / 62.4 / \textbf{62.3}

\\ MI-TI-DI-Admix
& 20.0 / 53.2 / 55.0 / 55.7 / \textbf{54.7}
& 9.4 / 54.8 / 60.1 / 61.8 / \textbf{63.1}
& 19.9 / 53.1 / 68.1 / 69.7 / \textbf{69.3}

\\
\bottomrule
\end{tabular} 
}
\label{targeted_Logit_D_V}
% \end{sc}
\end{small}
\end{center}
% \vskip -0.15in
\end{table}

\begin{table}[H]
% \vskip -0.15in
\begin{center}
\begin{small}
% \begin{sc}
\scalebox{0.9}{
\begin{tabular}{c|ccc}
\toprule
\multirow{2}{*}{\tabincell{c}{}} & \multicolumn{3}{c}{\textbf{VGG-16 $\boldsymbol{\rightarrow}$ Inception-v3}}  \\
& Baseline &  +RAP  & +RAP-LS \\
\midrule 
I
& 0.0 / 0.0 / 0.0 / 0.0 / \textbf{0.0}
& 0.0 / 0.1 / 0.0 / 0.1 / \textbf{0.1}
& 0.0 / 0.0 / 0.2 / 0.0 / \textbf{0.2}

\\ MI
& 0.0 / 0.0 / 0.0 / 0.0 / \textbf{0.0}
& 0.0 / 0.0 / 0.2 / 0.0 / \textbf{0.0}
& 0.0 / 0.0 / 0.2 / 0.5 / \textbf{0.3}

\\ TI
& 0.0 / 0.0 / 0.0 / 0.1 / \textbf{0.0}
& 0.0 / 0.1 / 0.1 / 0.1 / \textbf{0.1}
& 0.0 / 0.0 / 0.4 / 0.4 / \textbf{0.4}

\\ DI
& 0.0 / 0.0 / 0.0 / 0.0 / \textbf{0.0}
& 0.0 / 0.0 / 0.4 / 0.6 / \textbf{0.4}
& 0.0 / 0.0 / 0.7 / 0.7 / \textbf{1.1}

\\ SI
& 0.0 / 0.4 / 0.3 / 0.2 / \textbf{0.2}
& 0.0 / 2.0 / 1.5 / 2.0 / \textbf{1.7}
& 0.0 / 0.6 / 1.6 / 1.9 / \textbf{1.8}

\\ Admix
& 0.1 / 0.7 / 0.8 / 0.6 / \textbf{0.7}
& 0.0 / 2.7 / 2.2 / 2.3 / \textbf{2.4}
& 0.1 / 1.0 / 2.3 / 3.0 / \textbf{2.8}

\\ MI-TI-DI
& 0.1 / 1.0 / 0.8 / 1.1 / \textbf{0.7}
& 0.0 / 1.8 / 2.8 / 3.0 / \textbf{3.4}
& 0.1 / 0.9 / 3.4 / 4.0 / \textbf{4.6}

\\ MI-TI-DI-SI
& 1.7 / 7.7 / 9.1 / 9.8 / \textbf{9.6}
& 0.6 / 12.2 / 14.5 / 13.8 / \textbf{15.2}
& 1.7 / 8.6 / 11.4 / 12.1 / \textbf{13.7}

\\ MI-TI-DI-Admix
& 3.6 / 12.4 / 12.2 / 11.5 / \textbf{11.6}
& 1.1 / 14.5 / 16.1 / 15.9 / \textbf{17.1}
& 3.4 / 11.2 / 15.9 / 17.4 / \textbf{17.6}

\\
\bottomrule
\end{tabular} 
}
\label{targeted_Logit_V_I}
% \end{sc}
\end{small}
\end{center}
% \vskip -0.15in
\end{table}

\begin{table}[H]
% \vskip -0.15in
\begin{center}
\begin{small}
% \begin{sc}
\scalebox{0.9}{
\begin{tabular}{c|ccc}
\toprule
\multirow{2}{*}{\tabincell{c}{}} & \multicolumn{3}{c}{\textbf{VGG-16 $\boldsymbol{\rightarrow}$ ResNet-50}}  \\
& Baseline &  +RAP  & +RAP-LS \\
\midrule 
I
& 0.2 / 0.4 / 0.3 / 0.3 / \textbf{0.1}
& 0.0 / 1.0 / 0.8 / 0.8 / \textbf{0.7}
& 0.2 / 0.5 / 1.4 / 1.5 / \textbf{1.4}

\\ MI
& 0.4 / 0.5 / 0.6 / 0.5 / \textbf{0.5}
& 0.2 / 1.1 / 1.3 / 1.3 / \textbf{1.3}
& 0.4 / 0.2 / 2.1 / 2.4 / \textbf{1.9}

\\ TI
& 0.3 / 1.0 / 0.7 / 0.9 / \textbf{0.7}
& 0.0 / 1.4 / 1.5 / 1.4 / \textbf{1.2}
& 0.3 / 1.0 / 3.0 / 3.3 / \textbf{3.2}

\\ DI
& 0.5 / 2.8 / 3.1 / 3.4 / \textbf{2.8}
& 0.0 / 4.9 / 6.7 / 6.5 / \textbf{7.3}
& 0.5 / 3.9 / 9.5 / 10.1 / \textbf{9.7}

\\ SI
& 1.4 / 4.4 / 3.9 / 3.8 / \textbf{3.3}
& 0.4 / 9.2 / 9.0 / 9.1 / \textbf{9.8}
& 1.4 / 4.3 / 10.1 / 9.4 / \textbf{9.8}

\\ Admix
& 4.6 / 7.3 / 6.7 / 5.8 / \textbf{5.6}
& 0.7 / 10.6 / 11.3 / 10.9 / \textbf{11.1}
& 4.7 / 7.3 / 11.6 / 12.5 / \textbf{11.9}

\\ MI-TI-DI
& 1.8 / 10.2 / 11.7 / 11.9 / \textbf{11.8}
& 0.0 / 10.8 / 14.6 / 15.7 / \textbf{16.7}
& 1.8 / 9.5 / 20.2 / 21.6 / \textbf{22.9}

\\ MI-TI-DI-SI
& 8.8 / 30.1 / 31.6 / 30.3 / \textbf{31.0}
& 3.2 / 30.8 / 32.5 / 33.5 / \textbf{35.3}
& 9.0 / 29.5 / 36.9 / 38.5 / \textbf{38.7}

\\ MI-TI-DI-Admix
& 15.2 / 34.6 / 35.1 / 36.6 / \textbf{36.2}
& 5.4 / 34.7 / 37.3 / 38.1 / \textbf{39.0}
& 15.3 / 35.5 / 43.2 / 42.9 / \textbf{43.1}

\\
\bottomrule
\end{tabular} 
}
\label{targeted_Logit_V_R}
% \end{sc}
\end{small}
\end{center}
% \vskip -0.15in
\end{table}

\begin{table}[H]
% \vskip -0.15in
\begin{center}
\begin{small}
% \begin{sc}
\scalebox{0.9}{
\begin{tabular}{c|ccc}
\toprule
\multirow{2}{*}{\tabincell{c}{}} & \multicolumn{3}{c}{\textbf{VGG-16 $\boldsymbol{\rightarrow}$ DenseNet-121}}  \\
& Baseline &  +RAP  & +RAP-LS \\
\midrule 
I
& 0.1 / 0.2 / 0.4 / 0.3 / \textbf{0.2}
& 0.0 / 0.7 / 1.1 / 0.7 / \textbf{1.4}
& 0.1 / 0.3 / 1.2 / 1.5 / \textbf{1.7}

\\ MI
& 0.3 / 0.8 / 0.6 / 0.6 / \textbf{0.5}
& 0.0 / 1.1 / 1.4 / 2.1 / \textbf{2.3}
& 0.3 / 0.6 / 2.4 / 3.2 / \textbf{3.0}

\\ TI
& 0.1 / 0.6 / 1.1 / 1.0 / \textbf{0.8}
& 0.0 / 0.9 / 1.7 / 1.6 / \textbf{1.7}
& 0.1 / 0.9 / 2.5 / 2.7 / \textbf{2.9}

\\ DI
& 0.2 / 3.8 / 4.8 / 4.1 / \textbf{3.8}
& 0.0 / 5.0 / 7.6 / 7.8 / \textbf{8.4}
& 0.2 / 3.7 / 11.9 / 12.2 / \textbf{12.7}

\\ SI
& 1.3 / 9.0 / 8.9 / 7.7 / \textbf{7.2}
& 0.3 / 14.0 / 15.6 / 16.4 / \textbf{16.8}
& 1.3 / 8.2 / 17.0 / 17.4 / \textbf{17.8}

\\ Admix
& 4.9 / 14.3 / 13.4 / 13.2 / \textbf{13.0}
& 0.7 / 17.9 / 20.5 / 20.2 / \textbf{20.2}
& 4.9 / 14.0 / 23.9 / 24.2 / \textbf{23.6}

\\ MI-TI-DI
& 1.5 / 12.1 / 13.4 / 13.9 / \textbf{13.7}
& 0.1 / 9.7 / 15.7 / 17.4 / \textbf{19.4}
& 1.6 / 12.1 / 24.4 / 26.3 / \textbf{27.4}

\\ MI-TI-DI-SI
& 13.0 / 38.9 / 41.5 / 42.8 / \textbf{41.7}
& 3.8 / 37.8 / 42.0 / 43.8 / \textbf{44.4}
& 12.8 / 37.3 / 48.6 / 49.8 / \textbf{49.6}

\\ MI-TI-DI-Admix
& 19.0 / 45.5 / 47.0 / 47.7 / \textbf{48.0}
& 6.8 / 41.3 / 45.2 / 44.8 / \textbf{45.1}
& 19.1 / 45.3 / 52.9 / 54.9 / \textbf{55.2}

\\
\bottomrule
\end{tabular} 
}
\label{targeted_Logit_V_D}
% \end{sc}
\end{small}
\end{center}
% \vskip -0.15in
\end{table}

\begin{table}[H]
% \vskip -0.15in
\begin{center}
\begin{small}
% \begin{sc}
\scalebox{0.9}{
\begin{tabular}{c|ccc}
\toprule
\multirow{2}{*}{\tabincell{c}{}} & \multicolumn{3}{c}{\textbf{Inc-v3 $\boldsymbol{\rightarrow}$ ResNet-50}}  \\
& Baseline &  +RAP  & +RAP-LS \\
\midrule 
I
& 0.2 / 0.4 / 0.3 / 0.1 / \textbf{0.2}
& 0.0 / 0.2 / 0.7 / 0.6 / \textbf{0.9}
& 0.2 / 0.4 / 1.0 / 0.7 / \textbf{0.5}

\\ MI
& 0.1 / 0.3 / 0.3 / 0.2 / \textbf{0.2}
& 0.0 / 0.6 / 1.4 / 1.5 / \textbf{1.7}
& 0.1 / 0.3 / 0.8 / 1.6 / \textbf{1.5}

\\ TI
& 0.2 / 0.3 / 0.2 / 0.2 / \textbf{0.2}
& 0.0 / 0.2 / 0.6 / 0.9 / \textbf{0.5}
& 0.2 / 0.3 / 1.0 / 0.7 / \textbf{0.7}

\\ DI
& 0.2 / 1.5 / 1.4 / 1.9 / \textbf{1.6}
& 0.1 / 2.5 / 4.3 / 4.3 / \textbf{4.6}
& 0.2 / 1.5 / 5.0 / 5.1 / \textbf{6.4}

\\ SI
& 0.3 / 0.3 / 0.3 / 0.6 / \textbf{0.6}
& 0.4 / 1.9 / 2.6 / 2.6 / \textbf{2.9}
& 0.3 / 0.3 / 2.4 / 2.8 / \textbf{2.5}

\\ Admix
& 1.2 / 1.9 / 2.2 / 1.9 / \textbf{1.5}
& 0.6 / 5.0 / 4.9 / 5.2 / \textbf{4.9}
& 1.2 / 1.9 / 5.7 / 5.7 / \textbf{5.2}

\\ MI-TI-DI
& 0.6 / 1.6 / 2.0 / 2.4 / \textbf{1.8}
& 0.0 / 4.2 / 6.3 / 7.7 / \textbf{8.3}
& 0.6 / 1.7 / 6.2 / 7.0 / \textbf{7.5}

\\ MI-TI-DI-SI
& 1.5 / 4.7 / 5.5 / 5.8 / \textbf{5.6}
& 0.7 / 8.6 / 10.3 / 11.1 / \textbf{11.9}
& 1.5 / 5.0 / 10.0 / 9.6 / \textbf{10.7}

\\ MI-TI-DI-Admix
& 2.8 / 8.9 / 9.5 / 9.6 / \textbf{9.6}
& 1.4 / 12.6 / 14.0 / 13.6 / \textbf{13.6}
& 2.8 / 8.6 / 14.5 / 15.1 / \textbf{16.7}

\\
\bottomrule
\end{tabular} 
}
\label{targeted_Logit_I_R}
% \end{sc}
\end{small}
\end{center}
% \vskip -0.15in
\end{table}

\begin{table}[H]
% \vskip -0.15in
\begin{center}
\begin{small}
% \begin{sc}
\scalebox{0.9}{
\begin{tabular}{c|ccc}
\toprule
\multirow{2}{*}{\tabincell{c}{}} & \multicolumn{3}{c}{\textbf{Inc-v3 $\boldsymbol{\rightarrow}$ DenseNet-121}}  \\
& Baseline &  +RAP  & +RAP-LS \\
\midrule 
I
& 0.0 / 0.0 / 0.2 / 0.0 / \textbf{0.2}
& 0.0 / 0.2 / 0.4 / 0.6 / \textbf{0.6}
& 0.0 / 0.0 / 0.2 / 0.4 / \textbf{0.3}

\\ MI
& 0.0 / 0.1 / 0.2 / 0.1 / \textbf{0.1}
& 0.1 / 0.7 / 1.0 / 1.1 / \textbf{1.6}
& 0.0 / 0.1 / 1.0 / 1.1 / \textbf{1.5}

\\ TI
& 0.0 / 0.3 / 0.2 / 0.0 / \textbf{0.1}
& 0.0 / 0.3 / 0.3 / 0.3 / \textbf{0.7}
& 0.0 / 0.3 / 0.9 / 0.9 / \textbf{0.6}

\\ DI
& 0.1 / 1.3 / 2.5 / 3.0 / \textbf{2.8}
& 0.0 / 2.7 / 4.4 / 5.4 / \textbf{5.8}
& 0.1 / 1.3 / 5.9 / 7.0 / \textbf{7.5}

\\ SI
& 0.2 / 0.7 / 0.9 / 0.8 / \textbf{0.9}
& 0.0 / 2.4 / 3.3 / 2.9 / \textbf{2.7}
& 0.2 / 0.7 / 3.2 / 3.1 / \textbf{3.2}

\\ Admix
& 1.1 / 2.6 / 2.5 / 2.3 / \textbf{2.0}
& 0.5 / 7.2 / 7.7 / 7.0 / \textbf{6.9}
& 1.1 / 2.6 / 8.2 / 7.3 / \textbf{7.5}

\\ MI-TI-DI
& 0.5 / 3.1 / 3.8 / 4.5 / \textbf{4.1}
& 0.2 / 5.4 / 10.8 / 12.6 / \textbf{14.8}
& 0.5 / 3.3 / 10.6 / 11.8 / \textbf{13.4}

\\ MI-TI-DI-SI
& 1.9 / 9.0 / 9.4 / 9.5 / \textbf{10.4}
& 1.1 / 15.5 / 19.8 / 19.8 / \textbf{21.2}
& 1.9 / 9.0 / 19.1 / 20.2 / \textbf{20.9}

\\ MI-TI-DI-Admix
& 4.6 / 15.7 / 16.8 / 17.4 / \textbf{17.9}
& 2.4 / 23.2 / 24.5 / 26.6 / \textbf{27.5}
& 4.6 / 15.0 / 29.1 / 30.2 / \textbf{31.6}

\\
\bottomrule
\end{tabular} 
}
\label{targeted_Logit_I_D}
% \end{sc}
\end{small}
\end{center}
% \vskip -0.15in
\end{table}

\begin{table}[H]
% \vskip -0.15in
\begin{center}
\begin{small}
% \begin{sc}
\scalebox{0.9}{
\begin{tabular}{c|ccc}
\toprule
\multirow{2}{*}{\tabincell{c}{}} & \multicolumn{3}{c}{\textbf{Inc-v3 $\boldsymbol{\rightarrow}$ VGG-16}}  \\
& Baseline &  +RAP  & +RAP-LS \\
\midrule 
I
& 0.0 / 0.3 / 0.1 / 0.1 / \textbf{0.1}
& 0.0 / 0.2 / 0.8 / 0.6 / \textbf{0.5}
& 0.0 / 0.3 / 0.2 / 0.5 / \textbf{0.5}

\\ MI
& 0.1 / 0.1 / 0.2 / 0.2 / \textbf{0.2}
& 0.1 / 0.4 / 0.8 / 1.2 / \textbf{1.3}
& 0.1 / 0.1 / 0.4 / 0.8 / \textbf{1.0}

\\ TI
& 0.1 / 0.2 / 0.2 / 0.1 / \textbf{0.2}
& 0.1 / 0.4 / 0.5 / 0.6 / \textbf{0.8}
& 0.1 / 0.2 / 0.6 / 0.6 / \textbf{0.6}

\\ DI
& 0.3 / 2.0 / 2.8 / 2.3 / \textbf{2.6}
& 0.1 / 1.8 / 4.3 / 5.2 / \textbf{6.3}
& 0.3 / 2.0 / 6.8 / 7.3 / \textbf{8.1}

\\ SI
& 0.0 / 0.7 / 0.6 / 0.4 / \textbf{0.5}
& 0.2 / 2.0 / 1.5 / 1.5 / \textbf{1.5}
& 0.0 / 0.7 / 1.6 / 2.3 / \textbf{2.3}

\\ Admix
& 0.5 / 1.6 / 1.0 / 1.0 / \textbf{1.3}
& 0.4 / 3.2 / 3.8 / 4.1 / \textbf{3.3}
& 0.5 / 1.6 / 4.5 / 3.8 / \textbf{4.4}

\\ MI-TI-DI
& 0.3 / 2.0 / 2.4 / 2.7 / \textbf{2.9}
& 0.1 / 3.8 / 6.5 / 7.3 / \textbf{8.0}
& 0.3 / 2.0 / 8.0 / 7.9 / \textbf{9.8}

\\ MI-TI-DI-SI
& 0.7 / 3.7 / 3.6 / 4.1 / \textbf{4.2}
& 0.5 / 7.6 / 7.5 / 8.5 / \textbf{8.9}
& 0.7 / 3.2 / 6.7 / 8.1 / \textbf{8.6}

\\ MI-TI-DI-Admix
& 2.3 / 6.9 / 8.3 / 8.6 / \textbf{8.4}
& 1.3 / 10.5 / 12.5 / 12.0 / \textbf{12.0}
& 2.3 / 7.0 / 11.9 / 12.8 / \textbf{12.1}

\\
\bottomrule
\end{tabular} 
}
\label{targeted_Logit_I_V}
% \end{sc}
\end{small}
\end{center}
% \vskip -0.15in
\end{table}

\begin{table}[H]
\small{\caption{
The untargeted attack success rate ($\%$) of all baseline attacks with RAP. The results with $CE$ loss and 10/100/200/300/400 iterations are reported. We highlight the results with $K=400$ in bold.}
\label{untargeted_CE_R_I}}
% \vskip -0.15in
\begin{center}
\begin{small}
% \begin{sc}
\scalebox{0.9}{
\begin{tabular}{c|ccc}
\toprule
\multirow{2}{*}{\tabincell{c}{}} & \multicolumn{3}{c}{\textbf{ResNet-50 $\boldsymbol{\rightarrow}$ Inception-v3}}  \\
& Baseline &  +RAP  & +RAP-LS \\
\midrule 
I
% e = 16/255, k = 8
% & 25.9 / 35.5 / 35.3 / 34.7 / \textbf{34.6}
% & 11.1 / 37.5 / 43.8 / 45.0 / \textbf{45.9}
% & 25.9 / 35.5 / 40.0 / 43.3 / \textbf{46.2}

% e = 12/255, k = 6
& 25.9 / 35.5 / 35.3 / 34.7 / \textbf{34.6}
& 12.3 / 48.3 / 54.1 / 55.5 / \textbf{57.0}
& 25.7 / 36.0 / 54.1 / 56.5 / \textbf{57.2}

\\ MI
& 53.2 / 50.7 / 51.0 / 50.6 / \textbf{50.3}
& 26.2 / 58.7 / 68.9 / 73.4 / \textbf{75.9}
& 53.2 / 50.7 / 64.3 / 73.6 / \textbf{77.4}

\\ TI
% e = 16/255, k = 8
% & 30.0 / 45.3 / 44.0 / 45.3 / \textbf{45.5}
% & 14.2 / 44.5 / 50.2 / 53.1 / \textbf{54.3}
% & 30.0 / 45.3 / 47.4 / 50.5 / \textbf{51.9}

% e = 12/255, k = 6
& 30.0 / 45.3 / 44.0 / 45.3 / \textbf{45.5}
& 16.4 / 57.9 / 63.9 / 64.6 / \textbf{66.1}
& 30.0 / 45.1 / 62.3 / 65.3 / \textbf{67.0}

\\ DI
& 46.0 / 60.5 / 59.5 / 59.4 / \textbf{57.7}
& 27.3 / 80.7 / 82.8 / 83.4 / \textbf{82.9}
& 46.0 / 61.0 / 86.0 / 85.7 / \textbf{85.0}

\\ SI
& 50.1 / 66.0 / 65.6 / 66.0 / \textbf{65.9}
& 60.6 / 80.5 / 80.9 / 80.9 / \textbf{79.7}
& 49.9 / 66.6 / 85.2 / 85.0 / \textbf{84.4}

\\ Admix
& 66.6 / 78.7 / 79.2 / 78.0 / \textbf{77.7}
& 73.9 / 87.6 / 87.0 / 86.8 / \textbf{87.4}
& 67.6 / 79.4 / 91.8 / 92.3 / \textbf{92.6}

\\ MI-TI-DI
& 82.1 / 85.8 / 86.4 / 85.9 / \textbf{85.7}
& 61.9 / 93.9 / 95.3 / 95.6 / \textbf{96.0}
& 82.1 / 85.8 / 95.9 / 96.4 / \textbf{96.9}

\\ MI-TI-DI-SI
& 94.2 / 96.8 / 97.2 / 97.0 / \textbf{97.0}
& 92.3 / 98.9 / 98.9 / 99.0 / \textbf{99.1}
& 94.2 / 96.7 / 99.0 / 99.3 / \textbf{99.1}

\\ MI-TI-DI-Admix
& 97.3 / 98.6 / 98.5 / 98.5 / \textbf{98.3}
& 95.1 / 99.4 / 99.4 / 99.3 / \textbf{99.2}
& 97.3 / 98.5 / 99.8 / 99.8 / \textbf{99.8}

\\
\bottomrule
\end{tabular} 
}
% \end{sc}
\end{small}
\end{center}
% \vskip -0.15in
\end{table}

\begin{table}[H]
% \vskip -0.15in
\begin{center}
\begin{small}
% \begin{sc}
\scalebox{0.9}{
\begin{tabular}{c|ccc}
\toprule
\multirow{2}{*}{\tabincell{c}{}} & \multicolumn{3}{c}{\textbf{ResNet-50 $\boldsymbol{\rightarrow}$ DenseNet-121}}  \\
& Baseline &  +RAP  & +RAP-LS \\
\midrule 
I
% e = 16/255, k = 8
% & 67.4 / 79.9 / 79.1 / 79.0 / \textbf{79.2}
% & 21.9 / 68.2 / 74.6 / 79.6 / \textbf{81.8}
% & 67.4 / 79.9 / 70.9 / 75.6 / \textbf{79.5}

% e = 12/255, k = 6
& 67.4 / 79.9 / 79.1 / 79.0 / \textbf{79.2}
& 26.7 / 84.8 / 91.1 / 90.8 / \textbf{91.5}
& 67.8 / 80.1 / 89.8 / 91.3 / \textbf{91.9}

\\ MI
& 87.3 / 85.4 / 86.4 / 85.9 / \textbf{85.8}
& 45.2 / 85.3 / 91.3 / 93.9 / \textbf{95.0}
& 87.3 / 85.4 / 90.8 / 95.0 / \textbf{96.1}

\\ TI
% e = 16/255, k = 8
% & 73.2 / 83.0 / 82.2 / 81.6 / \textbf{82.0}
% & 24.5 / 73.3 / 80.1 / 82.0 / \textbf{82.9}
% & 73.2 / 83.0 / 75.2 / 80.2 / \textbf{81.6}

% e = 12/255, k = 6
& 73.2 / 83.0 / 82.2 / 81.6 / \textbf{82.0}
& 30.9 / 87.3 / 91.5 / 93.3 / \textbf{94.1}
& 72.9 / 82.4 / 90.9 / 94.2 / \textbf{95.1}

\\ DI
& 92.8 / 98.9 / 99.2 / 99.0 / \textbf{99.0}
& 52.6 / 99.0 / 99.6 / 99.7 / \textbf{99.6}
& 92.8 / 99.0 / 99.6 / 99.7 / \textbf{99.7}

\\ SI
& 89.1 / 95.7 / 95.6 / 95.3 / \textbf{94.9}
& 91.3 / 98.9 / 99.0 / 99.2 / \textbf{98.9}
& 89.1 / 95.7 / 99.7 / 99.7 / \textbf{99.7}

\\ Admix
& 96.6 / 98.9 / 98.5 / 98.1 / \textbf{97.9}
& 96.2 / 99.6 / 99.6 / 99.6 / \textbf{99.6}
& 96.4 / 98.5 / 99.9 / 99.9 / \textbf{99.9}

\\ MI-TI-DI
& 98.2 / 99.7 / 99.8 / 99.8 / \textbf{99.8}
& 86.4 / 99.9 / 100 / 100 / \textbf{100}
& 98.2 / 99.7 / 99.9 / 100 / \textbf{100}

\\ MI-TI-DI-SI
& 99.8 / 100 / 100 / 100 / \textbf{100}
& 98.8 / 100 / 100 / 100 / \textbf{100}
& 99.8 / 100 / 100 / 100 / \textbf{100}

\\ MI-TI-DI-Admix
& 99.9 / 100 / 100 / 100 / \textbf{100}
& 99.5 / 100 / 100 / 100 / \textbf{100}
& 99.9 / 100 / 100 / 100 / \textbf{100}

\\
\bottomrule
\end{tabular} 
}
\label{untargeted_CE_R_D}
% \end{sc}
\end{small}
\end{center}
% \vskip -0.15in
\end{table}

\begin{table}[H]
% \vskip -0.15in
\begin{center}
\begin{small}
% \begin{sc}
\scalebox{0.9}{
\begin{tabular}{c|ccc}
\toprule
\multirow{2}{*}{\tabincell{c}{}} & \multicolumn{3}{c}{\textbf{ResNet-50 $\boldsymbol{\rightarrow}$ VGG-16}}  \\
& Baseline &  +RAP  & +RAP-LS \\
\midrule 
I
% e = 16/255, k = 8
% & 68.2 / 77.4 / 78.1 / 77.4 / \textbf{78.0}
% & 28.0 / 69.5 / 77.5 / 78.6 / \textbf{79.9}
% & 68.2 / 77.4 / 74.7 / 79.0 / \textbf{81.1}

% e = 12/255, k = 6
& 68.2 / 77.4 / 78.1 / 77.4 / \textbf{78.0}
& 36.2 / 84.6 / 89.2 / 90.7 / \textbf{91.1}
& 68.4 / 77.3 / 87.1 / 90.9 / \textbf{92.9}

\\ MI
& 82.5 / 82.8 / 82.9 / 82.7 / \textbf{82.4}
& 53.1 / 85.5 / 92.2 / 93.1 / \textbf{93.9}
& 82.5 / 82.8 / 89.3 / 93.7 / \textbf{94.5}

\\ TI
% e = 16/255, k = 8
% & 70.6 / 80.5 / 79.8 / 80.8 / \textbf{81.0}
% & 32.7 / 74.0 / 78.8 / 79.9 / \textbf{82.7}
% & 70.6 / 80.5 / 75.6 / 79.4 / \textbf{80.6}

% e = 12/255, k = 6
& 70.6 / 80.5 / 79.8 / 80.8 / \textbf{81.0}
& 39.3 / 86.9 / 90.6 / 92.5 / \textbf{93.1}
& 71.1 / 80.0 / 89.0 / 91.9 / \textbf{93.3}

\\ DI
& 92.3 / 99.1 / 99.1 / 99.0 / \textbf{99.0}
& 64.4 / 99.4 / 99.7 / 99.7 / \textbf{99.6}
& 92.3 / 99.1 / 99.8 / 99.9 / \textbf{99.7}

\\ SI
& 82.2 / 90.0 / 88.9 / 89.6 / \textbf{88.6}
& 81.3 / 95.7 / 95.8 / 95.7 / \textbf{95.7}
& 82.1 / 89.3 / 97.7 / 97.8 / \textbf{97.2}

\\ Admix
& 92.3 / 95.4 / 96.0 / 95.6 / \textbf{95.8}
& 91.6 / 97.9 / 98.4 / 97.8 / \textbf{97.7}
& 92.7 / 95.9 / 98.9 / 99.0 / \textbf{99.0}

\\ MI-TI-DI
& 97.9 / 99.7 / 99.7 / 99.8 / \textbf{99.8}
& 85.9 / 99.5 / 100 / 100 / \textbf{100}
& 97.9 / 99.7 / 99.9 / 99.9 / \textbf{99.9}

\\ MI-TI-DI-SI
& 99.1 / 99.8 / 99.8 / 99.7 / \textbf{99.7}
& 97.4 / 99.7 / 99.9 / 99.9 / \textbf{99.9}
& 99.1 / 99.8 / 99.8 / 99.8 / \textbf{99.8}

\\ MI-TI-DI-Admix
& 99.2 / 99.8 / 99.8 / 99.8 / \textbf{99.8}
& 98.5 / 99.7 / 99.9 / 99.9 / \textbf{99.9}
& 99.2 / 99.8 / 99.9 / 99.9 / \textbf{99.9}

\\
\bottomrule
\end{tabular} 
}
\label{untargeted_CE_R_V}
% \end{sc}
\end{small}
\end{center}
% \vskip -0.15in
\end{table}

\begin{table}[H]
% \vskip -0.15in
\begin{center}
\begin{small}
% \begin{sc}
\scalebox{0.9}{
\begin{tabular}{c|ccc}
\toprule
\multirow{2}{*}{\tabincell{c}{}} & \multicolumn{3}{c}{\textbf{DenseNet-121 $\boldsymbol{\rightarrow}$ Inception-v3}}  \\
& Baseline &  +RAP  & +RAP-LS \\
\midrule 
I
% e = 16/255, k = 8
% & 31.2 / 48.5 / 46.9 / 46.3 / \textbf{46.5}
% & 13.5 / 44.2 / 48.6 / 52.0 / \textbf{52.4}
% & 31.2 / 48.5 / 48.8 / 51.8 / \textbf{54.2}

% e = 12/255, k = 6
& 31.2 / 48.5 / 46.9 / 46.3 / \textbf{46.5}
& 18.0 / 54.9 / 58.1 / 59.8 / \textbf{60.2}
& 31.6 / 46.9 / 58.9 / 61.0 / \textbf{61.1}

\\ MI
& 56.8 / 58.8 / 59.3 / 60.6 / \textbf{59.3}
& 32.2 / 65.6 / 74.1 / 78.9 / \textbf{80.4}
& 56.8 / 58.8 / 74.6 / 80.0 / \textbf{82.8}

\\ TI
% e = 16/255, k = 8
% & 37.7 / 54.0 / 55.1 / 54.6 / \textbf{54.2}
% & 15.0 / 51.4 / 57.2 / 58.6 / \textbf{60.4}
% & 37.7 / 54.0 / 54.1 / 57.7 / \textbf{59.5}

% e = 12/255, k = 6
& 37.7 / 54.0 / 55.1 / 54.6 / \textbf{54.2}
& 20.4 / 61.0 / 64.7 / 67.3 / \textbf{66.7}
& 38.2 / 54.5 / 65.4 / 67.6 / \textbf{70.0}

\\ DI
& 51.0 / 67.9 / 68.3 / 66.7 / \textbf{67.6}
& 31.4 / 84.0 / 86.8 / 86.7 / \textbf{86.6}
& 51.0 / 68.0 / 89.0 / 88.8 / \textbf{86.9}

\\ SI
& 54.7 / 71.5 / 71.6 / 70.3 / \textbf{71.6}
& 61.1 / 82.9 / 83.1 / 83.5 / \textbf{83.2}
& 53.9 / 71.0 / 86.4 / 87.0 / \textbf{87.4}

\\ Admix
& 72.5 / 82.0 / 82.6 / 82.2 / \textbf{82.0}
& 73.0 / 89.9 / 90.3 / 89.5 / \textbf{89.8}
& 71.7 / 82.8 / 93.9 / 93.2 / \textbf{93.8}

\\ MI-TI-DI
& 81.5 / 89.7 / 89.8 / 89.4 / \textbf{89.1}
& 62.5 / 94.8 / 96.8 / 97.1 / \textbf{97.1}
& 81.5 / 89.6 / 96.1 / 96.9 / \textbf{97.1}

\\ MI-TI-DI-SI
& 92.3 / 95.2 / 94.9 / 95.1 / \textbf{95.1}
& 88.6 / 97.7 / 98.0 / 98.0 / \textbf{98.3}
& 92.4 / 95.2 / 97.8 / 98.5 / \textbf{98.4}

\\ MI-TI-DI-Admix
& 95.8 / 97.7 / 97.2 / 97.3 / \textbf{97.9}
& 93.2 / 98.6 / 98.6 / 99.0 / \textbf{98.8}
& 95.4 / 97.6 / 99.0 / 98.9 / \textbf{98.9}

\\
\bottomrule
\end{tabular} 
}
\label{untargeted_CE_D_I}
% \end{sc}
\end{small}
\end{center}
% \vskip -0.15in
\end{table}

\begin{table}[H]
% \vskip -0.15in
\begin{center}
\begin{small}
% \begin{sc}
\scalebox{0.9}{
\begin{tabular}{c|ccc}
\toprule
\multirow{2}{*}{\tabincell{c}{}} & \multicolumn{3}{c}{\textbf{DenseNet-121 $\boldsymbol{\rightarrow}$ ResNet-50}}  \\
& Baseline &  +RAP  & +RAP-LS \\
\midrule 
I
% e = 16/255, k = 8
% & 76.1 / 88.0 / 87.5 / 87.1 / \textbf{87.4}
% & 28.2 / 77.5 / 82.8 / 85.4 / \textbf{86.1}
% & 76.1 / 88.2 / 80.4 / 83.8 / \textbf{86.2}

% e = 12/255, k = 6
& 76.1 / 88.0 / 87.5 / 87.1 / \textbf{87.4}
& 35.7 / 90.1 / 93.5 / 93.2 / \textbf{94.2}
& 76.1 / 88.0 / 91.2 / 92.9 / \textbf{94.3}

\\ MI
& 87.7 / 90.5 / 91.2 / 90.8 / \textbf{90.3}
& 55.6 / 91.1 / 96.2 / 96.9 / \textbf{97.6}
& 87.7 / 90.5 / 95.4 / 97.2 / \textbf{97.9}

\\ TI
% e = 16/255, k = 8
% & 79.2 / 90.4 / 90.0 / 89.9 / \textbf{89.6}
% & 27.6 / 79.0 / 84.6 / 86.7 / \textbf{87.7}
% & 79.2 / 90.4 / 82.2 / 85.0 / \textbf{87.5}

% e = 12/255, k = 6
& 79.2 / 90.4 / 90.0 / 89.9 / \textbf{89.6}
& 36.9 / 90.1 / 93.2 / 95.0 / \textbf{94.2}
& 79.0 / 89.8 / 92.7 / 94.3 / \textbf{94.8}

\\ DI
& 91.1 / 98.0 / 98.3 / 98.2 / \textbf{98.2}
& 57.0 / 98.6 / 99.3 / 99.7 / \textbf{99.6}
& 91.1 / 98.0 / 99.5 / 99.6 / \textbf{99.7}

\\ SI
& 89.6 / 95.2 / 94.8 / 95.3 / \textbf{95.1}
& 83.0 / 96.5 / 96.7 / 96.3 / \textbf{96.9}
& 89.4 / 95.0 / 98.7 / 98.8 / \textbf{98.8}

\\ Admix
& 96.3 / 97.6 / 97.7 / 97.7 / \textbf{97.0}
& 90.9 / 98.8 / 98.8 / 99.0 / \textbf{99.0}
& 95.7 / 97.9 / 99.3 / 99.2 / \textbf{99.2}

\\ MI-TI-DI
& 96.3 / 99.3 / 99.5 / 99.4 / \textbf{99.4}
& 84.4 / 99.2 / 99.8 / 99.8 / \textbf{99.8}
& 96.3 / 99.2 / 99.8 / 99.9 / \textbf{100}

\\ MI-TI-DI-SI
& 98.3 / 99.7 / 99.8 / 99.8 / \textbf{99.8}
& 95.8 / 99.7 / 99.9 / 99.9 / \textbf{99.9}
& 98.3 / 99.7 / 99.9 / 99.9 / \textbf{99.9}

\\ MI-TI-DI-Admix
& 99.2 / 99.7 / 99.8 / 99.8 / \textbf{99.8}
& 97.9 / 99.9 / 99.8 / 99.8 / \textbf{99.8}
& 99.0 / 99.7 / 99.9 / 99.9 / \textbf{99.9}

\\
\bottomrule
\end{tabular} 
}
\label{untargeted_CE_D_R}
% \end{sc}
\end{small}
\end{center}
% \vskip -0.15in
\end{table}

\begin{table}[H]
% \vskip -0.15in
\begin{center}
\begin{small}
% \begin{sc}
\scalebox{0.9}{
\begin{tabular}{c|ccc}
\toprule
\multirow{2}{*}{\tabincell{c}{}} & \multicolumn{3}{c}{\textbf{DenseNet-121 $\boldsymbol{\rightarrow}$ VGG-16}}  \\
& Baseline &  +RAP  & +RAP-LS \\
\midrule 
I
% e = 16/255, k = 8
% & 75.1 / 84.7 / 85.2 / 84.9 / \textbf{85.1}
% & 33.4 / 77.4 / 82.4 / 84.2 / \textbf{84.6}
% & 75.1 / 84.7 / 79.7 / 82.6 / \textbf{84.6}

% e = 12/255, k = 6
& 75.1 / 84.7 / 85.2 / 84.9 / \textbf{85.1}
& 42.2 / 87.5 / 90.7 / 91.2 / \textbf{91.7}
& 75.1 / 84.6 / 89.2 / 91.7 / \textbf{92.8}

\\ MI
& 85.1 / 87.2 / 88.6 / 87.9 / \textbf{87.5}
& 58.4 / 90.2 / 93.7 / 95.1 / \textbf{96.0}
& 85.1 / 87.2 / 94.2 / 97.0 / \textbf{97.6}

\\ TI
% e = 16/255, k = 8
% & 74.4 / 86.3 / 86.4 / 87.3 / \textbf{87.0}
% & 34.8 / 79.4 / 82.5 / 84.5 / \textbf{85.3}
% & 74.4 / 86.3 / 81.0 / 82.8 / \textbf{85.9}

% e = 12/255, k = 9
& 74.4 / 86.3 / 86.4 / 87.3 / \textbf{87.0}
& 44.2 / 87.8 / 89.6 / 91.0 / \textbf{92.1}
& 74.5 / 85.8 / 90.3 / 92.2 / \textbf{93.3}

\\ DI
& 90.8 / 98.0 / 98.4 / 98.1 / \textbf{98.1}
& 63.3 / 98.6 / 99.2 / 99.6 / \textbf{99.4}
& 90.8 / 97.9 / 99.4 / 99.2 / \textbf{99.4}

\\ SI
& 84.2 / 91.5 / 91.4 / 91.4 / \textbf{91.9}
& 78.5 / 93.9 / 94.5 / 95.2 / \textbf{95.0}
& 83.9 / 91.6 / 96.9 / 97.1 / \textbf{97.5}

\\ Admix
& 93.5 / 95.7 / 96.0 / 96.1 / \textbf{95.6}
& 87.8 / 97.4 / 97.5 / 97.6 / \textbf{97.7}
& 92.0 / 96.1 / 98.9 / 98.7 / \textbf{98.6}

\\ MI-TI-DI
& 95.1 / 99.0 / 99.2 / 99.2 / \textbf{99.2}
& 84.2 / 99.1 / 99.4 / 99.5 / \textbf{99.5}
& 95.1 / 99.0 / 99.9 / 100 / \textbf{100}

\\ MI-TI-DI-SI
& 97.9 / 99.5 / 99.4 / 99.4 / \textbf{99.2}
& 93.3 / 99.0 / 99.2 / 99.3 / \textbf{99.3}
& 97.9 / 99.4 / 99.7 / 99.7 / \textbf{99.7}

\\ MI-TI-DI-Admix
& 98.4 / 99.4 / 99.4 / 99.5 / \textbf{99.4}
& 96.1 / 99.7 / 99.7 / 99.6 / \textbf{99.6}
& 98.3 / 99.4 / 99.8 / 99.7 / \textbf{99.8}

\\
\bottomrule
\end{tabular} 
}
\label{untargeted_CE_D_V}
% \end{sc}
\end{small}
\end{center}
% \vskip -0.15in
\end{table}

\begin{table}[H]
% \vskip -0.15in
\begin{center}
\begin{small}
% \begin{sc}
\scalebox{0.9}{
\begin{tabular}{c|ccc}
\toprule
\multirow{2}{*}{\tabincell{c}{}} & \multicolumn{3}{c}{\textbf{VGG-16 $\boldsymbol{\rightarrow}$ Inception-v3}}  \\
& Baseline &  +RAP  & +RAP-LS \\
\midrule 
I
% e = 16/255, k = 8
% & 14.3 / 22.2 / 22.0 / 22.2 / \textbf{22.0}
% & 7.9 / 17.2 / 19.7 / 20.9 / \textbf{19.9}
% & 14.7 / 22.0 / 18.5 / 19.8 / \textbf{20.7}

% e = 12/255, k = 6
& 14.3 / 22.2 / 22.0 / 22.2 / \textbf{22.0}
& 9.4 / 23.8 / 26.1 / 23.7 / \textbf{24.7}
& 14.4 / 21.8 / 24.1 / 25.4 / \textbf{24.9}

\\ MI
& 32.3 / 31.3 / 31.0 / 30.1 / \textbf{30.0}
& 16.4 / 30.4 / 36.9 / 42.0 / \textbf{42.7}
& 32.4 / 30.7 / 35.0 / 39.2 / \textbf{42.2}

\\ TI
% e = 16/255, k = 8
% & 18.7 / 30.2 / 29.6 / 29.7 / \textbf{29.1}
% & 9.9 / 22.6 / 28.6 / 30.0 / \textbf{31.9}
% & 18.7 / 29.4 / 24.2 / 28.6 / \textbf{31.5}

% e = 12/255, k = 6
& 18.7 / 30.2 / 29.6 / 29.7 / \textbf{29.1}
& 11.9 / 32.1 / 35.7 / 34.9 / \textbf{36.2}
& 18.3 / 29.3 / 34.2 / 36.0 / \textbf{37.1}

\\ DI
& 18.1 / 29.7 / 29.9 / 30.4 / \textbf{29.9}
& 14.2 / 43.6 / 46.1 / 46.5 / \textbf{46.6}
& 18.0 / 29.2 / 50.1 / 51.5 / \textbf{51.6}

\\ SI
& 31.0 / 45.1 / 46.1 / 45.1 / \textbf{45.8}
& 46.7 / 70.9 / 72.0 / 73.4 / \textbf{74.0}
& 31.0 / 44.6 / 73.0 / 74.3 / \textbf{74.7}

\\ Admix
& 40.2 / 54.9 / 55.5 / 54.9 / \textbf{55.5}
& 57.0 / 78.0 / 77.6 / 77.9 / \textbf{77.6}
& 41.4 / 56.0 / 80.0 / 79.9 / \textbf{80.8}

\\ MI-TI-DI
& 50.7 / 55.9 / 57.2 / 56.7 / \textbf{56.8}
& 41.9 / 74.0 / 79.0 / 81.5 / \textbf{82.6}
& 50.7 / 56.4 / 77.8 / 80.0 / \textbf{81.4}

\\ MI-TI-DI-SI
& 77.6 / 85.3 / 85.7 / 85.0 / \textbf{85.0}
& 85.5 / 93.1 / 93.7 / 94.2 / \textbf{94.1}
& 78.0 / 85.0 / 94.4 / 94.6 / \textbf{95.2}

\\ MI-TI-DI-Admix
& 84.7 / 89.4 / 89.2 / 89.9 / \textbf{89.3}
& 88.4 / 94.9 / 95.1 / 95.2 / \textbf{95.0}
& 85.8 / 90.1 / 94.8 / 95.4 / \textbf{95.5}

\\
\bottomrule
\end{tabular} 
}
\label{untargeted_CE_V_I}
% \end{sc}
\end{small}
\end{center}
% \vskip -0.15in
\end{table}

\begin{table}[H]
% \vskip -0.15in
\begin{center}
\begin{small}
% \begin{sc}
\scalebox{0.9}{
\begin{tabular}{c|ccc}
\toprule
\multirow{2}{*}{\tabincell{c}{}} & \multicolumn{3}{c}{\textbf{VGG-16 $\boldsymbol{\rightarrow}$ ResNet-50}}  \\
& Baseline &  +RAP  & +RAP-LS \\
\midrule 
I
% e = 16/255, k = 8
% & 37.2 / 52.0 / 53.4 / 53.1 / \textbf{53.7}
% & 15.6 / 37.1 / 42.1 / 41.9 / \textbf{44.5}
% & 37.5 / 51.8 / 40.4 / 42.7 / \textbf{43.4}

% e = 12/255, k = 6
& 37.2 / 52.0 / 53.4 / 53.1 / \textbf{53.7}
& 17.8 / 48.5 / 53.9 / 53.7 / \textbf{53.0}
& 38.1 / 53.0 / 52.4 / 54.8 / \textbf{54.2}

\\ MI
& 60.2 / 64.3 / 63.5 / 62.0 / \textbf{62.5}
& 32.9 / 57.1 / 67.6 / 73.1 / \textbf{76.2}
& 60.4 / 62.0 / 66.3 / 73.2 / \textbf{76.4}

\\ TI
% e = 16/255, k = 8
% & 45.3 / 62.7 / 63.6 / 62.5 / \textbf{62.8}
% & 17.2 / 44.5 / 50.1 / 52.7 / \textbf{52.9}
% & 45.4 / 62.2 / 47.2 / 50.7 / \textbf{52.1}

% e = 12/255, k = 6
& 45.3 / 62.7 / 63.6 / 62.5 / \textbf{62.8}
& 19.4 / 56.6 / 63.0 / 65.6 / \textbf{64.8}
& 46.0 / 62.9 / 63.5 / 65.8 / \textbf{65.8}

\\ DI
& 51.5 / 72.9 / 73.2 / 72.5 / \textbf{72.2}
& 29.6 / 80.9 / 85.0 / 86.4 / \textbf{86.0}
& 51.4 / 73.8 / 88.9 / 89.2 / \textbf{88.8}

\\ SI
& 64.6 / 81.0 / 80.2 / 80.5 / \textbf{80.0}
& 68.1 / 91.9 / 92.3 / 92.4 / \textbf{92.7}
& 64.9 / 80.6 / 95.1 / 95.3 / \textbf{94.7}

\\ Admix
& 76.8 / 87.5 / 88.2 / 88.0 / \textbf{87.3}
& 79.4 / 93.8 / 94.4 / 95.2 / \textbf{94.6}
& 77.6 / 88.3 / 96.6 / 96.8 / \textbf{96.8}

\\ MI-TI-DI
& 81.1 / 89.9 / 89.8 / 90.3 / \textbf{90.0}
& 66.7 / 94.6 / 96.3 / 96.9 / \textbf{97.2}
& 81.4 / 88.5 / 96.5 / 97.3 / \textbf{97.7}

\\ MI-TI-DI-SI
& 95.1 / 97.6 / 98.0 / 97.9 / \textbf{97.6}
& 94.7 / 98.4 / 98.8 / 98.9 / \textbf{98.8}
& 95.2 / 97.5 / 99.3 / 99.4 / \textbf{99.4}

\\ MI-TI-DI-Admix
& 97.2 / 98.1 / 98.0 / 98.1 / \textbf{97.8}
& 96.1 / 99.1 / 99.2 / 99.3 / \textbf{99.2}
& 97.3 / 98.6 / 99.5 / 99.6 / \textbf{99.6}

\\
\bottomrule
\end{tabular} 
}
\label{untargeted_CE_V_R}
% \end{sc}
\end{small}
\end{center}
% \vskip -0.15in
\end{table}

\begin{table}[H]
% \vskip -0.15in
\begin{center}
\begin{small}
% \begin{sc}
\scalebox{0.9}{
\begin{tabular}{c|ccc}
\toprule
\multirow{2}{*}{\tabincell{c}{}} & \multicolumn{3}{c}{\textbf{VGG-16 $\boldsymbol{\rightarrow}$ DenseNet-121}}  \\
& Baseline &  +RAP  & +RAP-LS \\
\midrule 
I
% e = 16/255, k = 8
% & 35.4 / 50.4 / 49.8 / 48.4 / \textbf{49.1}
% & 13.4 / 32.0 / 37.7 / 39.6 / \textbf{41.8}
% & 35.2 / 49.7 / 34.8 / 38.8 / \textbf{40.6}

% e = 12/255, k = 6
& 35.4 / 50.4 / 49.8 / 48.4 / \textbf{49.1}
& 15.4 / 46.0 / 49.6 / 50.5 / \textbf{50.6}
& 35.2 / 50.3 / 49.7 / 52.9 / \textbf{51.4}

\\ MI
& 62.1 / 63.8 / 62.8 / 61.7 / \textbf{60.5}
& 26.6 / 51.1 / 63.4 / 70.0 / \textbf{73.0}
& 61.6 / 62.5 / 62.7 / 70.5 / \textbf{73.9}

\\ TI
% e = 16/255, k = 8
% & 43.5 / 58.6 / 58.7 / 57.2 / \textbf{55.9}
% & 17.0 / 40.6 / 48.2 / 50.4 / \textbf{49.1}
% & 44.4 / 58.6 / 44.2 / 47.9 / \textbf{50.1}

% e = 12/255, k = 6
& 43.5 / 58.6 / 58.7 / 57.2 / \textbf{55.9}
& 19.4 / 55.8 / 62.7 / 63.0 / \textbf{63.7}
& 44.3 / 58.3 / 60.3 / 63.8 / \textbf{62.1}

\\ DI
& 48.1 / 70.2 / 68.9 / 70.0 / \textbf{68.8}
& 26.5 / 79.9 / 82.3 / 84.2 / \textbf{85.0}
& 47.9 / 70.5 / 85.1 / 87.2 / \textbf{87.2}

\\ SI
& 65.3 / 82.3 / 82.4 / 82.0 / \textbf{82.1}
& 71.3 / 93.3 / 93.7 / 94.4 / \textbf{94.8}
& 65.5 / 82.2 / 95.2 / 95.4 / \textbf{95.7}

\\ Admix
& 79.6 / 89.4 / 88.6 / 88.4 / \textbf{88.2}
& 83.5 / 96.1 / 95.9 / 96.2 / \textbf{96.4}
& 79.2 / 88.9 / 97.4 / 97.4 / \textbf{97.2}

\\ MI-TI-DI
& 80.3 / 87.0 / 88.7 / 89.3 / \textbf{88.8}
& 62.9 / 94.0 / 95.9 / 96.4 / \textbf{97.0}
& 80.4 / 86.8 / 96.8 / 97.2 / \textbf{97.3}

\\ MI-TI-DI-SI
& 95.3 / 98.2 / 98.4 / 98.4 / \textbf{98.1}
& 95.9 / 99.2 / 99.2 / 99.2 / \textbf{99.2}
& 95.4 / 98.2 / 99.5 / 99.5 / \textbf{99.4}

\\ MI-TI-DI-Admix
& 97.1 / 98.6 / 98.8 / 99.1 / \textbf{98.9}
& 97.4 / 99.4 / 99.6 / 99.6 / \textbf{99.5}
& 97.3 / 98.5 / 99.5 / 99.5 / \textbf{99.6}

\\
\bottomrule
\end{tabular} 
}
\label{untargeted_CE_V_D}
% \end{sc}
\end{small}
\end{center}
% \vskip -0.15in
\end{table}

\begin{table}[H]
% \vskip -0.15in
\begin{center}
\begin{small}
% \begin{sc}
\scalebox{0.9}{
\begin{tabular}{c|ccc}
\toprule
\multirow{2}{*}{\tabincell{c}{}} & \multicolumn{3}{c}{\textbf{Inc-v3 $\boldsymbol{\rightarrow}$ ResNet-50}}  \\
& Baseline &  +RAP  & +RAP-LS \\
\midrule 
I
% e = 16/255, k = 8
% & 34.0 / 48.4 / 51.2 / 50.1 / \textbf{51.5}
% & 17.5 / 50.9 / 53.6 / 55.0 / \textbf{55.3}
% & 34.0 / 48.4 / 52.7 / 54.2 / \textbf{55.8}

% e = 12/255, k = 6
& 34.0 / 48.4 / 51.2 / 50.1 / \textbf{51.5}
& 22.7 / 58.6 / 60.9 / 61.1 / \textbf{62.1}
& 34.5 / 49.0 / 60.2 / 60.5 / \textbf{62.0}

\\ MI
& 58.5 / 59.1 / 60.4 / 60.3 / \textbf{62.0}
& 43.8 / 77.0 / 81.7 / 84.0 / \textbf{85.8}
& 58.5 / 59.1 / 80.0 / 82.6 / \textbf{84.8}

\\ TI
% e = 16/255, k = 8
% & 33.6 / 46.9 / 48.7 / 48.5 / \textbf{49.3}
% & 19.8 / 50.8 / 54.2 / 52.9 / \textbf{55.3}
% & 33.6 / 46.9 / 52.5 / 54.2 / \textbf{55.1}

% e = 12/255, k = 6
& 33.6 / 46.9 / 48.7 / 48.5 / \textbf{49.3}
& 21.8 / 58.9 / 60.2 / 61.7 / \textbf{63.4}
& 33.1 / 47.2 / 59.5 / 61.5 / \textbf{61.6}

\\ DI
& 48.4 / 65.8 / 67.2 / 68.4 / \textbf{68.4}
& 33.3 / 78.8 / 81.4 / 81.4 / \textbf{81.7}
& 48.3 / 65.7 / 80.7 / 82.3 / \textbf{81.8}

\\ SI
& 43.7 / 61.9 / 63.9 / 65.1 / \textbf{66.2}
& 45.8 / 67.0 / 69.4 / 69.5 / \textbf{69.8}
& 43.6 / 62.3 / 72.5 / 73.4 / \textbf{72.8}

\\ Admix
& 56.1 / 73.0 / 75.9 / 76.9 / \textbf{75.9}
& 57.0 / 77.5 / 79.8 / 80.3 / \textbf{80.2}
& 56.3 / 73.4 / 82.9 / 84.0 / \textbf{84.9}

\\ MI-TI-DI
& 72.2 / 79.5 / 81.9 / 81.9 / \textbf{82.9}
& 61.2 / 88.1 / 90.8 / 91.9 / \textbf{91.8}
& 72.2 / 79.4 / 89.9 / 91.5 / \textbf{90.6}

\\ MI-TI-DI-SI
& 82.9 / 88.3 / 88.3 / 88.4 / \textbf{89.0}
& 83.5 / 90.8 / 91.2 / 90.6 / \textbf{91.2}
& 82.8 / 88.1 / 91.9 / 92.6 / \textbf{92.3}

\\ MI-TI-DI-Admix
& 89.8 / 91.6 / 91.3 / 91.4 / \textbf{91.5}
& 89.0 / 93.9 / 94.0 / 94.0 / \textbf{94.1}
& 89.6 / 92.3 / 94.1 / 94.8 / \textbf{94.7}

\\
\bottomrule
\end{tabular} 
}
\label{untargeted_CE_I_R}
% \end{sc}
\end{small}
\end{center}
% \vskip -0.15in
\end{table}

\begin{table}[H]
% \vskip -0.15in
\begin{center}
\begin{small}
% \begin{sc}
\scalebox{0.9}{
\begin{tabular}{c|ccc}
\toprule
\multirow{2}{*}{\tabincell{c}{}} & \multicolumn{3}{c}{\textbf{Inc-v3 $\boldsymbol{\rightarrow}$ DenseNet-121}}  \\
& Baseline &  +RAP  & +RAP-LS \\
\midrule 
I
% e = 16/255, k = 8
% & 35.2 / 47.2 / 47.3 / 46.8 / \textbf{48.7}
% & 19.2 / 48.9 / 53.1 / 53.5 / \textbf{54.8}
% & 35.2 / 47.2 / 49.9 / 52.7 / \textbf{54.8}

% e = 12/255, k = 6
& 35.2 / 47.2 / 47.3 / 46.8 / \textbf{48.7}
& 21.4 / 54.3 / 57.2 / 59.4 / \textbf{60.8}
& 34.9 / 47.5 / 58.7 / 58.8 / \textbf{60.0}

\\ MI
& 57.4 / 56.2 / 56.5 / 56.8 / \textbf{56.7}
& 42.9 / 74.0 / 80.1 / 82.5 / \textbf{84.6}
& 57.4 / 56.2 / 77.4 / 81.9 / \textbf{84.6}

\\ TI
% e = 16/255, k = 8
% & 35.8 / 48.6 / 47.8 / 48.9 / \textbf{49.4}
% & 20.1 / 51.7 / 57.1 / 57.1 / \textbf{58.5}
% & 35.8 / 48.6 / 52.9 / 55.1 / \textbf{56.9}

% e = 12/255, k = 6
& 35.8 / 48.6 / 47.8 / 48.9 / \textbf{49.4}
& 22.1 / 59.6 / 63.3 / 65.7 / \textbf{63.4}
& 35.5 / 48.7 / 61.6 / 64.2 / \textbf{63.8}

\\ DI
& 53.2 / 72.1 / 71.8 / 71.5 / \textbf{71.9}
& 35.7 / 81.9 / 83.7 / 85.1 / \textbf{85.0}
& 53.2 / 71.8 / 84.1 / 85.2 / \textbf{84.0}

\\ SI
& 46.6 / 63.7 / 65.1 / 65.9 / \textbf{65.9}
& 52.6 / 72.4 / 73.5 / 74.5 / \textbf{74.9}
& 46.6 / 63.0 / 77.7 / 77.9 / \textbf{77.2}

\\ Admix
& 60.5 / 76.7 / 78.0 / 79.3 / \textbf{78.5}
& 63.9 / 83.2 / 83.4 / 84.1 / \textbf{83.7}
& 61.9 / 76.9 / 87.7 / 87.3 / \textbf{87.4}

\\ MI-TI-DI
& 76.7 / 84.7 / 85.7 / 85.7 / \textbf{85.7}
& 65.1 / 91.5 / 92.8 / 94.0 / \textbf{94.2}
& 76.7 / 84.6 / 92.6 / 92.9 / \textbf{93.3}

\\ MI-TI-DI-SI
& 89.0 / 91.9 / 91.7 / 91.8 / \textbf{92.0}
& 89.0 / 94.7 / 95.6 / 95.2 / \textbf{95.2}
& 89.0 / 91.4 / 95.1 / 95.4 / \textbf{95.6}

\\ MI-TI-DI-Admix
& 93.5 / 95.5 / 95.9 / 95.1 / \textbf{95.4}
& 93.3 / 96.8 / 96.9 / 96.4 / \textbf{96.2}
& 94.1 / 95.5 / 97.2 / 97.5 / \textbf{97.6}

\\
\bottomrule
\end{tabular} 
}
\label{untargeted_CE_I_D}
% \end{sc}
\end{small}
\end{center}
% \vskip -0.15in
\end{table}

\begin{table}[H]
% \vskip -0.15in
\begin{center}
\begin{small}
% \begin{sc}
\scalebox{0.9}{
\begin{tabular}{c|ccc}
\toprule
\multirow{2}{*}{\tabincell{c}{}} & \multicolumn{3}{c}{\textbf{Inc-v3 $\boldsymbol{\rightarrow}$ VGG-16}}  \\
& Baseline &  +RAP  & +RAP-LS \\
\midrule 
I
% e = 16/255, k = 8
% & 39.9 / 53.1 / 54.1 / 53.7 / \textbf{55.1}
% & 25.8 / 53.6 / 59.2 / 60.2 / \textbf{60.4}
% & 39.9 / 53.1 / 58.3 / 59.8 / \textbf{60.9}

% e = 12/255, k = 6
& 39.9 / 53.1 / 54.1 / 53.7 / \textbf{55.1}
& 29.1 / 63.0 / 65.8 / 66.9 / \textbf{65.9}
& 39.7 / 52.6 / 65.6 / 68.3 / \textbf{68.0}

\\ MI
& 60.7 / 62.2 / 63.8 / 62.1 / \textbf{63.1}
& 50.7 / 76.1 / 81.0 / 83.6 / \textbf{84.9}
& 60.7 / 62.2 / 79.8 / 84.0 / \textbf{84.6}

\\ TI
% e = 16/255, k = 8
% & 41.6 / 55.1 / 55.2 / 55.3 / \textbf{58.1}
% & 29.6 / 57.5 / 60.3 / 63.1 / \textbf{63.6}
% & 41.6 / 55.1 / 60.3 / 61.7 / \textbf{63.0}

% e = 12/255, k = 6
& 41.6 / 55.1 / 55.2 / 55.3 / \textbf{58.1}
& 31.1 / 65.9 / 67.1 / 68.2 / \textbf{68.6}
& 41.5 / 55.1 / 66.3 / 68.0 / \textbf{69.5}

\\ DI
& 54.9 / 73.4 / 74.5 / 76.0 / \textbf{76.1}
& 44.4 / 83.4 / 84.7 / 85.0 / \textbf{85.2}
& 54.9 / 73.0 / 85.7 / 87.2 / \textbf{86.4}

\\ SI
& 46.7 / 62.4 / 64.4 / 65.7 / \textbf{66.0}
& 47.4 / 67.6 / 69.2 / 68.6 / \textbf{69.2}
& 46.3 / 64.1 / 72.4 / 72.1 / \textbf{73.0}

\\ Admix
& 57.3 / 73.2 / 72.8 / 74.0 / \textbf{74.5}
& 57.3 / 75.4 / 75.9 / 77.5 / \textbf{77.2}
& 55.3 / 73.4 / 82.6 / 82.2 / \textbf{83.5}

\\ MI-TI-DI
& 74.7 / 82.7 / 84.7 / 84.6 / \textbf{85.1}
& 67.7 / 90.0 / 91.9 / 92.3 / \textbf{92.7}
& 74.7 / 82.5 / 90.4 / 90.8 / \textbf{91.0}

\\ MI-TI-DI-SI
& 79.8 / 88.0 / 87.6 / 87.5 / \textbf{87.6}
& 81.6 / 89.0 / 89.4 / 89.4 / \textbf{90.3}
& 79.7 / 87.8 / 92.4 / 92.5 / \textbf{92.9}

\\ MI-TI-DI-Admix
& 87.9 / 89.7 / 90.7 / 91.4 / \textbf{91.4}
& 87.0 / 92.2 / 92.3 / 92.5 / \textbf{93.2}
& 87.7 / 91.7 / 94.5 / 94.6 / \textbf{94.1}

\\
\bottomrule
\end{tabular} 
}
\label{untargeted_CE_I_V}
% \end{sc}
\end{small}
\end{center}
% \vskip -0.15in
\end{table}

\end{document}